%% file: main.tex
\documentclass[12pt, onehalfspace, oneside]{aucklandthesis}
\usepackage[utf8]{inputenc}
\usepackage[T1]{fontenc}
\usepackage{amsmath,amssymb,amsfonts,mathtools,dsfont, nicefrac}
\usepackage{graphicx} % for inserting graphics files
\usepackage{appendix} % for appendices
\usepackage{hyperref} % for formatting web addresses and other URLs
\usepackage{booktabs} % for table rules
\usepackage{pdfpages} % include pdf pages for co-authorship forms
\usepackage{lmodern} % for large chapter numbers
\usepackage[explicit]{titlesec} % for my custom titles
\usepackage{xcolor}
\usepackage{colortbl}

\usepackage{adjustbox}
\usepackage{arydshln}
\usepackage{ragged2e}
\usepackage{colortbl}

\hypersetup{
    colorlinks,
    linkcolor={black},
    citecolor={blue!50!black},
    urlcolor={blue!50!black}
}
\newlength\chapnumb
\titleformat{\chapter}[display] {}{}{0pt}
{\parbox[b]{\chapnumb}{%
   \fontsize{120}{110}\selectfont\thechapter}%
  \parbox[b]{\dimexpr\textwidth-\chapnumb\relax}{%
    \raggedleft%
    \hfill{\bfseries\LARGE#1}\\
    \rule{\dimexpr\textwidth-\chapnumb\relax}{0.4pt}}}
\titleformat{name=\chapter,numberless}[display] {}{}{0pt}
{\parbox[b]{\chapnumb}{%
   \mbox{}}%
  \parbox[b]{\dimexpr\textwidth-\chapnumb\relax}{%
    \raggedleft%
    \hfill{\bfseries\LARGE#1}\\
    \rule{\dimexpr\textwidth-\chapnumb\relax}{0.4pt}}}
    
\usepackage[font=footnotesize]{caption} % size of all captions; "small" is the next bigger option.

\makeatletter
\renewcommand\tableofcontents{%
  \chapter*{\contentsname}
  \@mkboth{\MakeUppercase\contentsname}{\MakeUppercase\contentsname}%
  \@starttoc{toc}%
}
\makeatother
\usepackage{xspace} % for correct spacing after definitions
\DeclareMathOperator{\argmax}{arg\,max\xspace}

\usepackage{natbib}
\usepackage{bibentry}
\nobibliography*

\usepackage{csquotes} % for nice indentation

\usepackage[inline]{enumitem}
\makeatletter
\DeclareRobustCommand{\ddots}
   {\sbox0{.}
    \mathinner{
    \raise\dimexpr2.82\p@+.94\wd0\relax\copy0
    \kern\dimexpr1.41\p@-.19\wd0\relax
    \raise\dimexpr1.41\p@+.47\wd0\relax\copy0
    \kern\dimexpr1.41\p@-.19\wd0\relax
    \box0}}
\DeclareRobustCommand{\vdots}
   {\sbox0{.}%
    \vbox{%
      \offinterlineskip
      \kern2\p@ %<-- this makes the box somewhat higher (plain TeX and LaTeX have 4\p@) 
      \hbox{.}%
      \kern\dimexpr2\p@+.666\wd0\relax
      \hbox{.}%
      \kern\dimexpr2\p@+.666\wd0\relax
      \hbox{.}}}
\makeatother

\begin{document}

% ====================================================
%
% FRONTMATTER
%
% Arabic pagination, starting with the title page
% which is counted but not numbered
%
% ====================================================

% Specify the title page content
\title{Teaching Smaller Language Models To Generalise To Unseen Compositional Questions}
%\subtitle{}
%\subtitle{Identification and Mitigation \\-- Beyond Sperm}
\author{Timothy John Hartill}
\degreesought{Doctor of Philosophy} 
\degreediscipline{Computer Science}
% REMOVE LATER!!
\degreecompletionyear{2024}
%\vspace{1em} This thesis is for examination purposes only and is confidential to the examination process.

% Print the title page
\maketitle

\pagenumbering{roman}
\pagestyle{plain}

% Abstract, up to 350 words
\input{Contents/abstract} % it's in a separate file

% Dedication (optional)
%\thesisdedication{Dedicated to Robbie.}

% Preface and/or acknowledgements (optional)
\clearpage\input{Contents/acknowledgements} % it's in a separate file

% Contents, lists of tables and figures
\settocdepth{subsection} % choose chapter, section, subsection 
\begin{KeepFromToc}
    \clearpage\tableofcontents
\end{KeepFromToc}
%\cleardoublepage\listoffigures %(optional)
%\cleardoublepage\listoftables %(optional)

% Glossary (optional)
%\input{glossary}

% CO-AUTHORSHIP FORMS

%\clearpage
%    \addcontentsline{toc}{chapter}{Statements of Contribution}
%    \includepdf[picturecommand={
%    \begin{tikzpicture}[overlay]
%        \node[draw=none] at (11em,47em) {\ref{sec:memorise}};
%    \end{tikzpicture}
%  }]{Co-Authorship_Forms/Co-Authorship_ICDM20.pdf}
%\clearpage\includepdf[picturecommand={
%    \begin{tikzpicture}[overlay]
%        \node[draw=none] at (10em,46.3em) {\ref{sec:ratd}};
%    \end{tikzpicture}
%  }]{Co-Authorship_Forms/Co-Authorship_PAKDD22.pdf}
%\clearpage\includepdf[picturecommand={
%    \begin{tikzpicture}[overlay]
%        \node[draw=none] at (11.5em,46.6em) {\ref{sec:combine}};
%        \node[draw=none] at (21em,48em) {Under Review; Preprint available at \href{https://doi.org/10.21203/rs.3.rs-2133331/v1}{doi.org/10.21203/rs.3.rs-2133331/v1}};
%    \end{tikzpicture}
%  }]{Co-Authorship_Forms/CO-Authorship_CANCELS.pdf}

%\includepdf{Co-Authorship_Forms/co-authorship-form-unseen-compositional-questions-v2-final}
%\includepdf{Co-Authorship_Forms/co-authorship-form-combining-rationales-dense-retrieval-v2}
%\includepdf{Co-Authorship_Forms/co-authorship-form-memorisation-v2_arxiv}

% ====================================================
%
% MAINMATTER
%
% Include external chapter files here using
% the \input{} command
%
% If you run out of memory during compilation,
% switch some or all chapters to \include{} instead of \input{}, 
% but watch out for pagination problems.
%
% ====================================================
\pagenumbering{arabic}

% ====================================================
%
% TIKZ TESTING AREA
% \cleardoublepage
% \begin{figure*}[t]
%     \centering
%     \resizebox{0.9\columnwidth}{!}{\input{Tikz/mimic_overview.tikz}}
% \end{figure*}
% \cleardoublepage
%
% ====================================================

\input{Contents/introduction}   % ~ 5-10 pages
\input{Contents/prelim}         % 
\input{Contents/relatedwork}    % ~ 20 pages
\input{Contents/chapter1_memorise}

\input{Contents/chapter2_ratd}

\input{Contents/chapter3_combine}

\input{Contents/conclusion}     % ~ 20 pages; include summary, contributions, future work, and limitations

% ====================================================
%
% ENDMATTER
%
% Appendices and bibliography 
% Pagination arabic, re-starts at 1
%
% ====================================================
\cleardoublepage % start afresh on a new page
\appendixpage % makes a page to mark beginning of appendices
\input{Contents/appendix1}

\bibliography{references.bib}
% {\footnotesize \bibliography{references}}

%\printbibliography%[title={Works cited}, heading=bibintoc]

\end{document}

%% file: Contents/abstract.tex
\chapter*{Abstract}

% aim, methods, findings, conclusions (no more than 350 words)
%  words

We are inspired by recent progress with pretrained large Language Models (LLMs), that are able to answer questions that are unlikely to have been encountered during training. However a diversity of potential applications exist in the broad domain of reasoning systems and considerations such as latency, cost, available compute resource and internet connectivity are relevant in determining an appropriate approach. We consider the setting where \textit{some} local compute capacity is available at inference time but internet connectivity is not. 

Similar to a general-purpose LLM, we assume that our much smaller Reasoning Models may be asked arbitrary questions from unknown distributions, hence we focus on evaluation in an \textit{unseen} setting where our evaluation datasets are disjoint from our training datasets. We equip our models to answer diverse questions through multitask training focused on instilling an ability to reason over a provided context to an answer. We acquire this context from two knowledge sources; a local Wikipedia corpus queried using a multi-hop dense retrieval system with novel extensions, and from rationales generated from a larger Language Model optimised to run in a lower resource environment.

Our main contributions to the study of question-answering in this setting are as follows: We propose novel methods to evaluate whether our model is capable of answering contextualised questions without memorisation, and show that it is. We establish a comprehensive set of baseline results on unseen evaluation datasets. We show that the addition of novel retrieval-augmented training datasets (\textit{RATD}) to the training regime of the Reasoning Model in conjunction with our retrieval system significantly improves results. We demonstrate further significant improvement through the application of methods for combining contextual knowledge from our two sources. The first method (\textit{RR}) involves training a novel Rationale Ranking model to score both generated rationales and retrieved contexts with respect to relevance and truthfulness. We then use the scores to derive combined contexts from both knowledge sources using a number of  strategies. We also show that utilising the \textit{RATD} datasets enables our model to become proficient at utilising information from combined contexts both separately and in conjunction with the \textit{RR} method.

%% file: Contents/acknowledgements.tex
\chapter*{Acknowledgements}

I am especially grateful to Pat Riddle whose guidance and tireless efforts were essential in maintaining a high standard in our experiments and in our writing. Pat's enthusiasm for rigorous scientific research was an inspiration to me throughout this endeavour. 

Thanks also to my many collaborators, particularly Neset Tan, Diana Benavides-Prado and Josh Bensemann who provided valuable feedback and suggestions at critical junctures.  

I am grateful to the authors of \citet{Pi2022-um} for providing their unreleased POET-SQL dataset and to Omar Khattab for similarly providing his Hover paragraph sequencing data.

Finally, to my wife Clare and my daughters Simone and Sophie, thank you for your fortitude, endless support, and patience throughout the journey.

%% file: Contents/introduction.tex
\chapter{Introduction}
\label{sec:intro}

\section{Background and Motivation}
\label{sec:intro:motivation_background}

When prompted with task demonstrations \citep{Brown2020-rl}, instructions \citep{Sanh2021-na, Wei2021-go, Ouyang2022-ti} or reasoning chains \citep{Wei2022-lz}, large Language Models (LLMs) have shown an ability to answer diverse questions unlikely to have been encountered during training \citep{Brown2020-rl, Sanh2021-na, Wei2021-go, Du2022-ld, Chowdhery2022-yw}. While impressive, this performance has required access to considerable computational resource, typically centralised and accessed over a network that is assumed to be continuously available. In this thesis, we consider the implications and opportunities that an alternative scenario might present; one in which internet connectivity is assumed to be unreliable, unavailable, or merely prohibitively expensive. To make progress in this scenario, utilising technology widely available at the time of writing, we assume that some local compute capacity is available at inference time, namely the equivalent of a single workstation with a large consumer-grade GPU card. Such resource-constrained environments are abundant, ranging from vehicles and fixed locations without continuous internet access, to sensitive applications involving highly confidential information not shareable over the internet. 

In our constrained environment, we utilise a smaller Language Model that can be run locally on our workstation to answer questions. We define smaller Language Models as generative Transformer models \citep{Vaswani2017-tr} with 400 million to 1 billion trainable parameters, i.e those that are large enough to be effective at answering questions whilst being able to perform training and inference with reasonable latency, cost and energy efficiency. We boldly assume that like a general-purpose LLM, our smaller Language Models will be expected to answer arbitrary questions from unknown distributions. This is uncommon in that, excepting \citet{Khashabi2020-gq}, few papers have reported zero-shot results for smaller Language Models, focusing instead on optimising performance via finetuning for particular tasks. However, duplication between test and training splits in natural language processing (NLP) datasets is frequent \citep{Lewis2021-ia, Lee2022-zm, Krishna2021-ue, Kambhatla2023-fc}, which leads to conjecture as to what exactly a model has learned in the fine-tuned setting. In addition to the possibility of answer leakage from directly memorised training samples, it has been shown that models are able to utilise more subtle cues, such as the writing style of a particular annotator who contributed to both train and test splits, for better results than are achievable where the test split is truly independent of the training split \citep{Geva2019-ll}. To minimise such issues as well as to facilitate comparison in a similar setting as other zero/few shot studies, we define an unseen question as one from an evaluation dataset that is disjoint from our training datasets.

LLMs have been shown to have strong performance in answering questions that are input without any supporting context i.e. open domain questions \citep{Roberts2020-ab}. By contrast, smaller Language Models, such as the BART model \citep{Lewis2020-gt} that we use throughout our experiments, are poor at answering such uncontextualised questions, particularly when the evaluation question is not a paraphrase of a memorised training sample \citep{Lewis2021-ia}. An alternative approach, which we follow and extend, has been to use the question text to query a knowledge source and retrieve information pertinent to answering the question. The problem is then transformed into a reading comprehension (RC) challenge whereby the question and the acquired context are input into a Language Model that would \textit{preferably} reason over the question and the provided context to infer an answer (hereafter, called a Reasoning Model). 

In the belief that regardless of how comprehensive any available knowledge source may be, there will be many questions that cannot be answered using information from a single retrieved document, we focus our study on \textit{compositional} questions. The classical \citet{Partee1984-xl} definition of compositionality as an ability to build up the meaning of a complex expression by combining the meanings of its parts has been challenging in practice to usefully apply to natural language tasks such as machine translation and our question-answering topic \citep{Dankers2022-bw, Hupkes2020-iu}. Recent work has alternatively described compositional or ``complex'' questions as those where answering requires decomposition into multiple reasoning steps \citep{Talmor2018-uy, Geva2021-sl}, or reasoning (sometimes termed composition) over more than one piece of information \citep{Yang2018-xq, Min2019-wt, Khot2020-sv, Zhao2023-zr}. The skills involved in such reasoning are diverse and multidimensional \citep{Rogers2023-xq}, encompassing for example fact composition \citep{Khot2020-sv}, numerical reasoning \citep{Dua2019-td, Zhao2023-zr}, logical operations \citep{Clark2020-qy, Sinha2019-qa} or set operations \citep{Sen2022-lv}. Noting that the complexity of reasoning needed is a function of both the question and the available evidence \citep{Min2019-wt}, and that Language Model training data is itself a source of evidence, we offer a modestly revised definition of a compositional question as follows: 

\begin{displayquote}
A question is compositional if it is unlikely to be answerable by our Reasoning Model with a memorised answer from a similar training example, or by retrieving any single document from any available knowledge source. 
\end{displayquote}

Here, a knowledge source refers to training data for any Language Model we utilise or the textual corpus accessed by our retrieval system. A document refers to an individual training sample, or corpus paragraph respectively.

Our first knowledge source is a corpus consisting of English Wikipedia paragraphs. Methods for retrieving information from such textual corpora have a long history in the information retrieval domain generally e.g. \citet{Sparck_Jones1972-dm}, and more recently for augmenting open domain questions \citep{Chen2017-gw, Karpukhin2020-wa}. In regard to the latter, early studies focused on the single-hop case where a single document from the corpus typically provides sufficient evidence to enable answering the question in a deductively valid fashion. This work has subsequently been extended to retrieval for multi-hop questions where multiple documents from the corpus are necessary to answer the question \citep{Qi2021-sm, Xiong2021-ex}. Here studies have focused on datasets such as HotpotQA \citep{Yang2018-xq} where the necessary number of documents, henceforth $n$, has often been limited to two. In our work, we extend $n$ to an arbitrary number of documents and introduce an Evidence Set Scoring model whose purpose is to quantify the sufficiency of the information accumulated up to each hop for answering a question. 

Corpora such as Wikipedia contain large amounts of factual information and it might be expected that effective retrieval from such sources would provide good information for answering questions of a factual nature. However such knowledge sources have been shown to be less effective for identifying other types of information such as commonsense, or ``world'' knowledge \citep{Piktus2021-lu}. We therefore evaluate a second knowledge source in combination with the first; rationales generated by larger Language Models conditioned on the question text. We define a rationale as a free-text explanation \citep{Wiegreffe2021-go} of approximately one to three sentences that aims to provide sufficient evidence from which to deduce an answer. Querying a LLM over the internet to generate rationales would of course defeat our purpose, but we study the case where a larger Language Model can be optimised to run in our constrained environment.

\section{Research Problem}
\label{sec:intro:problem}

The setting defined above poses a number of under-explored challenges that form the basis of our research. These can be summarised as: 

\subsubsection*{Smaller Language Model Viability As Reasoning Models}

\begin{itemize}
\item The extent to which RC questions can be answered by smaller Language Models without reference to one or more memorised training samples has not previously been documented.
\item How well smaller Language Models can perform the reasoning function in the unseen setting, and how performance can be improved has not been comprehensively studied. 
\item Few studies quantify the LLM performance gap to smaller Language Models when both are considered in similar unseen settings.
\end{itemize}

\subsubsection*{Knowledge Retrieval Limitations}

\begin{itemize}
\item Even the most comprehensive set of knowledge sources is unlikely to yield sufficient information to enable answering \textit{any} question deductively. This could be due to any combination of (1) incompleteness of the knowledge source, (2) incompleteness of the question specification, (3) sub-optimality in the retrieval algorithm, or (4) information retrieved being false. It is therefore desirable to consider the situation where information retrieved is partially evidential, contains irrelevant distractions, or false information. We evaluate novel mitigations for these challenges.
\item Research on improving performance in dense retrieval from textual corpora where the retrieval components are not fine-tuned on the same datasets as the target questions is limited (exceptions and alternatives to our approach in this regard are discussed in Section \ref{sec:relwork:retrieval}).
\end{itemize}

\subsubsection*{Knowledge Source Strengths and Weaknesses} 
\begin{itemize}
\item As we discuss in Section \ref{sec:relwork:knowledge_aug_from_llm}, a number of studies consider LLMs as knowledge sources, but these generally assume that the LLM is the single, or primary source. Perhaps because of this assumption there has not been much focus on quantifying the detailed strengths or weaknesses of LLMs as knowledge sources in contrast to other possible sources of contextual information. 
\item Conversely, approaches focusing on retrieval from textual corpora tend to benchmark themselves against LLMs in a closed book setting where the LLM is the Reasoning Model as well as the knowledge source. This has the effect of conflating LLM reasoning ability with LLM viability as a knowledge source. We offer an evaluation in a setting where these are disentangled.
\item Few other studies have considered approaches to \textit{combining} knowledge from disparate sources in constrained settings. Section \ref{sec:relwork:multiple_knowledge_sources} discusses those studies that we have been able to identify.
\end{itemize}

\section{Contributions}
\label{sec:intro:contributions}

In the setting discussed above, we address our research questions and make the following contributions to the research community:
\begin{enumerate}
    \item We demonstrate that a smaller Language Model is capable of performance beyond simple memorisation in deriving correct answers to challenging compositional questions. To achieve this we propose a method of identifying overlap between evaluation and training samples based upon semantic similarity of input and output tokens. We utilise this approach in conjunction with a technique to intervene with additional training datasets to create a Reasoning Model versus a baseline Reasoning Model with no intervention. Our approach enables us to mitigate effects of pretraining on results and to avoid comparing disparate populations of evaluation subsets as some prior studies have done. After demonstrating the effectiveness of our methods in identifying both memorisable, and unmemorisable samples we are able to show that improved performance on unmemorisable samples is not attributable to the effect of memorisation.
    \item We offer what is to our knowledge the most comprehensive set of baselines evaluating smaller Language Model zero-shot reasoning abilities versus LLM and other approaches published to date. Here our baseline (\textit{Base}) is a multitask-trained Reasoning Model that is trained in two stages on a large number of tasks, both existing and those that we develop.
    \item We propose the ``Iterator'', a dense retrieval, reranking and evidence set scoring system that aims to identify the relevant $n$ documents necessary to answer $n$-hop questions, where $n$ is arbitrary but we use $n=4$. 
     \item We use the Iterator against a corpus of English Wikipedia paragraphs both to develop contexts for unseen evaluation questions and to develop retrieval-augmented training datasets (\textit{RATD}) which are added to the existing \textit{Base} training regime in training the \textit{Base+RATD} model. \textit{RATD} datasets are intended to impart diverse reasoning strategies, such as an ability to identify and weigh partially evidential facts in long, noisy contexts. We show that when used in conjunction with our retrieval-augmented evaluation samples the \textit{Base+RATD} model significantly outperforms the \textit{Base} model on the established baselines. 
    \item We evaluate methods for combining information from two knowledge sources to develop contexts that are more helpful in answering questions. The first knowledge source is the above Iterator with Wikipedia while the second involves rationale generation from larger Language Models that are optimised to run locally in a resource-constrained environment. We propose ``Rationale Ranking'' (\textit{RR}), a method that both selects context components by relevance, and filters components that may be false. This is accomplished by training a Rationale Ranking model to score LLM-generated rationales and Iterator-generated contexts for \textit{truthfulness} in addition to the more common practice of quantifying relevance. A number of strategies are then evaluated for using the resulting scores to develop contexts that combine information from both knowledge sources. We show that the \textit{RR} method significantly outperforms the earlier \textit{Base+RATD} baselines. We also show that models trained using the earlier \textit{RATD} training method are able to generalise sufficiently such that they can successfully utilise combined contexts both in isolation from, and in conjunction with, \textit{RR} scoring. 
    \item We show that smaller Language Models trained for reasoning can manifest comparable or stronger performance on unseen questions to LLMs, when provided with the same knowledge to reason over that the LLM is capable of generating for itself.
    \item We present evidence to illustrate the respective strengths and weaknesses of LLMs and $n$-hop retrieval from a Wikipedia corpus as knowledge sources. The LLM tends to offer better performance when considering questions requiring commonsense knowledge (e.g. ``I’m crossing the river, my feet are wet but my body is dry, where am I?''). Retrieval from the Wikipedia corpus tends to be better at extracting knowledge necessary to answer $n$-hop factual questions where $n$ is higher than two (e.g. ``The Rhine forms a border between Aschenbrödel's composer's country and another country where women got the vote when?''). Moreover, we show that combining information from these sources significantly improves the average performance over evaluation datasets versus using a single source, and on individual evaluation datasets the combined context performance is often beyond what either knowledge source in isolation can deliver.
\end{enumerate}

Portions of this thesis have been published in a peer-reviewed international journal. In particular, our \textit{RATD} paper was accepted by Transactions on Machine Learning Research (TMLR) in August 2023 \citep{Hartill2023-pf}. Another paper of which portions are also contained in this thesis has been submitted to a well-regarded venue for peer review and is awaiting review completion.

\section{Thesis Overview}
\label{sec:intro:overview}
The remainder of this work is organized in the following chapters. \\

\noindent \textbf{Chapter \ref{sec:prelim}} provides preliminary explanations relevant to discussion in the following chapters, specifically the models we use and the unseen evaluation datasets we choose. \\

\noindent \textbf{Chapter \ref{sec:relwork}} reviews related research on the various topics that we utilise or extend in our research. We highlight the differences and similarities to our problem formulation. \\ 

\noindent \textbf{Chapter \ref{sec:memorise}} proposes a set of methods for determining whether a smaller Language Model is capable of reasoning over a provided question and context to an answer or whether it is only capable of providing a memorised answer from a similar training input. \\

\noindent \textbf{Chapter \ref{sec:ratd}} introduces a set of baselines for performance on challenging unseen compositional questions, comparing our approach of augmenting questions with a retrieved context using the Iterator against LLM and other approaches. We then discuss our method for improving performance via the addition of \textit{RATD} datasets to the training regime of our Reasoning Model and demonstrate that this significantly improves performance when combined with our retrieval method. \\

\noindent \textbf{Chapter \ref{sec:combine}} presents a set of methods for combining the retrieval knowledge source developed in the prior chapter with a second knowledge source consisting of rationales generated by larger Language Models. Here we show that further significant improvement against the baselines are possible and explore the strengths and weaknesses of each knowledge source with respect to the different types of questions encapsulated in each of our baselines. \\

\noindent \textbf{Chapter \ref{sec:conclusion}} concludes the thesis. Here, we discuss limitations and potentially fruitful avenues to be explored in future research. 

%% file: Contents/prelim.tex
\chapter{Preliminaries}
\label{sec:prelim}

The purpose of this chapter is to provide necessary definitions and background explanations relevant to the thesis. For the interested reader, Section \ref{sec:prelim:qa_history} provides a very brief history of computational approaches to answering questions. Since it does not contain novel ideas, it may be skipped. Section \ref{sec:prelim:llm_qa} provides summary background on Language Models and introduces nomenclature used later in this thesis. Finally, to avoid duplication, Section \ref{sec:prelim:unseen_eval_datasets} provides a description of each dataset we use in evaluation as different subsets of these are utilised in Chapters \ref{sec:memorise}, \ref{sec:ratd} and \ref{sec:combine}. Since we reuse or develop a large number of training datasets, the reader is referred to Chapter \ref{sec:ratd} for the Reasoning Model training process, and to Appendix \ref{sec:app_trainingdatasets} for further details on the individual training datasets.

\section{Computational Approaches to Question-Answering}
\label{sec:prelim:qa_history}

Excepting the recent trend towards using LLMs to answer questions directly using knowledge encoded in the model parameters, computational approaches to the question-answering challenge have relied upon external sources of knowledge. The earliest question answering system was BASEBALL \citep{Green1961-gh} which parsed a question into a structured representation which was then used to iteratively query a structured database. Another very early system is described in \citet{Simmons1964-fb}. It used content words extracted from each question to query an index of such terms and retrieve sentences that could be relevant to answering the question. The question and each sentence were then parsed using a dependency parser and sentences were scored with respect to the similarity of structure to the question parse. The highest scoring sentence was selected as most likely to contain the answer.

These two studies are representative of the two main historical themes in question-answering: Semantic parsing methods such as BASEBALL convert a question into a structured representation capable of being used as an exact query against a database to return an answer. Information Retrieval-based methods use some (not necessarily structured) representation of the question to retrieve a set of candidate documents, and then as in our case, use diverse RC mechanisms to extract or compute an answer from them \citep{Bordes2014-np, Jurafsky2023-oh}. 

Explorations of classical methods for RC Mechanisms where the context has been provided rather than retrieved can be found in \citet{Hirschman1999-nq, Riloff2000-uo}. These both rely on lexical overlap between question and context sentences. \citet{Ng2000-bx} claims to be the first machine learning method that is competitive for RC. They use a logistic regression model to score each question-context sentence pair where each pair is represented as a vector of 20 specifically constructed features such as a count of the number of words in common between the question and the sentence. 

In 1999 the Text REtrieval Conference (TREC) question answering track was launched with a goal of bringing together work being performed on Information Retrieval with work being done on RC \citep{Voorhees2001-qs}. Falcon \citep{Harabagiu2000-zw} is one such resulting project encompassing both of these aspects. 

More recently  \citet{Bordes2014-np, Bordes2014-nf} use neural models to embed bag-of-words representations of the question and subgraphs from the Freebase knowledge graph into a vector space such that the dot product of the resulting vector representations are higher where the subgraph contains the answer. Since that time, many different approaches to question-answering involving neural models have been studied. Prominent amongst these are approaches utilising Language Models, discussed in the next section, and approaches using graph neural networks \citep{Kipf2017-rr}. In the latter, a Language Model is typically used to create contextualised vector representations of the question and retrieved (or provided) contextual information. A graph is then constructed over both with novelty being introduced in the specification of nodes and edges. These representations are then passed through a graph neural network for further refinement. The final representations are subsequently used as input into further neural models for performing tasks such as answering the question and predicting which sentences in the retrieved context are relevant \citep{Fang2020-ba}.

\section{Language Models}
\label{sec:prelim:llm_qa}

%https://en.wikipedia.org/wiki/Language_model
%https://en.wikipedia.org/wiki/Word_n-gram_language_model
%\citep{Jurafsky2009-uz} 2nd ed
%\citep{Jurafsky2023-oh} 3rd ed: history of N-gram models: see paperpile annotations

Language Models estimate a probability function $P$ for a word, or token, in a sequence of such tokens \citep{Manning1999-tq}. Given a sequence of $s$ words $w_1,...,w_s$ denoted as $w_1^s$, the task has been formalised as learning the joint probability of the sequence from the product of the conditional probability of each word conditioned on the subsequence preceding it: 

\begin{equation}
\label{eq_prelim_lm1}
P(w_1^s) = \prod_{i=1}^s P(w_i|w_{1}^{i-1})
\end{equation}

%sim(e_{i}, t_{j}) = \frac{csim(e^{q}_{i},  t^{q}_{j}) + csim(e^{a}_{i},  t^{a}_{j})}{2.0} * 100.0    
% \sum_{i=0}^k c_i f_i(x)

According to \citep{Jurafsky2023-oh}, the mathematics of a tractible approximation of this was first formalised by Markov \citep{Markov1913-po}. Such $n$-gram models restrict the historical context considered in estimating the probability of the $i$th word to $n-1$ words by substituting the $P(w_i|w_{1}^{i-1})$ term in Equation \ref{eq_prelim_lm1} with $P(w_i|w_{i-n+1}^{i-1})$ where $n$ is typically one (substituting $P(w_i)$ for $P(w_i|w_{i-n+1}^{i-1})$), two (bigrams) or three (trigrams). The conditional probability for each $n$-gram is estimated based on a count of the number of occurrences of it in the corpus.

%\begin{equation}
%\label{eq_prelim_lm2}
%P(w_i|w_{1}^{i-1}) \approx P(w_i|w_{i-n+1}^{i-1})
%\end{equation}

In 2000, \citep{Bengio2000-vv} proposed a neural version of a Language Model where the probability distribution over possible next words from an input sequence is estimated by a feed-forward neural network. Each word in the vocabulary was represented by a dense vector $C(i) \in \mathbb{R}^d$ in which features are learned during training. The vector was stored in a matrix and accessed via the simple strategy of assigning each word in the vocabulary an index number. This is readily identifiable with the embedding tables used as the first layer in modern neural Language Models. In 2013 \citet{Mikolov2013-gy,Mikolov2013-gj} improved upon the utility of such word embeddings by proposing the Continuous-Bag-Of-Words (CBOW) model (where the embedding parameters are learned from predicting the current word from both prior and future words), and the Skip-gram model (where the training objective was to predict prior and future words from the current word). Embeddings created with models such as these and similar were commonly used as input representations in the next generation of neural Language Models that were built using recurrent neural networks (RNNs).

In 2014 \citet{Sutskever2014-mh} proposed a sequence-to-sequence Language Model built for the task of neural machine translation (NMT). It was built using the LSTM \citep{Hochreiter1997-et} version of a RNN and featured an encoder-decoder architecture where at each timestep up to a maximum input sequence length $t$, the embedding for a word from the input sequence $q:\{x_1^t\}$ is input into the encoder, which outputs a hidden representation $\textbf{h} \in \mathbb{R}^d$ where $d$ is the dimensionality of each input embedding as well as the hidden state. During training, the decoder takes the final $\textbf{h}$ as input along with the desired translation (or answer in our question-answering case) $a:\{y_1^m\}$. As with \citet{Bengio2000-vv} the decoder is trained to estimate the probability distribution over possible next words. This is applied autoregressively to generate a word per iteration: 

\begin{equation}
\label{eq_prelim_enc_dec_lm}
P(a|q) = \prod_{i=1}^m P(y_i|\textbf{h}, y_{1}^{m-1})
\end{equation}

Extending the RNN architecture, \citet{Bahdanau2015-mj} proposed an attention mechanism that uses a softmax function to produce a weighting over the sequence of all hidden states $\textbf{H} \in \mathbb{R}^{d \times t}$ produced by the encoder with the aim of weighting the most relevant parts of the corresponding input representations higher than others. This was shown to substantially improve performance on NMT tasks, and subsequently on other tasks such as question-answering as well. Adding the attention enhancement results in an update to the probability estimation function:

\begin{equation}
\label{eq_prelim_enc_dec_lm_H}
P(a|q) = \prod_{i=1}^m P(y_i|\textbf{H}, y_{1}^{m-1})
\end{equation}

In the question-answering domain, \citet{Iyyer2014-ws} and \citet{Hermann2015-lt} applied RNN architectures to RC tasks. \citet{Chen2017-gw} also used RNN models but here information retrieval was used to augment each question $q\textsubscript{i}$ with a retrieved context $c\textsubscript{i}$ where $i$ denotes the $i$th sample in a dataset. For brevity, throughout this thesis, we will denote input into a model using angle brackets e.g. in the \citet{Chen2017-gw} case the encoder input would be 
$\langle q\textsubscript{i},c\textsubscript{i} \rangle$, the decoder input would be $\langle \textbf{H}\textsubscript{i},a\textsubscript{i} \rangle$ and we will omit the batch dimension for readability. 

\citet{Vaswani2017-tr} proposed the first Transformer model,  which demonstrated improved performance on NMT tasks. Similar to \citep{Sutskever2014-mh}, this was an encoder-decoder model that estimates the probability function as per Equation \ref{eq_prelim_enc_dec_lm_H}. However the model differs greatly in that each of the encoder and decoder components primarily consist of alternating layers of self-attention and feed-forward layers. Self-attention relates each position in a single sequence to each other. \citet{Vaswani2017-tr} formalised this in the well-known equation: 

\begin{equation}
\label{eq_prelim_dot_product_attn}
Attention(\textbf{Q},\textbf{K},\textbf{V}) = softmax(\frac{\textbf{QK}^\top}{\sqrt{d_k}})\textbf{V}
\end{equation}

Here each input embedding is linearly projected onto query and key vectors $q, k \in \mathbb{R}^{d_k}$ and a value vector $v \in \mathbb{R}^{d_v}$. These are packed into matrices $\textbf{Q}$, $\textbf{K}$ and $\textbf{V}$. $\frac{1}{\sqrt{d_k}}$ is used as a scaling constant. Simplifying for brevity by ignoring positional encoding, multiple attention heads, layer normalisation and residual connections, the resulting weighted output is input into the subsequent feed forward layer. In the encoder, the process repeats until the final feed forward layer outputs $\textbf{H\textsubscript{i}} \in \mathbb{R}^{d \times t}$.

In 2019 \citet{Devlin2019-ox} proposed BERT, which is an implementation of the encoder component of the original Transformer. This paper introduced the masked Language Modelling (MLM) pretraining task in which the next-word modelling task introduced in Equation \ref{eq_prelim_lm1} is replaced with a bi-directional cloze-style objective \citep{Taylor1953-eu} reminiscent of that in the \citet{Mikolov2013-gy} CBOW model. In the MLM version of the cloze objective, tokens in the input sequence are randomly masked and the model is able to consider both prior and future tokens in estimating the probability distribution over possible tokens that each masked token could be. In this thesis we utilise later variants of BERT, namely RoBERTa \citep{Liu2019-ru} and ELECTRA \citep{Clark2020-vh} as described in Chapters \ref{sec:ratd} and \ref{sec:combine}.

Several variations of the MLM objective have seen wide adoption in encoder-decoder Transformer Language Models. Of particular note, \citet{Raffel2020-pm} evaluate a number of MLM styles and finally propose T5, a family of models that are pretrained using the version of MLM where the objective is to predict variable-length spans of text that have each been replaced by a single masking token. Similar to GPT \citep{Radford2018-kr}, described below, they then perform further training using supervised objectives over a variety of NLP tasks, and show that the resulting model has strong performance over all of them. At about the same time \citet{Lewis2020-gt} proposed BART, a similar model to T5 except that here the MLM pretraining objective was to predict the entire input sequence with all mask tokens substituted with the correct text. We use BART as our main Reasoning Model throughout this thesis. One difference to the original is that in our work, where we include a MLM task, we substitute the T5-style objective of predicting the unmasked answer spans in preference to the original BART objective of predicting the entire input sequence as it is less computationally intensive. 

Another line of Transformer model evolution has been the emergence of decoder-only Transformer Language Models. Unlike the encoder-decoder variants, these generally estimate the probability function using the original next-word objective similar to Equation \ref{eq_prelim_lm1}. GPT \citep{Radford2018-kr} was the first of these. In this study they showed that pretraining on a large corpus using the next-word objective followed by task-specific finetuning was effective in producing strong performance on individual tasks. A subsequent model, GPT2 \citep{Radford2019-ub}, was the first to show that a sufficiently large Language Model (1.5 billion trainable parameters) trained on a large corpus could become proficient on evaluation tasks in a zero-shot (unseen) setting. The GPT3 study \citep{Brown2020-rl} showed further improvement was possible by hugely scaling the model size to 175 billion parameters along with increasing the pretraining corpus size. This paper also introduced the idea of few-shot prompting where several exemplars of the problem to be solved along with the query are provided to the model as a prompt. In Chapter \ref{sec:combine} we utilise two such decoder-only LLMs, BLOOM \citep{BigScience_Workshop2022-dh} and StableVicuna \citep{Stability-AI2023-mp} in a resource constrained setting and with a focus upon their utility as knowledge sources.

\section{Unseen Evaluation Datasets}
\label{sec:prelim:unseen_eval_datasets}

For our experiments in Chapters \ref{sec:ratd} and \ref{sec:combine}, we focus our study on a set of unseen evaluation datasets that meet the following criteria: (1) Datasets collectively involve diverse textual and numerical reasoning strategies. (2) Questions are generally readily answerable by humans with access to a web browser and without specialised knowledge. (3) Questions tend to be compositional as per our definition in the Introduction. (4) Relevant comparison with prior work exists. 

Each evaluation dataset consists of a single split from the original dataset. This is typically the split most commonly used by others in published results. The particular split used is noted below for each dataset.  

Our experiments often involve augmenting the question component of each evaluation sample with contexts sourced by different means. This means that we must distinguish a number of different versions of each dataset. Therefore, in Chapter \ref{sec:ratd} we denote dataset variants that have been augmented via retrieval using our Iterator system as ``Dataset\textsubscript{R}'', and those with a gold context, ``Dataset\textsubscript{G}'', or similar. In Chapter \ref{sec:combine} we report results over the set of evaluation datasets with various context types in a single table. Hence for readability in that chapter we simplify the nomenclature to denote a set of datasets augmented with our retrieval as ``Iterator only'' in preference to the individual ``Dataset\textsubscript{R}'' format. We similarly denote datasets augmented with rationales generated by a LLM as ``Rationale only'', and those with contexts created by combining both knowledge sources as ``Rationale + Iterator''. We use the ``Dataset\textsubscript{R}'' nomenclature below when describing Iterator augmentation. Except for noting that corresponding ``Rationale Only'' and ``Rationale + Iterator'' variants are created for each of the datasets, we omit further mention of them in this section and refer the reader to Chapter \ref{sec:combine} for details of their construction.

All versions of our evaluation (and training) datasets are accessable at \href{https://github.com/timhartill/unseen_questions}{github.com/timhartill/unseen\_questions}.  \\

\noindent \textbf{StrategyQA} \citep{Geva2021-sl}, hereafter SQA, contains binary-labeled commonsense samples requiring a diversity of $n$-hop reasoning strategies (on average samples require content from 2.33 separate paragraphs to answer when considering retrieval from Wikipedia i.e. $n=2.33$). The form of questions is generally implicit, meaning they do not leak information as to how they could be decomposed (e.g. ``Did Aristotle use a laptop?'' versus ``Was Aristotle alive at the time that laptops were invented?''). Many samples involve reasoning to a plausible rather than an entailed conclusion even where gold paragraphs are provided \citep{Liang2022-lf} e.g. ``Is greed the most prevalent of the Seven Deadly Sins?''. 
To facilitate comparison with other zero-shot approaches we use the full training set for evaluation as per BIG-bench  \citep{Srivastava2022-kp} (denoted SQA for question-only and SQA\textsubscript{R} for question plus our retrieval). We also report results with two forms of gold context; using the provided summary notes which have a short paragraph, rationale-like form (SQA\textsubscript{GF}), and using the full paragraphs from each of three individual annotators (SQA\textsubscript{GP}) - for brevity we report the mean score over the three gold paragraph sets. \\

\noindent \textbf{CommonsenseQA} \citep{Talmor2019-rm} (CSQA) is a 5-way multi-choice (MC) dataset of commonsense questions derived from Conceptnet \citep{Speer2017-ll}. The task is to choose the best option of which more than one may sometimes be plausible, hence it may be necessary to consider knowledge related to each option before answering. Many of the questions involve commonsense knowledge that is unlikely to be retrievable from a generic corpus (``Where on a river can you hold a cup upright to catch water on a sunny day''). However retrieving specific related examples such as ``At the river, I filled my cup at a waterfall'' may sometimes be possible \citep{Piktus2021-lu}. CSQA augmented with our retrieval is denoted CSQA\textsubscript{R}. We report all results against the development split as is common practice. \\

\noindent \textbf{DROP} \citep{Dua2019-td} is a RC dataset wherein answering each question requires simple numerical or temporal reasoning. Questions only make sense in conjunction with the provided gold paragraph so we do not perform retrieval. Answers may be numbers, dates or text spans. Answers are often abstractive e.g. ``How many field goals were scored in the first quarter? ...The Lions scored first...with a 23-yard field goal...The Buccaneers tied it up with a 38-yard field goal...then took the lead...The Lions responded with a 28-yard field goal...'' The answer is 3 which isn’t explicit in the context. We use the full development split in all experiments except for those in Chapter \ref{sec:memorise} where preprocessing is performed as described in that chapter. \\

\noindent \textbf{IIRC} \citep{Ferguson2020-hv} contains questions where an initial paragraph is given and answers depend upon this plus additional paragraphs that must be retrieved  ($1 \leq n \leq 4+$). Each sample is provided with links to all supporting documents, and prior work leverages these to restrict the number of documents to be retrieved from. We instead use our Iterator to augment samples from the full Wikipedia corpus using the concatenation of question and initial paragraph as the query, without reference to the given links (IIRC\textsubscript{R}). We also report comparison against an oracle context (IIRC\textsubscript{G}) that we construct from the initial paragraph concatenated with the linked supporting documents. Answers may be numbers, binary, text spans or labeled unanswerable. For IIRC\textsubscript{G} unanswerable samples, we construct contexts using the initial paragraph fragment plus 1-2 random distractor paragraphs. We report all results against the test split. \\

\noindent \textbf{ARC-DA} \citep{Bhakthavatsalam2021-fq} is a question-only subset of ARC \citep{Clark2018-gy} where questions have been re-worded to make sense in an open domain context. The Worldtree database \citep{Xie2020-xb} provides explanatory fact sets for ARC samples which average six facts per sample. The original multichoice versions of ARC are part of our training regime, hence compositionality is doubtful and samples are only partially unseen in the sense that the question format is different (and we use the test split). Nonetheless we report results in the interests of exploring diversity. We experiment with Iterator-augmented (ARCDA\textsubscript{R}) versions as well as with a gold context that we construct from Worldtree  (ARCDA\textsubscript{G}) by concatenating the individual fact strings. \\

\noindent \textbf{Musique} \citep{Trivedi2022-mv} is an $n$-hop dataset ($n \leq 4$) constructed by combining single-hop questions from existing datasets including SQuAD \citep{Rajpurkar2016-fs} which is also part of our training regime. Moreover we utilise the training split of Musique in both our retriever and Reasoning Model training. However the provided development split has been constructed such that for all samples no single hop question, answer, or associated paragraph is common to the corresponding element of any training sample. Therefore we construct a new development set from the training set and experiment with the original Musique development split as ``partially seen'', this time where the form of questions is ``seen'' but the exact questions are not. Prior work generally uses specialised retrieval for Musique where selection is from the set of gold and distractor paragraphs provided for each sample. We experiment with our retrieval (Musique\textsubscript{R}), and with a gold context constructed from the concatenation of the supplied gold paragraphs (Musique\textsubscript{G}). \\ \\

\noindent In Chapter \ref{sec:memorise} we also make use of CommonsenseQA and DROP, and additionally consider the following datasets. We use the publicly available development split for each: \\

\noindent \textbf{DROP-CS} \citep{Gardner2020-rt} contains perturbed versions of DROP Test split samples e.g. by making a minor change to the context such that the label is changed. \\

\noindent \textbf{ROPES} \citep{Lin2019-ej} is a RC dataset that requires multi-step reasoning over a situation, often involving qualitative relations such as ``higher'' or ``lower''. Questions are human-authored based on passages from Wikipedia and science textbooks. \\

\noindent \textbf{NewsQA} \citep{Trischler2017-gu} is a RC dataset of human-authored questions about CNN articles. \\

\noindent \textbf{PIQA} \citep{Bisk2020-rl} is a two-option MC dataset covering physical commonsense questions. Samples are created by human annotators from prompts sourced from \texttt{instructibles.com}. \\

\noindent \textbf{QASC} \citep{Khot2020-sv} is an eight-option MC dataset covering human-authored science questions that require two facts to answer. Facts are sourced from a corpus derived from open web pages \citep{Clark2016-xg}. \\

%% file: Contents/relatedwork.tex
\chapter{Related Research}
\label{sec:relwork}

\section{Memorisation in Language Models}
\label{sec:relwork:memorisation}

As in our case, prior work on studying the effects of memorisation on model performance in the NLP domain has generally focused on identifying subsets of evaluation data that are either unlikely or likely to have been memorised from training data. Studies have then considered the performance of a subset in conjunction with the nature of the input samples. \citet{Lewis2021-ia} consider open-domain single-hop factual questions. By identifying test questions with answers matching training questions and then manually identifying those evaluation samples where the question is or isn't a paraphrase of a training question, they show that smaller Language Models (such as the BART model \citep{Lewis2020-gt} we also use) exhibit low performance on samples that don't have a match in the training set. Our Chapter \ref{sec:memorise} can be considered as an extension of this work in the area of RC questions that require reasoning over a context to answer. We show that in contrast to their findings on factual questions, a BART model is capable of \textit{improved} performance for RC samples without a memorisable match in the training set. \citet{Elangovan2021-gs} consider train-test overlap on different NLP tasks to ours. To evaluate similarity they utilise cosine similarity between sparse bag-of-words vectors constructed for each test and train sample. Similar to our study, a recent work, \citet{Kambhatla2023-fc}, considers cosine similarity over sentence embedding vectors as the similarity measure, although they only consider the input tokens whereas we consider both input and output. Additionally this study differs from our purpose in that it is focused on identifying dataset contamination between test and train splits within the same dataset, and in other methodological aspects such as controlling for the effects of pretraining as discussed further in Chapter \ref{sec:memorise}.  

The effect of evaluation dataset contamination in the pretraining datasets of large Language Models (LLMs) has been reported in a number of studies \citep{Brown2020-rl, Sanh2021-na, Wei2021-go, Du2022-ld, Chowdhery2022-yw}. These generally automate the process of contamination discovery by considering \textit{n}-gram overlap between evaluation datasets and pretraining data. A filtered, clean version of each evaluation dataset is sometimes then constructed and performance is compared to that of the full dataset. Generally these studies find that even where an evaluation dataset is found to heavily overlap with pretraining data, the performance gap between clean and full versions is small and each clean version may either slightly underperform or slightly overperform the full version. Although we are not disagreeing with the overall findings, one criticism of this approach is that \textit{n}-gram overlap can only detect test-train overlap where the overlap is an exact match of contiguous tokens, while paraphrases or overlaps between discontinuous tokens that otherwise overlap highly will not be detected. %On the other hand, as \citet{Chowdhery2022-yw} note, the degree of memorisation may be overstated in some of these studies such as where a sample's context is drawn from the web but the question itself is novel.

Also focusing on memorisability in pretraining data in the situation where the pretraining corpus is available, \citet{Carlini2023-fb} evaluate memorisation by prompting a model with a particular sequence and ascribing memorisation if the model continuation is an exact match to the ground truth continuation of that sequence. They show that the degree of memorisation increases both with the size of the model and with the number of duplicates of a sequence in the pretraining data. \citet{Lee2022-zm} show that training on de-duplicated pretraining data results in memorised text being generated ten times less frequently. \citet{Kandpal2023-md} show that single-hop factual question answering performance is correlated with the number of documents containing the question and answer entities seen in pretraining. In the domain of numerical reasoning, \citet{Razeghi2022-ue} show that numerical term frequency in the pretraining corpus also correlates with accuracy. The study goes on to remove evaluation samples that are likely to have been memorized i.e. those where the input terms and the answer co-occur in a pretraining document.  It was then found that the performance of the remaining unmemorisable samples continues to correlate with the frequency of the input terms in the pretraining corpus, suggesting that the performance improvement is not solely due to memorisation. 

As a reminder that \textit{spurious} memorisation can lead to lower results in downstream evaluation as well as inflating results, our own results in Chapter \ref{sec:ratd} show that removing near-duplicate Musique \citep{Trivedi2022-mv} training samples from a BART model training regime resulted in improved downstream performance where evaluation samples had input token overlap with the duplicated training samples but had different labels.  

Outside of the NLP domain, a number of studies have challenged the historical assumption that an ability to memorise the training set and an ability to generalise are mutually exclusive \citep{Zhang2021-ll}. In considering overparameterised models (those with more trainable parameters than samples they are trained on), \citet{Zhang2017-mb} found that such models are capable of perfectly memorising a training set with randomly assigned labels, without learning any ability to generalise. Models trained on the same training data except with correct labels assigned are of course able to generalise successfully to test samples. By varying the degree of randomness in assigning labels to training samples between these two extremes the authors found a correlation between generalisation error and the amount of label noise, showing that overparameterised neural networks are capable of both capturing the extant signal in the data, while at the same time memorising the noisy part. \citet{Feldman2019-fq} proposes that memorisation in long-tail distributions (i.e. the common case where classes consisting of small numbers of samples collectively comprise a significant fraction of the distribution) is actually necessary in minimising generalisation error, and empirically demonstrates this in \citet{Feldman2020-gp}. The focus of our study differs from these in that we are primarily interested in evaluating whether a model in our setting can exhibit an ability to generalise in the absence of an opportunity to memorise.

With a more distant connection with our work, \citet{Hacohen2020-rw} show that various neural models learn similar classification functions at particular stages of training. Exploring this idea in the NLP domain, \citet{Choshen2022-ex} study the order that linguistic phenomena are learned over the course of training and find that neural Language Models with differing architecture and training data tend to acquire particular linguistic abilities in a similar order. Future work might consider the relationship, if any, between such order of learning and the acquisition of skills involving memorisation versus those relating to more abstract RC skills such as logical operations, multi-step reasoning and so forth.

\section{Retrieval from Textual Corpora}
\label{sec:relwork:retrieval}

As discussed in Section \ref{sec:prelim:llm_qa}, \citet{Chen2017-gw} first used sparse retrieval, namely TF-IDF \citep{Sparck_Jones1972-dm}, against Wikipedia in the context of open domain question-answering. In dense retrieval, query and corpus documents are embedded into the same vector space and retrieval is typically performed through maximum inner product search (MIPS) over the resulting dense vectors. Several such approaches e.g. \citet{Karpukhin2020-wa} focus on retrieving the single most relevant document sufficient for answering a single-hop query. \citet{Lewis2020-is} combine single-hop dense retrieval with a generative Transformer using end-to-end backpropagation, a combination that they term retrieval-augmented generation (RAG). \citet{Xiong2021-ex} introduce multi-hop dense retrieval (MDR), to retrieve \textit{multiple} documents necessary to answer a complex multi-hop question. They focus on the two-hop situation where a maximum of two documents are sufficient to answer a question. In this situation training samples are input to a shared question and document encoder as: (1) Input $\langle q\textsubscript{i}\rangle$ with an objective of minimizing distance to the vector representing $d\textsubscript{i,0}$ (hereafter denoted $\langle q\textsubscript{i}\rangle \rightarrow d\textsubscript{i,0}$), where $d\textsubscript{i,t}$ is the \textit{t-th} supporting document of $q\textsubscript{i}$ to retrieve. (2) Input $\langle q\textsubscript{i}, d\textsubscript{i,0}\rangle\rightarrow d\textsubscript{i,1}$. We extend the MDR training regime and loss computation to enable retrieval of an arbitrary maximum number of documents i.e. $\langle q\textsubscript{i}, d\textsubscript{i,0}, ..., d\textsubscript{i,t}\rangle\rightarrow d\textsubscript{i,t+1}$.

\citet{Wang2018-zz} introduced the concept of a Reranker that refines retrieved results. IRRR \citep{Qi2021-sm} combined sparse retrieval and reranking into an iterative single model that can also answer multi-hop questions that have extractive answers. Baleen \citep{Khattab2021-jf}, is also iterative but uses a dense retrieval system based upon encoding a dense vector per input token. Their two-stage condenser system comprises a Reranker that scores the relevance of each sentence for each retrieved document followed by an additional module that scores relevance of each sentence from the top-scoring sentences selected over multiple documents from the first stage. It then generates a compressed context of relevant sentences, to be utilised by a separate QA Model. We take inspiration from Baleen's two-stage approach but other than using our own retriever, we differ most notably in that we introduce an Evidence Set Score into the second stage with the goal of quantifying the sufficiency of the entire set of selected sentences for answering a query, in addition to scoring the relevance of individual sentences.

Sparse retrieval offers the advantage that it can perform well in zero-shot settings where lexical overlap is sufficient to identify relevant documents. Several studies evaluate methods that improve the performance of dense retrieval models in zero-shot settings. A number of these use diverse unsupervised techniques involving creating queries and positive passages from unlabelled text e.g. \citep{Lee2019-vh, Ram2022-nf,Izacard2022-hu}. In a different approach, \citet{Chen2021-dh} trained a dense retriever to imitate a lexical-based model with good results. \citet{Thakur2021-qr} created the BEIR benchmark to further the study of retrieval in the zero-shot setting and some recent papers report results against this benchmark. We are unable to do so as some of our retriever training datasets are BEIR components, however we note as a future direction that our retriever training might benefit further from applying techniques that have been effective on BEIR.

\section{Knowledge Augmentation from LLMs}
\label{sec:relwork:knowledge_aug_from_llm}

\citet{Bosselut2019-wn} proposed COMET, a GPT-based Model \citep{Radford2018-kr} trained on triples from the ATOMIC  \citep{Sap2019-tk} and ConceptNet \citep{Speer2017-ll} knowledge graphs such that it would generate potentially novel triple completions. \citet{Shwartz2020-sn} compare augmentation methods using COMET, ConceptNet and their self-talk method where the question-answering Language Model is self-queried to produce additional information pertinent to answering the question. \citet{Liu2022-bc} generate knowledge statements from GPT-3 \citep{Brown2020-rl} conditioned on the question and use the augmented samples in separate smaller Reasoning Models. \citet{Yu2023-wv} also generate contextual information from a LLM, in this case by clustering supporting documents from dataset training splits and creating prompt exemplars from each cluster separately so that the LLM may generate diverse knowledge statements. Following the introduction of chain-of-thought (COT) prompting \citep{Wei2022-lz}, a number of recent papers \citep{Magister2023-uf, Li2023-ss, Hsieh2023-my, Wu2023-dg, Shridhar2023-qb} use this prompting style to distill training sets of rationale-augmented samples from internet-accessable LLMs such as GPT-3 or Palm \citep{Chowdhery2022-yw}, which are then typically used to train much smaller models in task-specific finetuned settings sometimes such that the label and the rationale are output to avoid the issue of having to generate a rationale from the LLM at test time. We note that our usage of LLM-generated rationales is rather different from these in that we assume a locally-accessable LLM  (with lower resource requirements) at test time and do not incorporate LLM-generated rationales in our Reasoning Model training.  We do however incorporate negative rationales generated by a LLM in our \textit{RR} Model training regime as discussed in Section \ref{sec:combine_rr_training}.
%\citep{Huang2022-gr} Large Language Models can self-improve. Train PaLM on rationales from palm...improves a large LM vs a smaller LM

\section{Multiple Knowledge Sources}
\label{sec:relwork:multiple_knowledge_sources}

Retrieval has been successfully used as a method for querying knowledge sources other than textual corpora. For example this approach has been used to obtain information from knowledge graphs by embedding the constituent triples as the document vectors in addition to, or instead of, standard text. \citet{Yu2022-be} augment commonsense questions with retrieved information from a commonsense-focused corpus consisting of information sourced from knowledge graphs, commonsense datasets and other textual sources. Perhaps most similar in spirit to our work \citet{Pan2023-sz} consider knowledge graphs, Wikipedia data, a dictionary, and others, as separate knowledge sources, each queried using dense retrieval. In contrast to our approach of considering various methods for combining information, they train a model to select the single most relevant source for augmenting each input sample. This is analogous to our ``Max Score'' method described in Section \ref{sec:context_combination_strategies}. Like us they train a smaller Reasoning Model with disparate training and evaluation datasets, although unfortunately their evaluation datasets differ from ours. Also in a similar direction to our ``Max Score'' method, \citet{Si2023-kz} route a query to four expert LLMs and select the single most likely answer using a smaller classifier trained for that purpose. \citet{Sun2018-gx} combine information from a textual corpus and a knowledge graph into a question-specific subgraph from which an answer is extracted. In a finetuned setting, \citet{ijcai2022p383} also consider multiple knowledge sources. They use an entity linking method to query ConceptNet and sparse retrieval over a dictionary and a set of commonsense datasets. The results are always concatenated which is similar to our Naïve Concatenation method (Section \ref{sec:context_combination_strategies}).

\section{Falsehood Detection}
\label{sec:relwork:falsehood_detection}

Our \textit{RR} Model, trained to score for truthfulness and relevance over instances from disparate knowledge sources, can be seen as a novel extension to a Reranking approach. However it also shares an objective with methods aiming to detect falsehood in LLM generations. Generally these methods fall into three categories. The first are methods based on the intuition that higher token log probabilities correspond to better text along a particular dimension such as truthfulness \citep{Yuan2021-dm,Fu2023-ir}. The second are factuality detection methods that evaluate LLM-generated assertions as true if they can be supported by a external reference (e.g fact retrieval from a reliable corpus). Recent studies here include \citep{Min2023-fc, Chern2023-jw}. A third category, broadly called self-checking involves prompting a LLM such as ChatGPT or GPT-4 \citep{OpenAI2023-cv} to recognize their own errors \citep{Chern2023-jw}, or refine their own outputs \citep{Chen2023-lc,Madaan2023-rm}, without recourse to external tools. In this category but with a different approach, \citet{Manakul2023-db} score the consistency between a reference statement and several stochastically sampled versions of it that may be likely to diverge more if the reference is a hallucination.

\section{Multitask Pretraining}
\label{sec:relwork:multitask_pretraining}

\citet{Raffel2020-pm} showed that when trained using self-supervised pretraining followed by supervised multitask training, a single sequence-to-sequence Transformer model without task-specific architectural modification was capable of performing well on all the diverse tasks it had been trained upon. Since then, a number of studies have shown the efficacy of supervised multitask training in facilitating generalisation in question-answering tasks \citep{Khashabi2020-gq, Sanh2021-na, Wei2021-go, Khashabi2022-bq}. Different to us, but orthogonal to our approach, many studies e.g. \citet{ Sanh2021-na, Wei2021-go, Ouyang2022-ti} make use of instruction-based tuning to facilitate generalisation. In order to focus on evaluation of differing training data regimes, we make use of a similar fixed prompting format to \citet{Khashabi2020-gq, Khashabi2022-bq} and utilise many of their converted QA datasets.  Perhaps most similar to our work, \citet{Wang2022-jm} combines multitask training over multi-choice datasets with external retrieval which they use to augment the training set. However their implementation diverges from ours in that they use sparse retrieval and then a fusion-based method similar to \citet{Izacard2021-zq} wherein multiple retrieved document vectors are used with gated cross-attention to focus on salient information. Their evaluation datasets are disjoint with ours and don't cover broader reasoning skills like numeracy, so comparison must be left to future work.

\citet{Longpre2021-on} created a synthetic dataset by substituting entity names in existing dataset contexts and updating corresponding labels to produce new unfactual but logically consistent samples. They show that training on the new dataset plus the original causes their model to rely on reasoning over the context more, and less on knowledge encoded in parameters. Recently, \citet{Li2022-dk} extended this approach to a fine-tuning framework for LLMs wherein the model is trained on relevant, irrelevant, and counterfactual but logically consistent contexts. Their approach differs from ours in that our \textit{RATD} datasets are constructed so as to encourage reasoning to a plausible conclusion whereas theirs are constructed with logical entailment in mind i.e. to ignore contexts where deductively valid reasoning is not possible in favor of knowledge stored in the LLM parameters.

\section{Numerical Literacy in Language Models}
\label{sec:relwork:num_literacy}

\citet{Yoran2022-mb}, \citet{Pi2022-um} and \citet{Geva2020-dd} all develop numeracy-focused pretraining datasets that we adapt and utilise. Generally these approaches have concentrated on finetuned settings and to our knowledge we are the first to study their performance against a diversity of unseen evaluation datasets. Recently \citet{Trivedi2022-rh} released numeracy-focused pre-training datasets constructed from ``Question Decomposition Meaning Representation'' (QDMR) representations of several existing datasets from \citet{Wolfson2020-dx}. These are structured representations of reasoning paths leading from questions to answers. They were released too late for us to include in our pretraining regime but we report  comparisons in Table \ref{ratd:tab:perf_ft_1}.

%% file: Contents/chapter1_memorise.tex
\chapter{Do Smaller Language Models Answer Contextualised Questions Through Memorisation Or Generalisation?}
\label{sec:memorise}

\section{Introduction}
\label{sec:memo:intro}

Memorisation has been described as the learning of a direct mapping between input features and particular outputs \citep{Chatterjee2018-iy, Elangovan2021-gs, Schwarzschild2021-pd, Lewis2021-ia}, in contrast with generalisation \citep{Elangovan2021-gs}, or the application of a method for deriving the output \citep{Schwarzschild2021-pd}. A number of studies have considered the impacts of memorisation from the perspective of the capacity of particular models to memorise pretraining data e.g. \citet{Carlini2023-fb, Chowdhery2022-yw} as well as through the lens of downstream evaluation dataset contamination e.g \citet{Brown2020-rl, Sanh2021-na, Wei2021-go, Du2022-ld, Chowdhery2022-yw}. A general finding has been that memorisation capacity scales with model parameter count, which implies that smaller models would suffer less from this problem. However observations from \citet{Lewis2021-ia}, as well as from our own work in Chapter \ref{sec:ratd}, on the BART model \citep{Lewis2020-gt} suggest that undetected memorisation could effect smaller Language Models sufficiently so as to be an issue in interpreting results.

\begin{figure*}[h]
\begin{center}
\includegraphics[width=\textwidth]{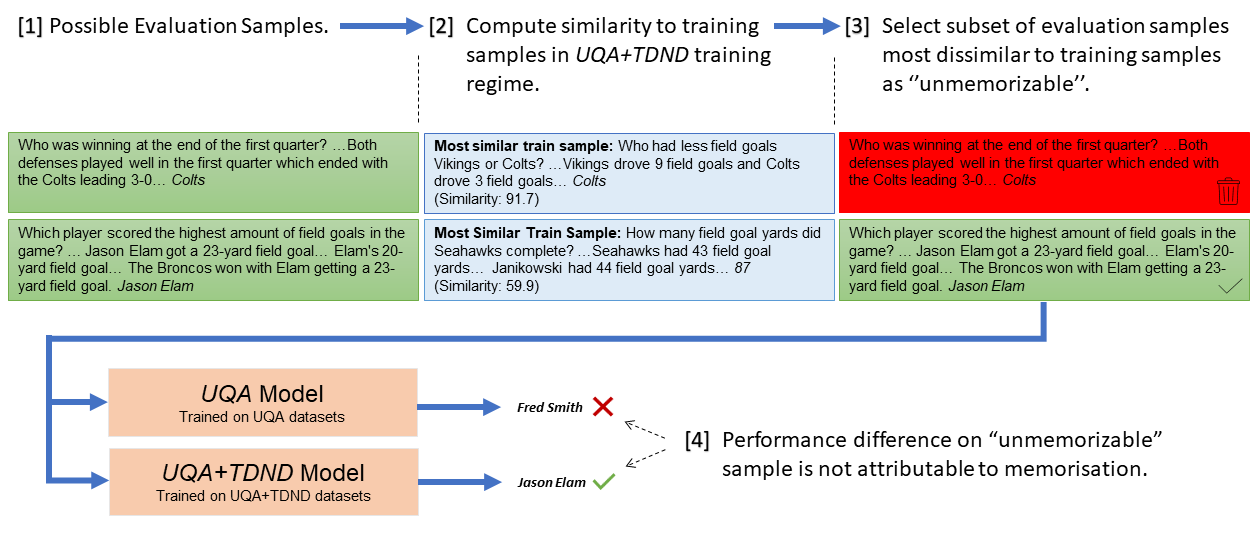}
\caption{Visualisation of key aspects of our methods. We consider two models, one trained on a set of question-answering datasets (\textit{UQA}) and the other trained on \textit{UQA} plus two additional datasets collectively referred to as \textit{TDND} (\textit{UQA+TDND}). \textit{TDND} samples are constructed so as to improve performance on some of our evaluation datasets and to be irrelevant for others. Our objective is to understand whether any improvement is attributable to memorisation or to \textit{TDND} samples imparting an improved ability to generalise. We select evaluation samples that are very unlikely to have become memorisable from our training datasets based on a semantic similarity score (Section \ref{sec:memo:sim_method}), and compare performance between the two models. Our method enables evaluating performance for each model on the \textit{same} subset of unmemorisable samples, and it does not require access to the pretraining corpus.}
\label{fig:memo:ovw_fig}
\end{center}
\end{figure*}

We consider the impact of memorisation on evaluation samples that preferably should involve reasoning from a question, over a provided context to an answer. Where the context is of a free-form nature we describe these as requiring reading comprehension (RC samples) and we denote samples where the context comprises multi-choice options as MC samples. We characterise an evaluation sample as memorisable if it is similar in terms of input and output to one or more training samples e.g. an evaluation sample consisting of the input ``What is a tool for indicating air pressure? (A) seismograph (B) barometer ...'' and label ``barometer'' is memorisable if a sample with input ``Which weather instrument measures air pressure? (A) barometer (B) rain gauge ...'' and label ``barometer'' exists in the training data. To identify memorisable evaluation samples we propose a method of scoring similarity between each evaluation and each training sample using semantic similarity as encoded in sentence embedding vectors produced by a Sentence Transformers model \citep{Reimers2019-vz}. This is discussed in more detail in Section \ref{sec:memo:sim_method}. 

The UnifiedQA project (\textit{UQA}) \citep{Khashabi2020-gq} demonstrated that it is possible to attain good performance on unseen evaluation datasets (those that have not been involved in training) after further training of a pretrained Language Model on a variety of question-answering datasets in a multitask fashion. One of the unseen RC datasets that \citet{Khashabi2020-gq} use for evaluation is DROP \citep{Dua2019-td}. Performance on DROP is rather poor in the \textit{UQA} setting. This dataset requires simple numerical literacy in order to correctly answer a question. A separate study, \citet{Geva2020-dd}, demonstrated significant performance improvement on DROP by pretraining on two synthetic datasets (collectively referred to here as \textit{TDND}) that they designed to impart simple numerical reasoning strategies. We add \textit{TDND} to the \textit{UQA} training mixture (denoted \textit{UQA+TDND}) and analyse the impact on subsets of DROP \citep{Dua2019-td}, ROPES \citep{Lin2019-ej}, and  several other unseen RC and MC datasets that are unlikely to be memorisable, even after the addition of the \textit{TDND} datasets. \\

\noindent In summary the major contributions of this chapter are:
\begin{enumerate}
\item We propose a method of identifying evaluation-train overlap based on semantic similarity of input and output tokens. 
\item We propose a method to intervene with additional training datasets versus a baseline, both to mitigate effects of pretraining on results, and to avoid the need to compare disparate populations of evaluation subsets. 
\item We demonstrate the effectiveness of our methods in identifying both memorisable, and unmemorisable samples. 
\item  We show that performance on unmemorisable subsets of DROP and ROPES is significantly improved by the addition of \textit{TDND} training datasets.
\end{enumerate}

\section{Method}
\label{sec:mem:method}

In the context of language models, \citet{Carlini2023-fb} characterise memorisation as the generation of an exact continuation of a text sequence, given the first part of the sequence as input. Several other studies (Section \ref{sec:relwork:memorisation}) test for potential memorisation (evaluation dataset contamination) as the presence of \textit{n}-gram(s) in training samples that co-occur in evaluation samples (where $n\geq8$). We take a view of potential memorisation as occurring where there is not only overlap in a \textit{contiguous} sequence of tokens but also where a \textit{discontinuous subset} of input tokens could directly produce a particular output. For example learning one or more training samples similar to  ``\textcolor{red}{\underline{Who}} had more field goals Vikings or \textcolor{red}{\underline{Colts}}? ...'' with label ``Colts'' could cause a model with evaluation input ``\textcolor{red}{\underline{Who}} was winning at the end of the first quarter? ... \textcolor{red}{\underline{Colts}} leading 3-0...'' to predict ``Colts'' without any semantic understanding of the question or the context. We develop an alternative method of evaluating evaluation-train similarity using cosine similarity of evaluation and train sample sentence embedding vectors. We find that this approach surfaces test-train overlaps where the tokens discontinuously (or contiguously) overlap  (see Section \ref{sec:memo:sim_method}).

In some prior work it has been necessary to compare disparate populations of evaluation samples in order to draw conclusions. For example \citet{Chowdhery2022-yw} note that in comparing the full version of an evaluation dataset to a filtered version consisting only of unmemorisable samples they are comparing different subsets. We address this issue by identifying evaluation samples that will not be rendered memorisable by the addition (``intervention'') of new training datasets and then using this same subset to evaluate the performance difference before and after our intervention. This approach has the added benefit that we do not need access to the pretraining corpus. A visual overview of our approach is provided in Figure \ref{fig:memo:ovw_fig}. 

Below we discuss how the training regimes for our ``before'' model (\textit{UQA}) and ``after'' model (\textit{UQA+TDND}) are constructed, our evaluation datasets, and our methods for identifying evaluation samples that are very unlikely to have become memorisable by the intervention of the additional training datasets.

\subsection{\textit{UQA} and \textit{UQA+TDND} Model Training}
\label{sec:memo:training_regime}

\begin{table}[h]
\centering
\begin{tabular}{@{}l|rr|rr@{}}
\toprule
\textbf{} & \multicolumn{2}{c|}{\textbf{UQA}} & \multicolumn{2}{c}{\textbf{UQA+TDND}} \\
\textbf{Run} & \textbf{Step} & \textbf{Dev Perf.} & \textbf{Step} & \textbf{Dev Perf.} \\ \midrule
1 & 140,000 & 65.80\% & 150,000 & 67.45\% \\
2 & 110,000 & 66.62\% & 140,000 & 68.76\% \\
3 & 140,000 & 66.13\% & 140,000 & 68.74\% \\ \bottomrule
\end{tabular}
\caption{Best model selection for three runs each of \textit{UQA} and \textit{UQA+TDND}. Step is the training step at which the best model is selected. Dev Perf is the mean accuracy over constituent development sets. The \textit{UQA+TDND} best model has usually but not always been trained for more steps than the \textit{UQA} best model.}
\label{tab:memo:best_model_selection}
\end{table}

Our main experiments evaluate the performance difference between two models; \textit{UQA} and \textit{UQA+TDND}.  Both are trained using the same hyperparameters (Appendix \ref{mem:sec:app_hyperparams}), the only differences being the respective sets of datasets used to train them. We experimented with differing combinations of hyperparameters on both training mixtures until we found a set that worked well over both.  Training for both models is performed in a multi-task manner, uniformly sampling over the training datasets. The best model from each run is selected as that with the highest mean performance over all development sets after 150,000 train steps which allows for some flexibility in tuning per training mixture as shown in Table \ref{tab:memo:best_model_selection}. We make use of a similar fixed prompting format to \citealp{Khashabi2020-gq, Khashabi2022-bq} (Appendix \ref{sec:app_inputformat}), and take as our \textit{UQA} baseline the same set of training datasets that they use. Specifically, \textit{UQA} consists of datasets of RC type; SQUAD 1.1 \citep{Rajpurkar2016-fs}, SQUAD 2 \citep{Rajpurkar2018-rt}, NarrativeQA \citep{Kocisky2018-rt}, along with MC datasets RACE \citep{Lai2017-pb}, ARC \citep{Clark2018-gy}, Regents \citep{Clark2016-xg} (``Sci-Elem'' and ``Sci-Mid'' in this Chapter) , OpenbookQA \citep{Mihaylov2018-uk}, MCTest \citep{Richardson2013-ct}, and one binary-labelled dataset, BoolQ \citep{Clark2019-vz}.

As noted, \citet{Geva2020-dd} developed two synthetic datasets designed to impart numerical reasoning ability of the sort needed to improve model performance on DROP \citep{Dua2019-td}. Of these, ``Textual Data'' (\textit{TD}) contains RC samples with similar vocabulary and involving similar reasoning skills to DROP (e.g. ``Who had the lowest number of field goal yards in total? ... Dolphins nailed 26 field goal yards and Vikings nailed 15 field goal yards...'', label ``Vikings''). The second dataset, ``Numerical Data'' (\textit{ND}) contains a large number of samples with inputs consisting of symbolic expressions (e.g ``argmin(undergrass 11952 bussu 3315)?'', label ``bussu''). \citet{Geva2020-dd} show that pretraining on \textit{TD} and \textit{ND} followed by finetuning on DROP leads to substantially higher performance. In our case, we convert the datasets (collectively \textit{TDND}) to our format; specifically  \textit{ND} is converted to our open domain format, and TD to RC format as detailed in Appendix \ref{sec:app_inputformat}. These are added to the \textit{UQA} training mixture to train our \textit{UQA+TDND} model.

Further detail on the datasets used in the training regime for both models may be found in Appendix \ref{app:memo_models}.

\subsection{Evaluation Dataset Preprocessing}
\label{sec:memo:unseen_eval_datasets}

%Several studies eg. \citet{Lewis2021-ia}, as well as our own (Table \ref{tab:qual0_indomain}), illustrate a tendency for train split samples to be duplicated or paraphrased as development/test samples. Since we wish to minimise evaluation dataset memorisability, we begin by choosing evaluation datasets that are disjoint from our training datasets (``unseen''). 

We selected evaluation datasets as described in Section \ref{sec:prelim:unseen_eval_datasets}, namely DROP, DROP-CS, ROPES, NewsQA, PIQA, CSQA and QASC, in all cases using the publically available development split.

We discovered that the DROP development split that we use here for evaluation contained over 800 exact duplicates. Because we were unsure whether duplicate samples were the result of some bias in dataset creation that could manifest itself when we select smaller ``unmemorisable'' subsets we de-duplicated all our evaluation splits and note that DROP-CS also contained a very small number of duplicates. An example for each dataset is shown in Table \ref{tab:memo:qual1_unseen_mostsim}.

\begin{table}[h]
\centering
\begin{tabular}{@{}lrrrr@{}}
\toprule
\textbf{Eval Dataset} & \textbf{All} & \textbf{\begin{tabular}[c]{@{}r@{}}Least\\ Similar\end{tabular}} & \textbf{Unmemorisable} \\ \midrule
DROP & 3277 & 867 & 652 \\
DROP-CS & 478 & 154 & 110 \\
ROPES & 1688 & 307 & 197 \\
NewsQA & 4341 & 1204 & 759 \\ 
%MMLU-M & 530 & 217 & 133 \\
PIQA & 1838 & 1354 & 588 \\
%SIQA & 1935 & 718 & 373 \\
CSQA & 1221 & 233 & 129 \\
QASC & 926 & 139 & 99 \\ \bottomrule
%QASC-IR & 926 & 95 & 72 \\
\end{tabular}
\caption{Evaluation Dataset sample counts. ``All'' is the total sample count after de-duplication and removal of samples with numeric answers. Least Similar is the subset of these with a Similarity Score of each evaluation sample to it's most similar training sample under 60.0. Unmemorisable samples are those Least Similar which also have no answer term overlap with the most similar training sample.}
\label{tab:memo:eval_dataset_counts}
\end{table}

When selecting ``unmemorisable'' subsets (see Section \ref{sec:memo:sim_method} below) we observed that samples with numeric answers were much more likely to be filtered out since many such answers tend to be commonly occurring small numbers (1, 2, 5...). To combat this bias we remove all samples with numeric answers from our DROP and DROP-CS evaluation.

The resulting sample counts are in Table \ref{tab:memo:eval_dataset_counts}. Elaboration as to how the ``Least similar'' and ``Unmemorisable'' subsets are derived follows in the next section.

\subsection{Similarity Computation Method}
\label{sec:memo:sim_method}

To evaluate similarity between evaluation and training samples, we use sentence embedding vectors produced by the ``sentence-transformers/stsb-roberta-large'' model \citep{Reimers2019-vz} from the Huggingface library \citep{Wolf2020-ro}. We quantify the ``memorisability'' of each evaluation sample from each training sample by computing a Similarity Score as:

\begin{equation}
\label{eq:memo:eq_1}
sim(e_{i}, t_{j}) = \frac{csim(e^{q}_{i},  t^{q}_{j}) + csim(e^{a}_{i},  t^{a}_{j})}{2} * 100    
\end{equation}

Here $e_i$ and $t_j$ are the embeddings for the $ith$ evaluation and $jth$ training samples, $q$ and $a$ refer to the question (including context) and answer components of each, and csim is the cosine similarity function. We consider both $q$ and $a$ equally as we are primarily interested in identifying evaluation-train pairs where a memorised answer could inflate results. Alternative formulations that consider $q$ only would also identify spuriously memorisable samples that could deflate results but this does not suit our purpose here.

We require a memorisability threshold $T$ for Similarity Scores, below which sample pairs are sufficiently dissimilar as to be unmemorisable. The choice of a value for $T$ involves a trade-off between confidence that no memorisable samples remain and diminishing sample counts. We identified a suitable value of $T$ through an iterative process of evaluating the ten most similar sample pairs for each evaluation dataset at a possible value for $T$ and increasing this value at each iteration until we found a value at which no memorisable sample pairs were identified but remaining sample counts are reasonable (Table \ref{tab:memo:eval_dataset_counts}). This value was identified as $T=60$. We cross-checked this by searching for the lowest Similarity Score for any sample pair where we considered the evaluation sample to be memorisable. This value was found to be substantially higher than 60, further increasing our confidence that evaluation subsets identifed at $T=60$ were unlikely to contain memorisable samples (the most similar pair for each subset at $T=60$ is shown in Appendix \ref{mem:sec:app_leastsim_mostsim}). We call the resulting subset of samples for each evaluation dataset ``Least Similar''. 

 %We then searched for the lowest Similarity Score for a sample pair where we considered the evaluation sample to be memorisable, which had a Similarity Score of approximately 81. Since this approach is of course fallible we substracted a substantial number from this score to arrive at a memorisability threshold of 60.0. We inspected the ten most similar sample pairs with similarity score under 60.0 for each of our evaluation datasets and were unable to identify any memorisable evaluation samples (the most similar pair for each dataset is shown in appendix \ref{sec:app_leastsim_mostsim}). We call the resulting subset of samples for each evaluation dataset ``Least Similar''. 

Acknowledging the possibility that some number of Least Similar samples could still be memorisable we then took a further subset of Least Similar samples where the answer has no word overlap with the most similar training sample. For brevity we call this further subset ``Unmemorisable'' as shorthand for ``unlikely to be memorisable from our training datasets, including \textit{TDND}''. We note that we are unable to eliminate evaluation samples that have answer overlap with \textit{any} training sample as this would eliminate too many samples. 

It is also worth clarifying that our definition of ``Unmemorisable'' does not preclude a given evaluation sample being memorisable from pretraining data. Since we are comparing performance before and after the intervention with \textit{TDND} datasets it is only strictly necessary that our Unmemorisable samples not be memorisable from \textit{TDND} although in practice we ensure they are not memorisable from any of our \textit{UQA+TDND} datasets.

\subsubsection{Similarity Computation Evaluation - In-Domain Datasets}

We initially evaluate the calibration of our method by considering similarity between the train and development/test splits of our training datasets. As Table \ref{tab:memo:qual0_indomain} shows, identical or near identical sample pairs occur for most training datasets and these tend to score close to 100.

%\newpage
\begin{table}[!ht]
\centering
\tiny
%\resizebox{\columnwidth}{!}{%
%\begin{adjustbox}{width=0.76\textwidth}
%\resizebox{\columnwidth}{!}{%
\begin{tabular}{p{0.1\linewidth} p{0.4\linewidth} p{0.4\linewidth}}
%\begin{tabular}{p{0.11\linewidth} p{0.58\linewidth} p{0.31\linewidth}}
\toprule
\textbf{Dataset} & \textbf{Eval Sample [Split]} & \textbf{Most Similar Train Sample} \\ \midrule
Sci-Elem & Green plants get the energy they need to make food from? \textit{sunlight} [Test] & \textcolor{red}{Identical except for order of multi-choice options. (99.48)} \\
Sci-Mid & Iron oxides such as rust form when iron metal reacts with   oxygen in the air. What are the chemical symbols for the two elements found   in iron oxide? \textit{Fe and O} [Test] & \textcolor{red}{Identical. (100.00)} \\
ARC-Easy & Which of the following elements is best able to combine with   itself and hydrogen {[}H{]} to form large molecules? \textit{carbon {[}C{]}}  [Test] & \textcolor{red}{Identical. (100.00)} \\
ARC-Hard & Students watched a bird fly to and from a large bush every few   minutes. The students told their teacher "The bird has a nest in that bush." This statement is an example of? \textit{an inference made from observations} [Test] & \textcolor{red}{Identical except that one multi-choice option is different. (99.91)} \\
BoolQ & Has an mlb game ever ended in a tie? …The longest game by   innings in Major League Baseball was a 1--1 tie… \textit{Yes} [Dev] & \textcolor{red}{Identical. (100.00)} \\
MCTest & What did Hannah and Mary chase at the park? …Hannah and Mary   ran around chasing butterflies for a little time… \textit{butterflies} [Dev] & What did my granddaughter try to catch? ... granddaughter Tina ... catch ... butterfly... \textit{butterfly} (87.53) \\
OBQA & Oak tree seeds are planted and a sidewalk is paved right next   to that spot until eventually the tree is tall and the roots must extend past   the sidewalk which means? \textit{parts may break the concrete} [Test] & \textcolor{red}{Identical except for order of multi-choice options. (99.95)} \\
RACE & The last sentence in the passage shows that  \_ ? … Little Tommy … said "Well on the first day of school when I saw that man nailed to the plus sign I knew they weren't joking. " \textit{Tommy was afraid of being nailed} [Test] & \textcolor{red}{Identical. (99.99)} \\
SQuAD & Under Elie Metchnikoff's cellular theory what cells were   responsible for immune response? … According to the cellular theory of   immunity … by Elie Metchnikoff it was … phagocytes… \textit{phagocytes} [Dev] & \textcolor{red}{Question is a paraphrase ("Cellular immunology expressed the theory that what cells caused immune responses?"), context and answer are identical. (99.75)} \\ \bottomrule
\end{tabular}%
%}
%\end{adjustbox}
\caption{In-domain Test-Train Overlap. Most similar test-train pairs for each constituent training dataset as measured by Similarity Score (in brackets). The actual evaluation split used is in square brackets. For readability, multi-choice options are removed, remaining context is truncated and answers are in \textit{italics}. The same pair was identified in both SQuAD 1.1 and SQuAD 2 hence shown once. Train samples that are identical or paraphrases to evaluation samples from the same dataset are highlighted in \textcolor{red}{red}.}
\label{tab:memo:qual0_indomain}
\end{table}

\begin{table}[h]
\centering
\tiny
%\resizebox{\columnwidth}{!}{%
%\begin{adjustbox}{width=0.85\textwidth}
%\resizebox{\columnwidth}{!}{%
\begin{tabular}{p{0.1\linewidth} p{0.4\linewidth} p{0.4\linewidth}}
%\begin{tabular}{p{0.1\linewidth} p{0.45\linewidth} p{0.45\linewidth}}
\toprule
\textbf{Eval Dataset} & \textbf{Eval Sample} & \textbf{Most Similar Train Sample} \\ \midrule
DROP & Which household was second most common? … there were 19306 households … 39.9\% were non-families… \textit{non-families} & \textcolor{red}{SQuAD 1.1: What is the second highest demographic for households? … There were 230233 households … 37.4\% were non-families… \textit{non-families} (94.40)} \\
DROP-CS & Which team went scoreless in the third quarter? … Buffalo … connected … 8-yard TD pass for the only score of the period… \textit{Vikings} & TD: Who had the lowest number of field goal yards in total? … Dolphins nailed 26 field goal yards and Vikings nailed 15 field goal yards… \textit{Vikings} (89.96) \\
ROPES & Will Seattle have more or less sulfur oxides in the air than St. Louis? … Seattle has installed a new wind farm and zero emission solar farm to generate power while St. Louis recently installed a coal fired power plant … \textit{less} & SQuAD 1.1: Were sulfonamides more or less toxic than arsphenamine? … Compared to arsphenamine the sulfonamides … were far less toxic … \textit{less} (81.13) \\
NewsQA & What was the score in the Werder Bremen Athletic Bilbao game?   … Werder Bremen beat Athletic Bilbao 3-0 … \textit{3-0} & SQuAD 2: What was the winning score for the game with Real Madrid at Bernabeu stadium? … The pinnacle of the … season … the … Bernabéu Stadium in a 3–0 win over Real Madrid… \textit{3-0} (88.06) \\
%MMLU-M & … patients who drank at least three cups of green tea per day had a much lower incidence of depression … which of the following statements should not be made? \textit{People who want to avoid depression should consider drinking green tea on a daily basis} & RACE: The purpose of the passage is to tell us \_ ? … both studies of green tea and its history show that it is a great thing to drink for your health. \textit{drinking green tea is healthy} (79.97) \\
PIQA & Trees?  \textit{provide homes for animals} & \textcolor{red}{RACE: The story is about \_ ?  … Some animals live in holes in trees … \textit{the homes of some animals} (77.04)} \\
%SIQA & Why did Lee do this? Lee wore a jacket because it was winter and it was cold. \textit{stay warm} & Sci-Elem: Growing thicker fur in the winter helps some animals to? \textit{keep warm} (82.72) \\
CSQA & The water in clouds turn in to what when it gets cold? \textit{snowflake} & \textcolor{red}{ARC-Hard: Which form of water is most likely to appear when the temperature is below freezing? \textit{snow} (87.27)} \\
QASC & What is a tool for indicating air pressure? \textit{barometer} & \textcolor{red}{Sci-Elem: Which weather instrument measures air pressure? barometer (95.14)} \\ \bottomrule
%QASC-IR & What reduces the cause of tuberculosis in milk? …   Pasteurization destroys the bacteria in milk that cause tuberculosis … \textit{Pasteurization} & \textcolor{red}{SQuAD 2: What treatment did the milk that was fed to infants undergo to control bacterial disease? … scurvy among infants … the vast majority of sufferers were being fed milk that had been heat-treated… Pasteurisation was effective against bacteria but it destroyed the vitamin C. \textit{Pasteurisation} (87.08)} \\
\end{tabular}%
%}
%\end{adjustbox}
\caption{Overlap between unseen evaluation and train datasets. Most similar overall sample pair for each evaluation dataset as measured by Similarity Score (in brackets). For readability, multi-choice options are removed, remaining context is truncated and answers are in italics. \textcolor{red}{Red} denotes train samples that could potentially make the corresponding evaluation sample memorisable through contiguous or discontiguous sets of input tokens.}
\label{tab:memo:qual1_unseen_mostsim}
\end{table}

\subsubsection{Similarity Computation Evaluation - Evaluation Datasets}

Turning to our evaluation datasets, we first consider the most similar overall eval-train pair for each evaluation dataset (i.e. the unfiltered versions without removal for Least Similar or Unmemorisable subsets). Generally we find the incidence of identical or near identical pairs is much lower than is the case for the above in-domain evaluation, however memorisable evaluation samples certainly exist as shown in Table \ref{tab:memo:qual1_unseen_mostsim}. In contrast to the above in-domain evaluation where contiguous overlaps of tokens in similar pairs are common, it can be seen that memorisable samples in Table \ref{tab:memo:qual1_unseen_mostsim} generally would not have been detected without a method that can pick up discontinuous token overlaps.

For brevity, the supporting table of Least Similar evaluation-train pairs is in Appendix \ref{mem:sec:app_leastsim_mostsim}, having already noted that we cannot identify any memorisable evaluation samples in that category. Similarly, Appendix \ref{mem:sec:app_unmemorisable_mostsim} shows the most similar evaluation-train pair for Unmemorisable evaluation samples. Unsurprisingly we cannot identify any memorisable evaluation samples here either.

\section{Main Experiment}
\label{sec:memo:main_exper}

All evaluation datasets of RC format are evaluated using the F1 score as formulated by \citet{Rajpurkar2016-fs}. The MC datasets are evaluated by taking the option with the highest overlap with the predicted answer and then scoring as exact match.

The \textit{UQA} and \textit{UQA+TDND} Models are based on BART \citep{Lewis2020-gt}. All models use the Huggingface \citep{Wolf2020-ro} implementations. We train three models for each of \textit{UQA} and \textit{UQA+TDND} respectively using different random seeds and take the mean over each set as our main measure. We ascribe statistical significance to performance change between \textit{UQA} and \textit{UQA+TDND} if it is at the 95\% confidence level (confidence intervals and standard deviations are in Appendix \ref{mem:sec:confidence_intervals}).

\subsection{Experimental Results and Discussion}

Table \ref{tab:memo:main_all} shows the effect of adding the \textit{TDND} datasets to the training regime. Considering the unfiltered evaluation sets comprised of ``All Samples'', it is no surprise that DROP and DROP-CS show a large performance improvement (15.7\% and 19.3\% respectively) since the \textit{TDND} datasets are specifically designed for that purpose. Moving to the Unmemorisable subsets, there is still a 9\% performance improvement for DROP showing that while there is some diminishment, a material performance improvement that is not attributable to memorization remains. DROP-CS improvement is similar but this result is not significant due to the small sample size. While our experiment cannot tell us what mechanism is responsible for this ability to generalise, the intuitive explanation is that \textit{TDND} datasets have as intended imparted relevant numerical reasoning strategies.

\begin{table}[h]
\centering
\resizebox{\textwidth}{!}{%
\begin{tabular}{@{}lr|rrr|rrr|rrr@{}}
\toprule
\textbf{} & \textbf{} & \multicolumn{3}{c|}{\textbf{All Samples}} & \multicolumn{3}{c|}{\textbf{Least Similar}} & \multicolumn{3}{c}{\textbf{Unmemorisable}} \\
\textbf{Eval Dataset} & \textbf{Random} & \textbf{UQA} & \textbf{\begin{tabular}[c]{@{}r@{}}UQA\\ +TDND\end{tabular}} & \textbf{\begin{tabular}[c]{@{}r@{}}\%\\ Change\end{tabular}} & \textbf{UQA} & \textbf{\begin{tabular}[c]{@{}r@{}}UQA\\ +TDND\end{tabular}} & \textbf{\begin{tabular}[c]{@{}r@{}}\%\\ Change\end{tabular}} & \textbf{UQA} & \textbf{\begin{tabular}[c]{@{}r@{}}UQA\\ +TDND\end{tabular}} & \textbf{\begin{tabular}[c]{@{}r@{}}\% \\ Change\end{tabular}} \\ \midrule
DROP &  & 40.2 & 46.5 & \textbf{15.7} & 41.0 & 43.9 & \textbf{7.1} & 41.7 & 45.5 & \textbf{9.0} \\
DROP-CS &  & 32.0 & 38.2 & \textbf{19.3} & 36.3 & 41.8 & 15.3 & 38.5 & 42.2 & 9.6 \\
ROPES &  & 41.2 & 51.9 & \textbf{26.1} & 46.5 & 55.3 & \textbf{18.9} & 41.9 & 52.6 & \textbf{25.7} \\
NewsQA &  & 57.3 & 56.6 & -1.3 & 52.8 & 50.3 & -4.7 & 53.4 & 51.4 & -3.7 \\ 
%MMLU-M & 25.0 & 28.4 & 27.2 & -4.4 & 27.2 & 25.3 & -6.8 & 25.8 & 24.8 & -3.9 \\
PIQA & 50.0 & 63.5 & 62.3 & -1.9 & 62.2 & 61.7 & -0.8 & 60.3 & 60.4 & 0.1 \\
%SIQA & 33.3 & 53.5 & 54.1 & 1.3 & 52.7 & 52.6 & -0.3 & 56.6 & 55.0 & -2.7 \\
CSQA & 20.0 & 55.6 & 55.4 & -0.4 & 61.5 & 61.2 & -0.5 & 60.7 & 61.0 & 0.4 \\
QASC & 12.5 & 37.7 & 36.2 & -3.8 & 35.7 & 34.1 & -4.7 & 36.4 & 33.7 & -7.4 \\ \bottomrule
%QASC-IR & 12.5 & 57.7 & 55.7 & -3.4 & 54.7 & 52.6 & -3.8 & 49.5 & 49.1 & -0.9 \\
\end{tabular}%
}
\caption{Effect of intervention with \textit{TDND} datasets on All, Least Similar, and Unmemorisable evaluation samples. Figures are the mean over three model runs trained with different random seeds. Statistically significant changes at the 95\% confidence level are marked in \textbf{bold} i.e. the improvement for DROP and ROPES is significant in Least similar and Unmemorisable subsets, changes for other datasets are not.}
\label{tab:memo:main_all}
\end{table}

ROPES shows an even larger performance improvement than DROP over All Samples which is largely retained for the unmemorisable subset (26.1\% $\rightarrow$ 25.7\%). Noting that like DROP, ROPES also requires multi-step reasoning over a context and often involves qualitative relations like ``less'' or ``lower'' \citep{Lin2019-ej} it is reasonable to say that benefits imparted by \textit{TDND} samples are responsible for the improvement. For example a typical \textit{TD} sample might involve a judgement such as ``Who had the lowest number of field goal yards in total? ... Dolphins nailed 26 field goal yards and Vikings nailed 15 field goal yards...''

%\begin{table}[h]
%\small
%\centering
%\caption{Standard Deviations for figures presented in Table \ref{tab:main_all}.}
%\label{tab:main_std_1}
%\resizebox{\textwidth}{!}{%
%\begin{tabular}{l|rr|rr|rr|rr}
%\toprule
%\textbf{} & \multicolumn{2}{c|}{\textbf{All Samples}} & \multicolumn{2}{c|}{\textbf{Least Similar}} & \multicolumn{2}{c}{\textbf{Unmemorisable}} \\
%\textbf{Eval Dataset} & \textbf{UQA} & \textbf{\begin{tabular}[c]{@{}r@{}}UQA\\ +TDND\end{tabular}} & \textbf{UQA} & \textbf{\begin{tabular}[c]{@{}r@{}}UQA\\ +TDND\end{tabular}} & \textbf{UQA} & \textbf{\begin{tabular}[c]{@{}r@{}}UQA\\ +TDND\end{tabular}} \\ \midrule
%DROP & 1.0 & 1.0 & 1.8 & 2.0 & 1.3 & 2.2 \\
%DROP-CS & 3.7 & 2.5 & 4.2 & 3.4 & 4.2 & 3.9 \\
%MMLU-M & 1.2 & 0.5 & 1.4 & 1.6 & 1.4 & 1.2 \\
%PIQA & 0.8 & 0.5 & 1.1 & 0.9 & 1.9 & 1.2 \\
%SIQA & 0.8 & 0.2 & 0.5 & 0.1 & 0.6 & 0.1 \\
%CSQA & 1.3 & 0.1 & 0.4 & 2.5 & 0.4 & 4.1 \\
%QASC & 1.0 & 0.7 & 2.9 & 0.9 & 3.8 & 2.7 \\
%QASC-IR & 0.6 & 1.4 & 0.9 & 0.9 & 1.3 & 2.4 \\
%ROPES & 1.7 & 3.1 & 3.5 & 6.5 & 1.7 & 6.2 \\
%NewsQA & 1.3 & 0.9 & 2.4 & 1.9 & 2.1 & 1.6 \\ \bottomrule
%\end{tabular}%
%}
%\end{table}

\subsection{Chapter Limitations}
\label{sec:memo:limitations}

Since our similarity computation (Equation \ref{eq:memo:eq_1}) considers both the question and the answer components it is able to identify evaluation samples that contribute to \textit{inflated} results from the model emitting memorised but correct answers. However using the Equation \ref{eq:memo:eq_1} formulation, we cannot say what could be \textit{deflating} results (e.g. NewsQA and QASC in Table \ref{tab:memo:main_all}). For example, it could be an effect of spurious memorisation where an incorrect answer is emitted based on one or more superficially similar training samples, random perturbation, or it could equally be some other factor such as the result of the incorrect application of some method learned as a result of the \textit{TDND} intervention.

\section{Conclusion}
\label{sec:memo:conclusion}

We have proposed a method of identifying evaluation-train overlap based on semantic similarity of input and output sequences that is reinforced by the further elimination of evaluation samples with overlap in answer terms to the most similar training sample. We have shown that this method is able to identify evaluation samples that are memorisable through both contiguous and non-contiguous token overlap with similar training examples.

To avoid the pitfall of having to compare disparate populations of evaluation samples, as well as to eliminate any dependency on knowing the contents of the pretraining dataset, we have also proposed a method for determining whether or not performance improvement is attributable to memorisation. This involves an intervention through the addition of training datasets that might be expected to improve performance on some evaluation datasets but not on others and measurement of the resulting performance difference. We have shown that for contextualised questions there is significant performance improvement on unmemorisable subsets of DROP and ROPES i.e the improvement is not attributable to memorisation.

%% file: Contents/chapter2_ratd.tex
\chapter{Using Retrieval-Augmented Training Datasets To Improve Reasoning Performance}
\label{sec:ratd}

The research presented in this chapter has been adapted from
\begin{displayquote}
\bibentry{Hartill2023-pf}.
\end{displayquote}
The results of this chapter are available in the GitHub repository \href{https://github.com/timhartill/unseen_questions}{github.com/timhartill/unseen\_questions}

\section{Introduction}
\label{ratd_sec:intro}
As noted, LLMs show an ability to answer questions unlikely to have been encountered during training. Rather than encoding all knowledge in the parameters of a LLM, an alternative approach has been to transform the original question-answering problem into a RC problem by retrieving relevant information for answering a particular query from an external corpus, and training a smaller Reasoning Model to reason over the concatenation of the query and retrieved information to derive an answer e.g. \citet{Chen2017-gw}. In this chapter we extend retrieval methods as described in Section \ref{sec:relwork:retrieval} in conjunction with a supervised multitask pretraining regime for the Reasoning Model involving 79 tasks for our baseline and 93 tasks for the improved model.

The viability of this approach outside of fine-tuned settings is currently subject to limitations, both in the retrieval component, as discussed below, and with respect to the inabilities of smaller language models to perform the reasoning function as well as larger models. We aim to quantify performance limitations and evaluate mitigations for some of them. 

There are at least two significant challenges in retrieval to be overcome. Firstly, no matter how large the corpus is, there will always be missing information, particularly so in our setting where neither datasets nor corpus have been normalised such that sufficient information is in the corpus to make each question answerable through deductively valid means. Secondly, as long as humans ask questions with ambiguous references e.g. ``Who is the spouse of the Green performer?'' \citep{Trivedi2022-mv}, retrieval will necessarily be imperfect even where sufficient knowledge exists in the corpus and the retrieval method is otherwise perfect.

We evaluate a method for addressing these issues. Specifically, we measure the effect of adding datasets to our Reasoning Model training regime that are designed to impart heuristic strategies for reasoning to a plausible rather than an entailed answer. We construct these datasets by building contexts for training questions using our retrieval system against a fixed corpus of English Wikipedia paragraphs. The resulting \textit{RATD} samples are included in training our Reasoning Model irrespective of whether they contain partial, full, or no evidence. Our approach carries the advantage that a diversity of reasoning strategies may be imparted. Such strategies include ignoring an irrelevant context completely or weighing partially evidential facts; e.g. reasoning toward answering ``Do teenagers always rebel against their parents?'' \citep{Talmor2021-al} can be aided by the retrieval of knowledge that ``Adolescents who have a good relationship with their parents are less likely to engage in various risk behaviours'', even though there is no entailment implied. 

Generally our method is most applicable to question-answering tasks where the desired answer is short i.e. from a word to a short sentence, and the question itself does not come already supplied with a fully evidential context. We also assume that it is possible to retrieve sufficient information from our corpus so as to make a question answerable within a modest sequence length (we limit ourselves to a 512 token maximum) e.g. we are unlikely to be able to answer a question such as ``How many songs have a person's name in the title?'' even through retrieving every instance is theoretically possible. 

We focus our study on a subset of the unseen evaluation datasets previously described in Section \ref{sec:prelim:unseen_eval_datasets}, namely StrategyQA \citep{Geva2021-sl}, Musique \citep{Trivedi2022-mv}, IIRC \citep{Ferguson2020-hv}, ARC-DA \citep{Bhakthavatsalam2021-fq}, DROP \citep{Dua2019-td}, and CommonsenseQA \citep{Talmor2019-rm}. \\

\noindent In summary the major contributions of this chapter are: 
\begin{enumerate}
\item We offer what is to our knowledge the most comprehensive set of baselines evaluating smaller Language Model zero-shot reasoning abilities published to date.
\item We show that augmenting the training regime with \textit{RATD} datasets significantly improves performance from the baselines. 
\item We demonstrate that training for numerical literacy and unanswerability is brittle in the unseen setting in the absence of sufficiently similarly formatted training examples. 
\item We propose effective extensions to the retrieval approach as described below. 
\end{enumerate}

\begin{figure*}[h]
% \vskip 0.05in
\begin{center}
\includegraphics[width=\textwidth]{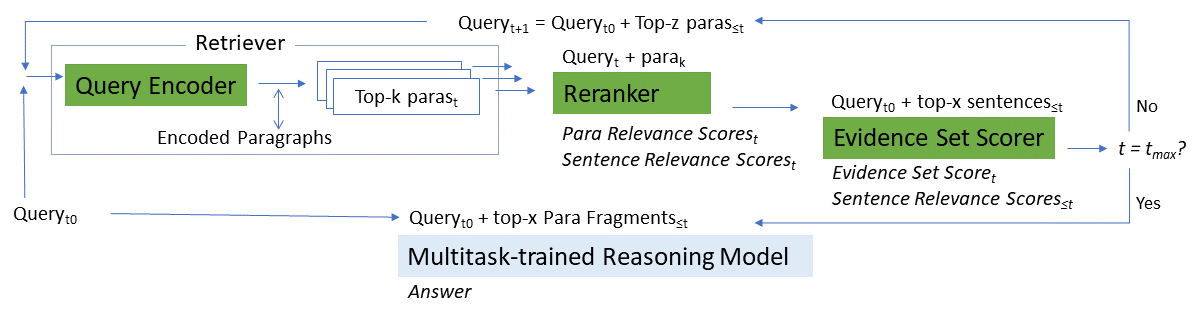}
\caption{Major system components: The Iterator (green boxes) and Reasoning Model (blue box). An initial query for hop $t$=0 is input into the Retriever. The Reranker scores each of the retrieved $k$ paragraphs and constituent sentences. Top-$x$ sentences (Evidence Set\textsubscript{$\leq$t}) are selected from top-ranked sentences from the Reranker and from the prior hop Evidence Set\textsubscript{$<$t}. The query + Evidence Set\textsubscript{$\leq$t} are input into the Evidence Set Scorer which computes an overall Evidence Set Relevance Score $e$ and individual sentence relevance scores. Paragraphs associated with the top five sentences of Evidence Set\textsubscript{$\leq$t} are appended to the query and the process repeats t\textsubscript{max} times. Finally, paragraph fragments recovered from the Evidence Set for hop t=$\argmax(e)$ are concatenated with the original query and input into the Reasoning Model for answer generation.}
\label{ovw_fig}
\end{center}
% \vskip -0.2in
\end{figure*} 
% \vskip 0.1in

\section{Method}
We develop and train the Retrieval, Reranking, Evidence Set Scoring (collectively the ``Iterator''), and Reasoning Model components separately as visualised in Figure \ref{ovw_fig}. Comparisons with retrieval systems in our setting are limited since gold paragraph annotation does not exist. Moreover, excepting \citet{Khashabi2020-gq, Khashabi2022-bq} papers tend not to report zero-shot results for smaller language models such as the BART \citep{Lewis2020-gt} Reasoning Model we use. Therefore we initially evaluate the performance of components on in-domain settings with comparisons to strong prior work, and report results in this section. In subsequent sections we move to the major focus of our study, namely to evaluate our method of adding \textit{RATD} datasets to improve reasoning in the setting where questions are unseen, sufficient evidence to deductively answer a query may not be retrievable, and the model is too small to effectively answer open domain questions without a context to reason over.

\subsection{Retrieval}
For the retrieval component of the Iterator, as discussed in Section \ref{sec:relwork:retrieval}, we extend MDR \citep{Xiong2021-ex} from a two hop maximum to enable training on samples with an arbitrary maximum number of hops ($t\textsubscript{max}$). Training is over a mixture of datasets with questions involving one to four hops to answer; HotpotQA \citep{Yang2018-xq}, Hover \citep{Jiang2020-on}, Natural Questions \citep{Kwiatkowski2019-ef}, and Musique \citep{Trivedi2022-mv}. Hence in practice we set $t\textsubscript{max}$ = 4. Multi-hop questions contain multiple possible reasoning paths through the labelled gold paragraphs, some of which the encoder is able to learn to generalise from (``learnable'') and some not \citep{Xiong2021-ex}. For example, given a set of supporting documents for a 4-hop $q\textsubscript{i}$ as $\{d\textsubscript{i,0}, d\textsubscript{i,1}, d\textsubscript{i,2}, d\textsubscript{i,3}\}$, semantic overlaps between $q\textsubscript{i}$ and the documents might enable learnable reasoning paths of $\langle q\textsubscript{i}, d\textsubscript{i,0}, d\textsubscript{i,1}, d\textsubscript{i,2}, d\textsubscript{i,3}\rangle$ or $\langle q\textsubscript{i}, d\textsubscript{i,1}, d\textsubscript{i,0}, d\textsubscript{i,3}, d\textsubscript{i,2}\rangle$ but not $\langle q\textsubscript{i}, d\textsubscript{i,2}, d\textsubscript{i,0}, d\textsubscript{i,1}, d\textsubscript{i,3}\rangle$ or others. Our training regime samples a learnable reasoning path and builds training samples for subsets; e.g. from $\langle q\textsubscript{i}, d\textsubscript{i,1}, d\textsubscript{i,0}, d\textsubscript{i,3}, d\textsubscript{i,2}\rangle$ we would build four single-hop samples $\langle q\textsubscript{i}\rangle \rightarrow d\textsubscript{i,1}$, $\langle q\textsubscript{i}, d\textsubscript{i,1}\rangle \rightarrow d\textsubscript{i,0}$, $\langle q\textsubscript{i}, d\textsubscript{i,1}, d\textsubscript{i,0}\rangle \rightarrow d\textsubscript{i,3}$ and $\langle q\textsubscript{i}, d\textsubscript{i,1}, d\textsubscript{i,0}, d\textsubscript{i,3}\rangle \rightarrow d\textsubscript{i,2}$. We based document sequencing for learnable reasoning paths for Musique on the decomposed reasoning steps provided with that dataset. For HotpotQA and Hover we used the ordering  that was used in \citet{Xiong2021-ex} and \citet{Khattab2021-jf} respectively, while Natural Questions is treated as single-hop.

For each training sample, positive documents from other training examples in the batch are used as negatives, to which are added two adversarially negative paragraphs specific to that question. Where adversarial negative documents were not otherwise available we created them from our Wikipedia corpus by taking the first paragraph of directly hyperlinked documents from each gold paragraph. Specifically, we used this strategy to create negative documents for Hover as well as to create additional negatives for Musique. We used adversarial negatives for HotpotQA and Natural Questions supplied from \citet{Xiong2021-ex} and \citet{Karpukhin2020-wa} respectively.

Our objective function is similar to others e.g. \citep{Xiong2021-ex, Karpukhin2020-wa}. For hop $t$ of the $i-th$ training sample it models the probability of each next document given a query as: 

\begin{equation}
\label{eq:ratd:eq_1}
P(dvec\textsubscript{i,t+1}|qvec\textsubscript{i,t}) = \frac{exp(dvec\textsubscript{i,t+1} \cdot  qvec\textsubscript{i,t})}{\sum_{dvec\in D_i}exp(dvec \cdot qvec\textsubscript{i,t})}
\end{equation}

 Where $qvec\textsubscript{i,t} = enc(\langle q\textsubscript{i}, d\textsubscript{i,0}, ..., d\textsubscript{i,t}\rangle)$, $dvec\textsubscript{i,t+1} = enc(\langle d\textsubscript{i,t+1}\rangle)$, \textit{enc} is the shared encoder, $qvec\textsubscript{i,t}$ is the encoded query vector, $dvec\textsubscript{i,t+1}$ is the encoded next document vector, $D\textsubscript{i}$ is the set of positive and negative document vectors for $q\textsubscript{i}$ and $\cdot$ denotes the inner product operation.

\subsection{Reranking and Evidence Set Scoring}
\label{sec:ratd:reranking}

To refine retrieved documents we implement a two-stage system comprising Paragraph Reranker and Evidence Set Scoring models. Both models were trained using a mixture of datasets that come with sentence-level annotations, namely HotpotQA, Hover and FEVER \citep{Thorne2018-ws}.

Training samples for the Reranker are built from learnable reasoning paths. For single-hop samples the Reranker is trained with input $\langle q\textsubscript{i}, d\textsubscript{i,0}\rangle \text{ to score } d\textsubscript{i,0}$ relevance. Multi-hop questions can have different phrasing to single-hop questions and so we cannot rely purely on single hop samples for training for proficiency in scoring relevance for the first hop of a multi-hop sample. Therefore, for two-hop paths, samples are randomly built to one or two hops i.e. $\langle q\textsubscript{i}, d\textsubscript{i,0}\rangle \text{ to score } d\textsubscript{i,0}$ relevance, or $\langle q\textsubscript{i}, d\textsubscript{i,0}, d\textsubscript{i,1}\rangle \text{ to score } d\textsubscript{i,1}$. To remediate imbalance in hop distribution three and four hop samples are always built to the respective maximum hop count. Each query is paired with both a positive paragraph to be scored, and a substituted negative paragraph. The sampling function implements a form of shared normalization \citep{Clark2018-oq} such that pairs are positioned in the same training batch.

In the Reranker, a paragraph relevance score ($p$) in addition to individual sentence relevance scores ($s\textsubscript{p}$) are learned. The objective function for each is binary cross-entropy with the overall loss being an unweighted summation (see Appendix \ref{app:para_reranker_model} for details). 

Turning to inference, intuitively, a high-scoring sentence in a relevant paragraph is more likely to be evidential than a high scoring sentence in an irrelevant paragraph. We manually observed that $p$ is often more accurate than $s\textsubscript{p}$ and hence experimented with tuning a weight, $w$, in a sentence scoring function $s = wp+(1-w)s\textsubscript{p}$. For in-domain datasets such as HotpotQA we found non-zero values of $w$ that improved both sentence and paragraph recall by over 2\%, and F1 score by over 6\%, providing evidence that our observation was correct. However the optimal value of $w$ varied between 0.0 and 0.9 over in-domain datasets and tuning $w$ for any of our unseen datasets using their gold annotations would compromise our experimental setup. Hence we simply score each sentence in our main experiments as $s = 0.5p+0.5s\textsubscript{p}$.

For the second stage Evidence Set Scorer, at each hop $t$ the Evidence Set\textsubscript{$\leq$t} is selected from top-ranked sentences from the Reranker and from the prior Evidence Set\textsubscript{$<$t}, if any. The query and Evidence Set\textsubscript{$\leq$t} are input into the Evidence Set Scorer which scores evidence set relevance ($e$), and sentence relevance ($s\textsubscript{e}$) in the context of the evidence set. We retain $p$ for each selected sentence from the Reranker since sentences from highly relevant paragraphs are morely likely to be evidential. The sentences for the $t+1$ evidence set are thus selected by ranking according to $0.5p+0.5s\textsubscript{e}$ and then taking a maximum of five sentences that score over a threshold. The 0.5 coefficients were chosen after a similar evaluation as was done for the Reranker scoring function described above. We observed instances where the evidence set weakened as well as where it strengthened with additional hops, so we then take the evidence set from hop $t=\argmax(e)$ rather than assuming that $t\textsubscript{max}$ always selects the best. 

We observed that a high-scoring sentence is sometimes contextualized by adjacent sentences and collectively they create a stronger rationale. Hence final context for each query, both for \textit{RATD} dataset creation and for creating context for unseen evaluation samples, is created by recovering a paragraph fragment for each selected sentence by prepending/appending the preceding and subsequent sentence from the associated full paragraph where these exist, and then concatenating the document title with the resulting fragment. Ordering of paragraph fragments is by $0.5p + 0.5s\textsubscript{max}$ where $s\textsubscript{max}$ is the maximum Evidence Set Scorer sentence relevance score per paragraph. Using these paragraph fragments it is possible to fit contexts of approximately 6-7 paragraph fragments within a 512-token maximum
sequence length. In the case of datasets such as IIRC \citep{Ferguson2020-hv} that provide an initial paragraph in addition to the question, the initial paragraph is prepended to the context.

The Evidence Set Scoring model is trained with Evidence Sets built as combinations of positive and negative sentences, including replacing positive sentences with negative sentences from positive paragraphs and negative sentences from negative paragraphs. Each question is paired with both a fully evidential set of sentences and a partially evidential (or non-evidential) set of sentences sampled such that pairs are in the same training batch. The objective functions for both $e$ and $s\textsubscript{e}$ are binary cross-entropy and as with the Reranker the final loss is an unweighted summation. The label for $e$ is 1.0 if a subset of the Evidence Set is fully evidential, 0.0 otherwise.

Further details of the Iterator components are in Appendix \ref{sec:app_iterator}.

\subsection{Iterator In-domain Evaluation}
\begin{table*}[h]
\centering
\resizebox{\columnwidth}{!}{%
\begin{tabular}{l|rrrr|rrrr}
\hline
 & \multicolumn{4}{c|}{\textbf{Sentence EM}} & \multicolumn{4}{c}{\textbf{Sentence F1}} \\
\textbf{Model $\downarrow$ \# of Hops $\rightarrow$} & \textbf{2} & \textbf{3} & \textbf{4} & \textbf{All} & \textbf{2} & \textbf{3} & \textbf{4} & \textbf{All} \\ \hline
Baleen 4-hop + FLIPR retriever & 47.3 & 37.7 & \textbf{33.3} & 39.2 & 81.2 & \textbf{82.5} & \textbf{80.0} & \textbf{81.5} \\
Iterator + MDR retriever & 64.6 & 39.3 & 14.8 & 40.1 & 81.7 & 72.1 & 59.0 & 71.4 \\
Iterator + our retriever & \textbf{66.7} & \textbf{45.4} & 27.5 & \textbf{46.8} & \textbf{82.5} & 75.7 & 68.7 & 75.8 \\ \hline
\end{tabular}
}
\caption{In-domain Retrieval and Reranking Evaluation on Hover development set with $k$ = 25. Baleen is finetuned on Hover, MDR is trained on HotpotQA, and our retriever is trained on a mixture of HotpotQA, Hover, Musique and Natural Questions.}
\label{tab:iter_perf_1}
\end{table*}

We initially evaluate performance of the Iterator in an in-domain setting using the Hover development set against the HotpotQA Wikipedia Abstracts Corpus \citep{Yang2018-xq}, since Hover contains samples with up to four hops and it is possible to compare against the published Baleen \citep{Khattab2021-jf} performance. Here the number of paragraphs retrieved on each hop ($k$) is 25. Results (Table \ref{tab:iter_perf_1}) indicate that our Iterator is competitive with Baleen in this setting with our two-hop performance better using both Exact Match and F1 but their four-hop performance dominating. A reason we are stronger overall than Baleen on EM while the reverse is true for F1 is due to our choice of ranking function - Baleen ranks sentences entirely using $s\textsubscript{e}$ whereas we utilise a linear combination of our Reranker paragraph score $p$ and $s\textsubscript{e}$. Unsurprisingly our retriever performance is progressively better than MDR as the number of hops increases.

Our main experiments below use a corpus consisting of English Wikipedia paragraphs from the August 1 2020 dump. Details are in Appendix \ref{sec:app_corpus}.

\subsection{Reasoning Models}
A number of studies have shown the efficacy of supervised multitask training in facilitating generalisation in question-answering tasks \citep{Khashabi2020-gq, Sanh2021-na, Wei2021-go, Khashabi2022-bq}. We adopt this approach for training our Reasoning Models which we characterise as models that take a question and context pair as input $\langle q\textsubscript{i}, c\textsubscript{i}\rangle$ and generate an answer $a\textsubscript{i}$.

To facilitate numerical computation we adapt the Reasoning Model tokenizer for digit tokenisation \citep{Wallace2019-bk, Geva2020-dd} in all experiments. 

Noting that some of the numerical pretraining tasks take much longer to train to a reasonable degree of proficiency than our textual question-answering tasks, we continue training our Reasoning Models from their original pretraining checkpoint with two additional stages of multitask pretraining.

\subsubsection{Stage 1 Pretraining}
In Stage 1 we train using tasks that are aimed at imparting by abstraction a diversity of foundational reasoning skills, with a bias towards simple numerical literacy. Specifically we utilise existing tasks from \citet{Yoran2022-mb}, \citet{Pi2022-um} and \citet{Geva2020-dd} as well as some we create ourselves (see Appendix \ref{app:ratd_combine_models} for details). Stage 1 training is on a total of 33 tasks. One of these is a version of the original self-supervised masked language modelling task which is sampled with probability $\lambda$ = 0.35 so the model retains language understanding skills. The remaining tasks are sampled using an error-based sampling regime \citep{Gottumukkala2020-mt} whereby tasks with low accuracy in the previous validation step are oversampled in the subsequent training steps and vice-versa.

\subsubsection{Stage 2 Pretraining}
In Stage 2, we add five open domain (i.e. question-only) question-answering tasks to the above foundational Stage 1 tasks (for 38 tasks in total, denoted \textit{Group 1}). We add the open domain tasks with the primary aim of teaching the model about the expected form of answer for a given question type e.g. yes or no for ``Could an Aardvark use a knife and fork?'' noting that it has been shown that smaller models cannot learn such open domain tasks well \citep{Lewis2021-ia}. To avoid the possibility of catastrophic forgetting, we continue to train on \textit{Group 1} in conjunction with a new set of tasks, \textit{Group 2}, which is sampled with $\lambda$ = 0.8. \textit{Group 2}, described further below, contains tasks aimed at teaching more question-answering specific reasoning skills, with a bias towards RC datasets.

Our purpose in having two groups is to enable us to implement differing sampling strategies within a single training regime. For \textit{Group 1} we utilise uniform sampling over all tasks and for \textit{Group 2} we use error-based sampling. This combination represents our solution to the issue noted in \citet{Yoran2022-mb}, namely that excessive oversampling will occur for tasks that the model cannot learn well. In addition we find uniform sampling useful for regulating the sampling of the tasks that the model has already learned in Stage 1.

\subsubsection{\textit{Base} and \textit{Base+RATD} Models}
We now discuss two resulting models, both continue training from the best Stage 1 checkpoint and use the same \textit{Group 1} tasks but different in \textit{Group 2} tasks.

The first, our \textit{Base} model, uses 41 tasks in \textit{Group 2} for an overall total of 79 tasks (38 \textit{Group 1} + 41 \textit{Group 2}). \textit{Group 2} consists of a diverse range of question-answering datasets. Of note, to facilitate an ability to identify relevant information and perform deductively valid reasoning, for HotpotQA, Hover, FEVER, Musique, Natural Questions, CREAK \citep{Onoe2021-il} and TriviaQA \citep{Joshi2017-xb}, we construct fully evidential contexts with many irrelevant distractors using a combination of gold and distractor paragraph fragments such that we are as close to our maximum sequence length of 512 tokens as possible without truncating sentences. Since some evaluation samples have a label of ``unanswerable'', we also create versions of HotpotQA, Hover, FEVER and Musique by similar construction to the fully evidential samples but with key gold sentences or paragraphs removed. These are assigned an ``unanswerable'' label.

For our second model, \textit{Group 2} consists of the 41 tasks in the above \textit{Base} Group 2 plus an additional 14 \textit{RATD} datasets for a total of 55 tasks. Our resulting \textit{Base+RATD} model is thus trained on a total of 93 tasks (38 \textit{Group 1} + 55 \textit{Group 2}). As described above, the \textit{RATD} dataset contexts are constructed using our Iterator against the full Wikipedia corpus. Recalling that none of our original datasets are normalised against the version of Wikipedia we use, the resulting contexts are noisy, often containing partial or no relevant evidence and many distractors. We hypothesise that the utility of these is to impart a variety of heuristic strategies using a context form similar to that which our downstream unseen evaluation datasets will have. Thus our \textit{Base+RATD} model may be equipped for reasoning to a plausible answer from partial information as well as the deductively valid answer derivable for the majority of datasets used to train the \textit{Base} model. 

Details of all datasets utilised in Reasoning Model training are in Appendix \ref{sec:app_trainingdatasets}.

\subsubsection{Reasoning Model In-domain Evaluation}

\begin{table}[h]
\centering
%\resizebox{\columnwidth}{!}{%
\small
\begin{tabular}{lrrrr}
\hline
\textbf{Pretraining Regime} & \textbf{Params} & \textbf{DROP} & \textbf{IIRC\textsubscript{G}} & \textbf{IIRC\textsubscript{R}} \\ \hline
POET-SQL (BART)\textsuperscript{\textit{a}} & 440M & 82.2 &  & \\
PReasM (T5-large)\textsuperscript{\textit{b}} & 770M & 72.3 & 75.0 & 45.1 \\
PReasM w/digit tok. (T5-large)\textsuperscript{\textit{c}} & 770M & 80.0 & 73.3 & 40.9 \\
PReasM + Teabreac (T5-large)\textsuperscript{\textit{d}} & 770M & 83.2 & 77.9 & 47.6 \\
Teabreac (T5-3B)\textsuperscript{\textit{d}} & 3B & \textbf{86.7} & 79.5 & 51.0 \\ \hdashline
Ours: \textit{Base} (BART) & 440M & 79.2 & \textbf{80.2} & \textbf{53.6} \\
Ours: \textit{Base+RATD} (BART) & 440M & 79.6 & 80.1 & 52.8 \\ \hline
\end{tabular}%
%}
\caption{Comparison of our Reasoning Model performance to related pretraining methods in \textit{finetuned} setting on DROP dev set and IIRC test set (F1). Our IIRC\textsubscript{R} uses our retrieval from English Wikipedia paragraphs whereas other studies shown use different techniques to retrieve only from provided supporting documents. \textsuperscript{\textit{a}} \citet{Pi2022-um}; \textsuperscript{\textit{b}} \citet{Yoran2022-mb} trained without digit tokenisation; \textsuperscript{\textit{c}} from \citet{Trivedi2022-rh} wherein PReasM is retrained with digit tokenisation; \textsuperscript{\textit{d}} \citet{Trivedi2022-rh}.}
\label{ratd:tab:perf_ft_1}
\end{table}

For comparison with related approaches, we fine-tune our models on DROP \citep{Dua2019-td} and separately on IIRC\textsubscript{G} and IIRC\textsubscript{R} \citep{Ferguson2020-hv}. IIRC\textsubscript{G} is an oracle setting, with context consisting of gold sentences and surrounding text. IIRC\textsubscript{R} has a retrieved context using respective retrieval methods from each study as discussed in Section \ref{sec:prelim:unseen_eval_datasets}. As shown in Table \ref{ratd:tab:perf_ft_1} we are competitive with other approaches in this in-domain setting: We are slightly behind on DROP compared to POET \citep{Pi2022-um} and Teabreac \citep{Trivedi2022-rh}, however we are state of the art on IIRC\textsubscript{G} and IIRC\textsubscript{R}.

\section{Experiments}
\label{ratd_experiments}

Our experiments are aimed at answering three main research questions: \\
\textbf{R1.} What is the impact of adding \textit{RATD} datasets to the Reasoning Model \textit{Base} training regime? \\
\textbf{R2.} How effective is pretraining for numerical literacy in the unseen setting for smaller language models? \\
\textbf{R3.} What is the performance differential between our Reasoning Model with differing evaluation dataset context configurations and high-performing models in a similar unseen setting?

For each evaluation dataset, where possible we report our results against other zero/few-shot work. If known, we also report the current state of the art. As applicable for each dataset we report results without retrieval, with our retrieval (denoted Dataset\textsubscript{R}), and with gold context (denoted Dataset\textsubscript{G} or similar).

To facilitate comparison against prior work on DROP \citep{Dua2019-td} and IIRC \citep{Ferguson2020-hv} we use the numeracy-focused F1 calculation introduced in \citet{Dua2019-td} whereby if the gold label is a number, the predicted answer must contain that number irrespective of other token overlap. For consistency we retain this method for reporting F1 for other datasets noting this is equivalent to standard F1 where the gold answer is not a number and disadvantageous to our results where the gold answer is a number. For datasets with binary labels we adopt the calculation used in \citet{Khashabi2020-gq} where to count as a match the predicted answer must appear in the gold label and the opposing answer must not. For multi-choice evaluation, we take the option with the highest overlap with the predicted answer and then score as exact match.

Where comparing performance of our \textit{Base} against \textit{Base+RATD} models we use the paired bootstrap test \citep{Efron1993-du} to test for statistical significance (p < 0.05).

\subsection{Models}
The Retriever component of the Iterator is built upon RoBERTa-base \citep{Liu2019-ru} and both the Reranker and Evidence Set Scorer use ELECTRA-large \citep{Clark2020-vh}. Unless noted otherwise, all results are reported against the same two final Reasoning Models which are based on BART \citep{Lewis2020-gt}. All models use the the Huggingface \citep{Wolf2020-ro} implementations.

\subsection{Experimental Results}
\subsubsection{StrategyQA and CommonsenseQA}

\textit{Base+RATD} significantly outperforms \textit{Base} on StrategyQA for SQA, SQA\textsubscript{R} and SQA\textsubscript{GP} (Table \ref{tab:comp_sqa}). On SQA\textsubscript{R} (which uses our retrieved contexts) our much smaller \textit{Base+RATD} model slightly exceeds performance of the two 11 billion parameter models and is comparable with OPT 175B \citep{Zhang2022-ts}.

Our \textit{Base} model fails to improve with SQA\textsubscript{GP} (which has contexts of gold paragraphs) versus the question-only SQA version. The improvement on SQA\textsubscript{GP} with the addition of \textit{RATD} draws attention to the fact that outside of our \textit{RATD} datasets the majority of our multihop training samples are aimed at imparting deductively valid forms of reasoning which, as noted above, are often inapplicable for SQA\textsubscript{GP}.  
\begin{table}[h]
\centering
%\resizebox{\columnwidth}{!}{%
\begin{tabular}{@{}lrrr@{}}
\toprule
% &  & \multicolumn{2}{r}{\textbf{Training Regime}} \\
\textbf{Model} & \textbf{Params} & \textbf{Base} & \textbf{Base+RATD} \\ \midrule
Random & & 50.0 & 50.0 \\\hdashline
PaLM - COT+Self-cons.\textsuperscript{\textit{a}} & 540B & 81.6 &  \\
U-PaLM - 5 shot\textsuperscript{\textit{b}} & 540B & 78.3 &  \\
PaLM - 5 shot\textsuperscript{\textit{c}} & 540B & 73.9 &  \\
OPT - 5 shot\textsuperscript{\textit{d}} & 175B & 58.5 & \\
T0++\textsuperscript{\textit{e}} & 11B & 54.4 &  \\
UnifiedQA v2\textsuperscript{\textit{f}} & 11B & 57.9 &  \\
PaLM - 5 shot & 8B & 55.4 &  \\
UnifiedQA v2 & 770M & 51.6 &  \\\hdashline
Ours: SQA & 440M & 51.6 & \textbf{53.9} \\
Ours: SQA\textsubscript{R} (Our retrieval) & 440M & 48.4\textsuperscript{\textit{g}} & \textbf{58.9} \\\hdashline
Ours: SQA\textsubscript{GF} (Gold facts) & 440M & \textbf{72.8} & 71.2 \\
Ours: SQA\textsubscript{GP} (Gold paras) & 440M & 51.6 & \textbf{55.8} \\ \bottomrule
\end{tabular}%
%}
\caption{StrategyQA performance comparison (Accuracy). StrategyQA contains binary-labelled, multi-hop commonsense questions. Bold figures denote the better of our two models. All \textit{Base} versus \textit{Base+RATD} differences are statistically significant. \textsuperscript{\textit{a}} \citet{Wang2022-dr}; \textsuperscript{\textit{b}} \citet{Tay2022-pf}; \textsuperscript{\textit{c}} \citet{Chowdhery2022-yw}; \textsuperscript{\textit{d}} from \citet{Taylor2022-bm}; \textsuperscript{\textit{e}} \citet{Sanh2021-na}; \textsuperscript{\textit{f}} \citet{Khashabi2022-bq} \textsuperscript{\textit{g}} Below-random performance on our \textit{Base} model with Q+retrieval is due to the model predicting text other than yes or no. Prepending ``Yes or no -'' to each question improves the score from 48.4 to 54.9. The corresponding \textit{Base+RATD} figure is 58.8 which retains statistical significance.}
\label{tab:comp_sqa}
\end{table}

\begin{table}[h]
\centering
%\resizebox{\columnwidth}{!}{%
\begin{tabular}{@{}llrr@{}}
\toprule
% &  & \multicolumn{2}{r}{\textbf{Training Regime}} \\
\textbf{Model} & \textbf{Params} & \textbf{Base} & \textbf{Base+RATD} \\ \midrule
Random & & 20.0 & 20.0 \\\hdashline
Prior work (finetuned)\textsuperscript{\textit{a}} & 418M & 91.2 &  \\
PaLM - 0/5 shot\textsuperscript{\textit{b}} & 540B & 69.2/81.5 &  \\
GPT3 - 0/few shot\textsuperscript{\textit{c}} & 175B & 81.5/85.0 &  \\
UnifiedQA v1\textsuperscript{\textit{d}} & 11B & 76.2 &  \\
PaLM - 0/5 shot & 8B & 66.0/77.6 &  \\
GPT3 - 0/few shot & 760M & 61.8/62.7 &  \\
UnifiedQA v1 & 770M & 60.9 &  \\ \hdashline
Ours: CSQA & 440M & 61.1 & \textbf{64.0} \\
Ours: CSQA\textsubscript{R} (Our retrieval) & 440M & 62.4 & \textbf{63.6} \\ \bottomrule
\end{tabular}%
%}
\caption{CommonsenseQA development set performance comparison (Accuracy). CommonsenseQA contains multi-choice commonsense questions. Bold figures denote the better of our two models. \textit{Base+RATD} improvement is statistically significant for CSQA but not for CSQA\textsubscript{R} (adding retrieved context improves \textit{Base} but not \textit{Base+RATD}). \textsuperscript{\textit{a}} \citet{Xu2021-ol}; \textsuperscript{\textit{b}} \citet{Chowdhery2022-yw}; \textsuperscript{\textit{c}} \citet{Brown2020-rl}; \textsuperscript{\textit{d}} \citet{Khashabi2020-gq}}
\label{tab:comp_csqa}
\end{table}

As described in Section \ref{sec:prelim:unseen_eval_datasets}, the contexts of SQA\textsubscript{GF} are of a condensed, rationale-like form, distinct from the standard verbose paragraph form of SQA\textsubscript{GP}. Model performance on SQA\textsubscript{GF} hugely outperforms our other configurations. This shows that with a context of a form the model has learned to reason with, it is possible to solve challenging implicit questions. As to where our models may have learned to reason with this short context form we note that some of the training datasets contain similar short form contexts e.g. BoolQ \citep{Clark2019-gl}, which like StrategyQA has binary labels. Our \textit{Base} model has 84.9\% development set accuracy on BoolQ.

As Table \ref{tab:comp_csqa} shows, augmenting CommonsenseQA samples with retrieval (CSQA\textsubscript{R}) yields mixed results. Others e.g. \citet{Piktus2021-lu} have observed that the best zero/few shot performance on this type of dataset has been achieved with much larger models rather than external retrieval and our analysis bears this out.

The addition of extra reasoning strategies via the \textit{RATD} datasets is more successful; as with StrategyQA, performance on CommonsenseQA is improved with the \textit{Base+RATD} model. 

\subsubsection{DROP and IIRC}
As with PaLM, our \textit{Base} and  \textit{Base+RATD} models are trained using digit tokenization. On DROP both our models outperform all models not trained using this method including GPT3 175B and InstructGPT 175B \citep{Ouyang2022-ti} (Table \ref{tab:comp_drop}). Performance of our models approaches that of PaLM 8B and PaLM 540B in the zero shot setting but both are  superior to ours with a 5-shot prompt. 

\begin{table}[h]
\centering
\resizebox{\columnwidth}{!}{%
\begin{tabular}{@{}lrrr@{}}
\toprule
% & \multicolumn{1}{l}{} & \multicolumn{2}{r}{\textbf{Training Regime}} \\
\textbf{Model} & \textbf{Params} & \textbf{Base} & \textbf{Base+RATD} \\ \midrule
PaLM - 0/5 shot\textsuperscript{\textit{a}} & 540B & 43.7/70.8 &  \\
GPT3 - 0/few shot\textsuperscript{\textit{b}} & 175B & 23.6/36.5 &  \\
InstructGPT PPO+ptx - 0/few shot\textsuperscript{\textit{c}} & 175B & 15.2/33.3 &  \\
UnifiedQA v1\textsuperscript{\textit{d}} & 11B & 32.5 &  \\
PaLM - 0/5 shot & 8B & 45.1/69.4 &  \\
UnifiedQA v1 & 770M & 24.6 &  \\
GPT3 - 0/few shot & 760M & 14.4/24.0 &  \\ \hdashline
Ours & 440M & \textbf{40.7} & 40.0 \\ \bottomrule
\end{tabular}%
}
\caption{DROP development set performance comparison (F1). DROP primarily tests numeracy in reading comprehension. Reduced performance on \textit{Base+RATD} versus \textit{Base} is statistically significant. \textsuperscript{\textit{a}}\citet{Chowdhery2022-yw}; \textsuperscript{\textit{b}}\citet{Brown2020-rl}; \textsuperscript{\textit{c}}\citet{Ouyang2022-ti}; \textsuperscript{\textit{d}} \citet{Khashabi2020-gq}}
\label{tab:comp_drop}
\end{table}

Ablative experiments on our training regime components (Table \ref{tab:drop_ablations}) indicate that digit tokenization, numerical literacy training datasets and two stage training are all important in achieving the best DROP performance in our setting.

\begin{table}[h]
\centering
\begin{tabular}{@{}lrrr@{}}
\toprule
% & \multicolumn{1}{l}{} & \multicolumn{2}{r}{\textbf{Training Regime}} \\
\textbf{Model} & \textbf{All Ans. Types} & \textbf{Numeric Ans. Only} \\ \midrule
Two Stage: +DT +NumLit & \textbf{40.0} & \textbf{25.4} \\
One Stage: +DT +NumLit & 38.2 & 22.9 \\
Two Stage: -DT +NumLit & 34.7 & 16.6\\
One Stage: +DT -NumLit & 29.0 & 11.2\\ \bottomrule
\end{tabular}%
%}
\caption{DROP development set (F1). Ablative results on our Reasoning Models trained using \textit{Base+RATD} datasets trained in one or two stages, with/without digit tokenization (+/-DT), and with/without numerical literacy training datasets (+/-NumLit). Note that the -NumLit setting is only relevant for single-stage training.}
\label{tab:drop_ablations}
\end{table}

\begin{table}[h]
\centering
\resizebox{\columnwidth}{!}{%
\begin{tabular}{@{}llrr@{}}
\toprule
% &  & \multicolumn{2}{r}{\textbf{Training Regime}} \\
\textbf{Model} & \textbf{Params} & \textbf{Base} & \textbf{Base+RATD} \\ \midrule
Prior work: IIRC\textsubscript{R}\textsuperscript{\textit{a}} & 123M & 51.6 &  \\ 
Ours: Finetuned IIRC\textsubscript{R} (Our retrieval)\textsuperscript{\textit{b}} & 440M & 53.6 &  \\ \hdashline
Ours: IIRC\textsubscript{R} (Our retrieval) & 440M & 23.8 & \textbf{25.5} \\ \midrule
Ours: Finetuned IIRC\textsubscript{G} (Gold context)\textsuperscript{\textit{b}} & 440M & 80.2 &  \\ \hdashline
Ours: IIRC\textsubscript{G} (Gold context) & 440M & \textbf{59.6} & 58.1 \\ \bottomrule
\end{tabular}%
}
\caption{IIRC test set evaluation (F1). IIRC tests diverse reasoning requiring retrieval. Both \textit{Base} to \textit{Base+RATD} comparisons are statistically significant. \textsuperscript{\textit{a}} \citet{Ferguson2022-ac} use a finetuned Reasoning Model and specialised retrieval with corpus restricted to documents linked from each initial paragraph. \textsuperscript{\textit{b}} To the best of our knowledge our \textit{Base} model finetuned on IIRC\textsubscript{R} and separately on IIRC\textsubscript{G} are both SOTA at the time of writing so we report these given unavailability of unseen comparisons.}
\label{tab:comp_iirc}
\end{table}

\begin{table}[h!]
\centering
%\resizebox{\columnwidth}{!}{%
\begin{tabular}{@{}llrrr@{}}
\toprule
\textbf{Dataset} & \textbf{Ans. Type} & \textbf{Base+RATD} & \textbf{Finetuned} \\ \midrule
DROP & Span (2962) & 67.4 & 82.3 \\
 & Multi-span (567) & 42.0 & 72.2 \\
 & Num (5850) & 25.4 & 79.0 \\
 & Date (157) & 62.4 & 74.0 \\  \cmidrule{2-4}
 & All (9536) & 40.0 & 79.6 \\ \midrule
IIRC\textsubscript{G} & Span (544) & 59.8 & 74.3 \\ 
 & Binary (66) & 57.1 & 64.7 \\
 & Num (277) & 2.9 & 67.4 \\
 & No answer (414) & 92.8 & 98.8 \\ \cmidrule{2-4}
 & All (1301) & 58.1 & 80.1 \\  \midrule
IIRC\textsubscript{R} & Span (544) & 48.9 & 44.8 \\
 & Binary (66) & 68.2 & 57.6 \\
 & Num (277) & 3.8 & 41.5 \\
 & No answer (414) & 2.6 & 69.9 \\ \cmidrule{2-4} 
 & All (1301) & 25.5 & 52.8 \\ \bottomrule
\end{tabular}%
%}
\caption{Breakdown by answer type on DROP development set and IIRC test set (F1). Sample counts are in brackets. Finetuned models are  trained from the \textit{Base+RATD} checkpoint.}
\label{tab:answer_type}
\end{table}

Table \ref{tab:comp_iirc} shows performance on IIRC. A first glance suggests that poor retrieval is the major cause of low performance on IIRC\textsubscript{R}, however inspection of retrieved items suggests that retrieval is often fully evidential. The breakdown by answer types in Table \ref{tab:answer_type} indicates that a major cause of failure is that in contrast to DROP, almost all numeric answers are predicted incorrectly for both IIRC\textsubscript{G} (gold contexts) and IIRC\textsubscript{R} (retrieved contexts). Finetuning alleviates the issue, confirming that the model is capable of performing the necessary computation when trained with sufficiently similar examples.

Our \textit{Base+RATD} model generally correctly predicts unanswerability for IIRC\textsubscript{G} but almost never does for IIRC\textsubscript{R}. The IIRC\textsubscript{R} context frequently contains either enough information to make the question answerable, or more frequently such relevant information as to make it \textit{appear} answerable. Similar to the numerical computation issue, adding sufficiently similar training examples via finetuning enables the model to distinguish unanswerable samples. Appendix \ref{ratd:sec:appendixfailures} illustrates failure cases for numeric and unanswerable types.

\subsubsection{ARC-DA and Musique}

Table \ref{tab:comp_arcda_mus} shows model performance on our ``partially seen'' datasets, ARC-DA and Musique. On ARC-DA, adding \textit{RATD} datasets significantly improves results in both retrieved and gold settings. By contrast, Musique performance significantly degrades with \textit{Base+RATD}. Noting that Musique is the only evaluation dataset for which we create \textit{RATD} datasets, we hypothesise that in the case of highly similar training examples to particular evaluation samples, the model prediction is the memorised answer of a similar training example. We confirm this by examining the predicted answers of the 1,670 Musique evaluation samples that scored 0 F1 against \textit{Base+RATD}. Of these the predicted answers of 716 samples are an exact match to a Musique training sample gold answer (e.g. ``Who is the spouse of the Green performer?'' is incorrectly answered as ``anna gordy gaye'' because this is the label to a number of training questions of ``Who is the spouse of ...'' form). An ablative experiment, wherein we trained a version of \textit{Base+RATD} without the Musique \textit{RATD} datasets, results in improved performance versus \textit{Base} and the original \textit{Base+RATD} on Musique (Table \ref{tab:comp_arcda_mus}) without material impact to other evaluation dataset results. 

\begin{table}[h]
\centering
%\resizebox{\columnwidth}{!}{%
\begin{tabular}{@{}lrrr@{}}
\toprule
%\textbf{} & \textbf{} & \multicolumn{2}{r}{\textbf{Training Regime}} \\
\textbf{Model} & \textbf{Params} & \textbf{Base} & \textbf{Base+RATD} \\ \midrule
UnifiedQA+ARCDA/MC with IR\textsuperscript{\textit{a}} & 11B & 61.4 &  \\ \hdashline
Ours: ARCDA\textsubscript{R} (Our retrieval) & 440M & 28.8 & \textbf{31.6} \\ %\midrule
Ours: ARCDA\textsubscript{G} (Gold context) & 440M & 56.8 & \textbf{59.1} \\ \midrule
Musique - EX(SA)\textsuperscript{\textit{b}} & 102M & 49.8 &  \\ \hdashline
Ours: Musique\textsubscript{R} (Our retrieval) & 440M & 24.3 & 22.2 (28.2) \\ %\midrule
Ours: Musique\textsubscript{G} (Gold context) & 440M & 60.8 & 43.8 (62.4) \\ \bottomrule
\end{tabular}%
%}
\caption{ARC-DA (test accuracy) and Musique (development F1) comparisons. ARC-DA is science question answering and Musique involves multi-hop question answering. All \textit{Base} to \textit{Base+RATD} differences are statistically significant. Musique performance degradation in \textit{Base+RATD} is caused by \textit{adding} Musique \textit{RATD} in training; results for an ablative model trained with all datasets \textit{except} for Musique \textit{RATD} is shown in brackets in the \textit{Base+RATD} column. \textsuperscript{\textit{a}} \citet{Bhakthavatsalam2021-fq}: Training includes ARC-DA. \textsuperscript{\textit{b}} \citet{Trivedi2022-mv}: EX(SA) uses specialised retrieval from each Musique sample's gold and distractor paragraphs.}
\label{tab:comp_arcda_mus}
\end{table}

The Musique training split has 19,938 samples but only 2,057 unique labels, and questions with the same answer tend to be of similar form, such as the above ``Who is the spouse of...'' example. Therefore we consider the question of whether the poor performance of \textit{Base+RATD} here is a general weakness of our method or whether it is specific to the particular bias of Musique. We trained another \textit{Base+RATD} model, this time with the Musique \textit{RATD} training dataset substituted with a filtered variation that only contains samples with unique labels. Similar to the above Musique \textit{RATD} ablation, this version also significantly improves against the original \textit{Base+RATD} (+3.0 F1 for Musique\textsubscript{R} and +10.6 F1 for Musique\textsubscript{G}) without impact to other results. Hence, assuming appropriate consideration of existing dataset bias when selecting \textit{RATD} training samples, we affirm the robustness of our method. 

\section{Conclusion}
\label{sec:ratd_conclusion}

We have argued that an ability to reason over imperfect and incomplete information is a critical skill with which question-answering models must be endowed. To facilitate such ability we create \textit{RATD} datasets that are designed to impart heuristic reasoning strategies with context of a form similar to that which retrieved contexts for downstream tasks will have. We show that training on \textit{RATD} datasets  improves performance on all unseen evaluation datasets with retrieved contexts. This sometimes comes at a small cost in situations where questions come with gold contexts that are in a form that our model is already good at utilizing (SQA\textsubscript{GF}, DROP, and IIRC\textsubscript{G}) although we suggest that in practice such gold contexts are the less common case. \textbf{(R1)}

We also show that even with our large and diverse pre-training regime, questions involving numerical computation and those labelled unanswerable remain sensitive to the similarity of training samples. \textbf{(R2)}

Our results demonstrate that generic retrieval without normalisation can outperform specialised methods (e.g. we are state of the art on fine-tuned IIRC\textsubscript{R}) and that our overall method can yield performance on par or better than that of much larger models without fine-tuning (e.g. SQA\textsubscript{R}, DROP). \textbf{(R3)}

%% file: Contents/chapter3_combine.tex
\chapter{Combining Rationale Generation and Dense Retrieval}
\label{sec:combine}

\section{Introduction}

\textit{``It was soon realized that the problem of systematically acquiring information from the environment was much less tractable than the mental activities the information was intended to serve'' - \citet{Moravec1988-yw} }

Moravec's paradox is the observation that problems such as developing an ability to reason, that might have been assumed to be one of the most difficult challenges in artificial intelligence has been easier to resolve than the challenge of acquiring more basic knowledge such as sensory information. It is motivating to consider this in the context of recent advances in using both LLMs and retrieval against large textual corpora for information acquisition in the question-answering domain. %Prior work has shown both approaches to be effective, and it may be argued that the knowledge acquisition problem is now as tractable as the problem of utilising this information in reasoning to an answer.

\begin{figure*}[h]
% \vskip 0.05in
\begin{center}
\includegraphics[width=\textwidth]{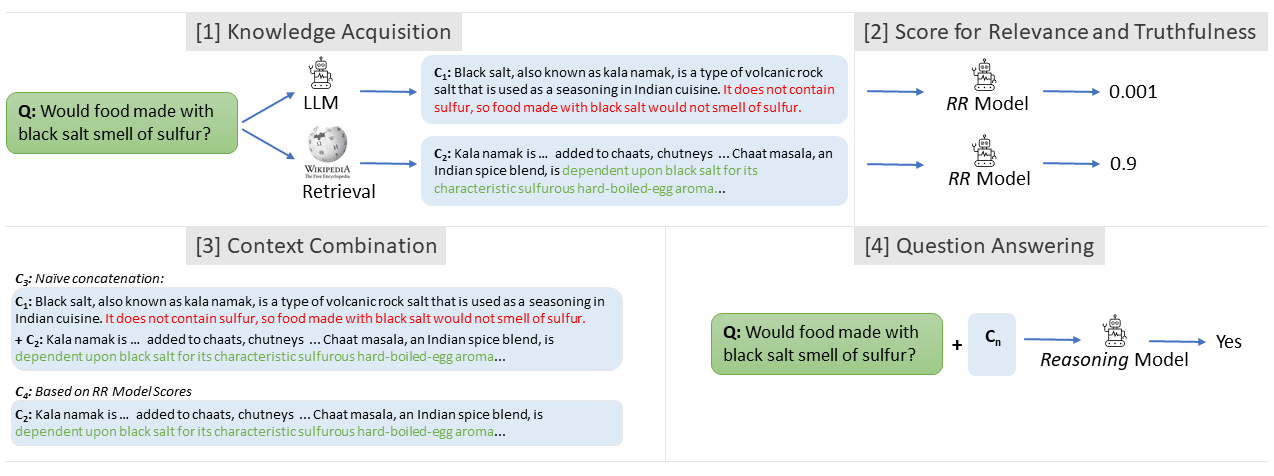}
\caption{Overview of our approach. Given an unseen question \textbf{Q}: [1] we acquire explanatory contexts, \textbf{C\textsubscript{1}} and \textbf{C\textsubscript{2}}, from two knowledge sources. [2] We score the acquired contexts for relevance and truthfulness using a Rationale Ranking (\textit{RR}) model that we train on diverse relevant/irrelevant samples that make both truthful and false assertions. [3] We evaluate and select methods for combining or filtering \textbf{C\textsubscript{1}} and \textbf{C\textsubscript{2}}. [4] We evaluate the performance of different contexts (\textbf{C\textsubscript{n}}) on a set of Reasoning Models that are trained on different mixtures of training datasets, including a mixture containing \textit{RATD} datasets, and a mixture without these. In the diagram, \textcolor{red}{red} denotes false information and {\color[HTML]{10E689}green} highlights relevant and truthful evidence.}
\label{ovw_fig_combine}
\end{center}
% \vskip -0.2in
\end{figure*} 
% \vskip 0.1in

In this chapter, we focus on methods to improve the performance of a smaller Language Model (i.e. Reasoning Model) which, given a question and an acquired explanatory context as input, is expected to reason to provide an answer. To acquire the explanatory context, we consider two knowledge sources both individually and in combination; retrieval of an explanatory context from a corpus of English Wikipedia paragraphs via our Iterator as introduced in Chapter \ref{sec:ratd}, and rationale generation from LLMs. Retrieval has generally been a relatively resource-efficient activity but until recently even inference on LLMs has required considerable computational resources. Recent innovations such as those involving 8-bit matrix multiplication (INT8) \citep{Dettmers2022-te} enable the use of LLMs as frozen knowledge bases in constrained settings. For example inference on the 13 billion parameter StableVicuna model \citep{Stability-AI2023-mp} that we convert to INT8 and use in some experiments runs in approximately 18 GB of GPU RAM, well within the current capacity of large consumer GPU cards.

We choose retrieval from a reliable corpus and LLMs as our knowledge sources since we hypothesise that they may offer differing and complimentary characteristics. Studies such as \citet{Khattab2021-jf}, and our own described in Chapter \ref{sec:ratd}, have shown that multi-hop retrieval systems can be proficient at identifying the relevant $n$ documents necessary to answer $n$-hop factual questions where $n$ can be greater than two, e.g. those found in the Hover \citep{Jiang2020-on} or Musique \citep{Trivedi2022-mv} datasets (``The Rhine forms a border between Aschenbrödel's composer's country and another country where women got the vote when?''). However we are unaware of any corresponding studies on LLMs that demonstrate similar proficiency in generating sufficient information to answer such $n$-hop questions. Conversely, it has been shown that LLMs can be strong at answering commonsense questions without using external retrieval \citep{Lourie2021-kb}, while for such questions retrieval from large textual corpora offers limited benefit as noted by \citet{Piktus2021-lu}, and by us in Chapter \ref{sec:ratd}. 

We explore two methods of combining information from our knowledge sources: (1) Rational Ranking (\textit{RR}), and (2) training with retrieval-augmented data. Our \textit{RR} method involves training a smaller Transformer to score both rationales and retrieved explanatory contexts with respect to relevance and truthfulness. We then evaluate a number of simple strategies to create combined contexts such as including either or both components that score over a threshold, or selecting the single top-scoring component. We focus on identifying combination methods that work best in the general case, i.e. are most likely to work well for an arbitrary unseen question for which we provide no means of predicting which combination method will work best. We find that we are able to identify such a method for each of our Reasoning Models and quantify the performance improvement (Section \ref{rr_ratd_improvement quantification}). Our second method (\textit{RATD}) consists of training our Reasoning Model with our retrieval-augmented datasets previously described in Chapter \ref{sec:ratd}. These datasets were originally developed to impart diverse reasoning strategies such as an ability to identify and weigh partially evidential facts in long, noisy contexts. When our rationales and retrieved contexts are combined, the resulting context is similar in length and form to the \textit{RATD} contexts, therefore we find that training on them enables a single Reasoning Model to utilise our various context formats effectively, including the case where the context consists of the naïve concatenation of rationale and retrieved context that does not consider the \textit{RR} model scores. \\

\noindent The major contributions of this chapter are: 

\begin{enumerate}
\item We propose \textit{RR}, a novel method that both selects context components by relevance, and filters components that may be false. 
\item We apply the \textit{RATD} method that we previously developed to facilitate reasoning over contexts that potentially combine information from multiple knowledge sources. 
\item We demonstrate that both methods in isolation significantly improve reasoning performance in smaller Language Models from strong baselines in the same unseen setting (Section \ref{rr_ratd_improvement quantification}).
\item We show that smaller Language Models trained for reasoning can manifest comparable or stronger performance on unseen questions to a LLM, when provided with the same knowledge to reason over that the LLM is capable of generating for itself (Section \ref{gr_ratd_vs_llm}). 
\item We illustrate the respective strengths and weaknesses of LLMs and multi-hop retrieval from a Wikipedia corpus as knowledge sources (Section \ref{gr_ratd_vs_llm}). 
\item We show that combining information from these sources significantly improves the average performance over evaluation datasets versus using a single source. Additionally, on individual evaluation datasets the combined context performance is often beyond what either knowledge source in isolation can deliver (Section \ref{gr_ratd_vs_llm}).
\end{enumerate}

\section{Method}
\label{sec:method_overview}

To answer an unseen question, $q\textsubscript{i}$, we acquire two contexts:  $c\textsubscript{i,1}$ is obtained by prompting a LLM, and $c\textsubscript{i,2}$ is obtained via dense retrieval. Next, we score $c\textsubscript{i,1}$ and $c\textsubscript{i,2}$ for relevance and truthfulness using the \textit{RR} model. We utilise the \textit{RR} scores in various methods for combining or filtering $c\textsubscript{i,1}$ and $c\textsubscript{i,2}$ into a set of new contexts. Finally, we input the concatenation of $q\textsubscript{i}$ and each resulting context into a set of Reasoning Models and evaluate performance in answering $q\textsubscript{i}$ correctly. A visual overview of our approach is provided in Figure \ref{ovw_fig_combine} where $q$ and $c$ are capitalised and simplified for readability. In the following sections we describe how the two knowledge sources are implemented, how the \textit{RR} model is constructed, trained and initially evaluated, and how the Reasoning Models are trained. We describe our context combination methods further in Section \ref{sec:context_combination_strategies}.

\subsection{Rationale Generation}
\label{sec:combine_rat_generation}

We utilize two LLMs, BLOOM \citep{BigScience_Workshop2022-dh} and StableVicuna \citep{Stability-AI2023-mp}, a much smaller model then BLOOM that has been further tuned from the Vicuna v0 13B model \citep{Chiang2023-xe} which in turn was adapted from the LLama \citep{Touvron2023-nj} foundation model. We chose these two models because they are representative of differing approaches to developing LLMs and they may offer divergent characteristics in rationale generation. At 176 billion parameters, BLOOM was the largest language model we had access to at the time that we could run under INT8. It was trained on 410 billion tokens and the version we used did not undergo further training on instructional data or human feedback. Llama by contrast was trained on one trillion tokens. From the Llama checkpoint, Vicuna underwent further training on user-provided ChatGPT conversations. Finally StableVicuna was developed from Vicuna by further training in both supervised and reinforcement learning from human feedback (RLHF) \citep{Ouyang2022-ti} settings on a mixture of the human-generated OpenAssistant Conversations Dataset \citep{Kopf2023-bx}, as well as human-LLM conversations from the GPT4All \citep{Anand2023-ma} and Alpaca \citep{Taori2023-au} projects. We used StableVicuna under both INT8 and FP16 versions, the former offering a smaller GPU memory footprint at around 18GB while the latter uses almost twice as much memory but we find inference much faster, thus offering a clear trade-off in a resource-constrained setting.

To generate rationales from each model, we used greedy decoding on chain-of-thought (COT) prompts \citep{Wei2022-lz} to generate the rationale followed by the phrase ``So the answer is'' and the answer (examples are in Appendix \ref{sec:app_llmrationaleprompts}). This enabled us to evaluate the LLM answers directly from the same prompts and with the same rationale that our Reasoning Model would use, allowing a comparison under a similar set of assumptions. Occasionally a model would fail to generate the separate answer. In this case, to be favorable to the direct LLM method, the full rationale was used as the answer in calculating metrics. Generated rationale length is a maximum of 128 tokens, which we found to be long enough to accommodate all the rationales we checked.

To maintain the integrity of our unseen settings we ensured that no examples used in prompts were from any of our evaluation datasets. The prompts used were identical between our LLMs excepting that examples for StableVicuna prompts are denoted as: \\

\noindent \texttt{\#\#\# Human:\ [question]\ \#\#\# Assistant:\ [rationale].\ So the answer is [answer].} \\

\noindent BLOOM prompts are denoted as: \\

\noindent \texttt{Q:\ [question]\ A:\ [rationale].\ So the answer is [answer].} \\

Our primary measure of context quality is an ability to improve question-answering performance, however we conducted a high-level qualitative examination of rationales generated by BLOOM and StableVicuna. This suggested that they both tend to produce more rationales containing sufficient information to answer each question on some datasets (e.g. ARC-DA) and more incomplete rationales on the same (e.g. Musique). We observed that BLOOM was generally more prone to generating falsehoods. Examples from both models may be found in Appendix \ref{app:eval_rationale_examples}. We note that robust examination of rationale quality is presently challenging to perform and believe research into automated methods in this area represents a promising future direction.

\subsection{Retrieval}
\label{sec:combine_retrieval}

For our ``retrieval'' knowledge source, as noted we simply reuse contexts previously generated by the Iterator for experiments described in Chapter \ref{sec:ratd}, both for each evaluation sample and also for the creation of \textit{RATD} datasets for the training regimes. As a reminder, Iterator-generated contexts are formatted as a list of paragraph fragments that are recovered from the top-scored sentences, each prepended by the title of the corresponding document and containing the top-scoring set of sentences along with preceding and successor sentences where these exist. The top-scored sentences are identified by taking the Evidence Set from the top-scored hop. Contexts contain as many fragments as will fit into a 512-token sequence length. They are semi-structured as follows: \\

\noindent \texttt{[Doc 1 title]:\ [One to three sentences from a document 1 paragraph].\ [Doc 2 title]:\ ...} \\

%The corpus utilised by the Iterator is obtained from the August 1 2020 English Wikipedia dump and consists of approximately 35 million paragraphs. 

\subsection{Rationale Ranker}
\label{sec:combine_rr_training}

Our \textit{RR} model takes a question and context pair as input $\langle q\textsubscript{i}, c\textsubscript{i}\rangle$ and produces a score $s\textsubscript{i}$. It is trained with a binary cross-entropy objective where samples are labelled 1.0 if $c\textsubscript{i}$ is truthful and fully evidential in answering $q\textsubscript{i}$ or 0.0 otherwise. The model is trained on a mixture of existing datasets for which we acquire or construct positive $c\textsubscript{i}$ (i.e. a set of relevant and truthful gold sentences that are sufficient to answer $q\textsubscript{i}$), and negative $c\textsubscript{i}$ (which omit some or all gold sentences and may be irrelevant, false or both with respect to $q\textsubscript{i}$ answerability). We used shared normalization \citep{Clark2018-oq} such that each $q\textsubscript{i}$ is sampled in the same batch paired with a positive and separately a negative $c\textsubscript{i}$. We found that without shared normalization, model training would collapse and it would predict every $c\textsubscript{i}$ as negative. This may have occurred because without seeing positive and negative $c\textsubscript{i}$ for the same $q\textsubscript{i}$ in the same batch the pattern to be learned is insufficiently signalled.   

\begin{table}[h]
\centering
\resizebox{\textwidth}{!}{%
\begin{tabular}{l|rl|rl}
\toprule
 & & \textbf{Positive Contexts} & & \textbf{Negative Contexts} \\
\textbf{Training Mixture} & \textbf{Count} & \textbf{Construction Methods} & \textbf{Count} & \textbf{Construction Methods} \\ \midrule
Creak\textsuperscript{\textit{a}} (Commonsense) & 10173 & Creak facts\textsuperscript{\textit{a}} & 81408 & LLM-sampled \\
HotpotQA\textsuperscript{\textit{b}} (Multi-hop factual) & 34304 & R4C facts\textsuperscript{\textit{g}}, Iterator-like, Rationale-like & 41839 & LLM-sampled, LLM-greedy, Iterator-like, Rationale-like \\
FEVER\textsuperscript{\textit{c}} (Single-hop factual) & 60986 & Eraser facts\textsuperscript{\textit{h}}, Iterator-like, Rationale-like & 121427 & LLM-sampled, Iterator-like, Rationale-like \\
QASC\textsuperscript{\textit{d}} (Multi-choice science) & 47830 & QASC facts\textsuperscript{\textit{d}}, eQASC facts\textsuperscript{\textit{i}} & 193214 & LLM-sampled, LLM-greedy \\
ARC\textsuperscript{\textit{e}} (Multi-choice science) & 6469 & WorldTree facts\textsuperscript{\textit{j}} & 24492 & LLM-sampled, LLM-greedy \\
Hover\textsuperscript{\textit{f}} (Multi-hop factual) & 28171 & Iterator-like, Rationale-like & 28171 & Iterator-like, Rationale-like \\ \midrule
\textbf{Total} & \textbf{187933} & \textbf{} & \textbf{490551} & \textbf{} \\ \bottomrule
\end{tabular}%
}
\caption{\textit{RR} model training dataset composition. The construction methods denoted ``... facts'' involve creating rationales from gold sentences or structured triples sourced from the cited study. Iterator-like contexts and Rationale-like are constructed from the training datasets' gold (and associated negative) paragraphs. LLM-sampled and LLM-greedy contexts are negative rationales generated by BLOOM using nucleus sampling and greedy decoding respectively. \textsuperscript{\textit{a}}\citet{Onoe2021-il}; \textsuperscript{\textit{b}}\citet{Yang2018-xq}; \textsuperscript{\textit{c}}\citet{Thorne2018-ws}; \textsuperscript{\textit{d}}\citet{Khot2020-sv}; \textsuperscript{\textit{e}}\citet{Clark2016-xg, Clark2018-gy}; \textsuperscript{\textit{f}}\citet{Jiang2020-on}; \textsuperscript{\textit{g}}\citet{Inoue2020-by}; \textsuperscript{\textit{h}}\citet{DeYoung2020-cx}; \textsuperscript{\textit{i}}\citet{Jhamtani2020-dh}; \textsuperscript{\textit{j}}\citet{Xie2020-xb}}
\label{tab:rr_train_datasets}
\end{table}

Since the model must score both rationale-style $c\textsubscript{i}$ and Iterator-generated $c\textsubscript{i}$ on the same scale, we develop training samples that are similar to both types. Obtaining positive $c\textsubscript{i}$ for training questions is generally straightforward. These are constructed from gold sentences and paragraphs associated with each dataset. Negative $c\textsubscript{i}$ that cover both irrelevance and falsehood are harder to obtain. We construct negative $c\textsubscript{i}$ by two methods; (1) generating them from BLOOM using specially constructed few-shot prompts containing examples of negative rationales (e.g. Appendix \ref{app:neg_rationale_prompt}), and (2) creating them synthetically by substituting gold sentences with negative ones using datasets such as HotpotQA that come with sentence level annotations. The synthetic method can only produce irrelevant negatives whereas the LLM-generating method produces both irrelevant and false rationales. For LLM generation we use both greedy decoding and nucleus sampling \citep{Holtzman2019-nw} to create negatives. We find that greedy decoding produces positive-appearing but negative samples but (obtusely) the LLM has a tendency to produce accidentally positive rationales which we must filter out\footnote{We eliminate rationales where the stemmed text contains the stemmed answer string, excepting samples with yes/no labels. We use the snowball stemmer from NLTK \citep{Bird2009-cm}.}. Nucleus sampling by contrast (temperature=0.95 and p=0.96) produces a diversity of false and irrelevant samples that are less likely to be accidental positives. However here falsehoods tend to have an exaggerated quality which could make them less adversarial for the model, so we create samples via both decoding methods (examples in Appendix \ref{app:neg_rationale_examples}). Dataset construction is summarised in Table \ref{tab:rr_train_datasets}.

We employ diverse combination methods involving the trained \textit{RR} model scores to create contexts for our evaluation datasets that combine rationales and Iterator-generated contexts, as described in Section \ref{sec:context_combination_strategies}. 

\subsubsection{Rationale Ranker Evaluation}

Our \textit{RR} development set consists of 89,470 samples taken from the respective development splits of our training datasets. Contexts are created using the same methods as illustrated in Table \ref{tab:rr_train_datasets} for corresponding training splits. We sample a single positive or negative context for each development question such that there are equal counts of positive and negative contexts. As shown in Table \ref{tab:rr_indomain_acc}, accuracy is high in this in-domain setting.

% \usepackage{booktabs}
\begin{table}[h]
\centering
\begin{tabular}{@{}rrr@{}}
\toprule
\textbf{Positive Context} & \textbf{Negative Context} & \textbf{Total} \\ \midrule
91.5 & 93.0 & 92.3 \\ \bottomrule
\end{tabular}
\caption{\textit{RR} model Accuracy on the in-domain development set (score threshold $t=0.5$). Total is micro-accuracy. High accuracy is attainable in detecting both positive and negative contexts.}
\label{tab:rr_indomain_acc}
\end{table}

% \usepackage{booktabs}
% \usepackage[normalem]{ulem}
\begin{table}[h]
\centering
\begin{tabular}{@{}lrr@{}}
\toprule
\textbf{Model} & \textbf{\begin{tabular}[c]{@{}r@{}}TruthfulQA\\ MC1\end{tabular}} &  \\ \midrule
GPT-4 RLHF\textsuperscript{\textit{a}} & 60.0 &  \\
GPT-3.5 RLHF\textsuperscript{\textit{a}} & 47.0 &  \\ \hdashline
GPT-4 No RLHF\textsuperscript{\textit{a}} & 30.0 &  \\
GPT-3 175B\textsuperscript{\textit{b}} & 21.0 &  \\
GPT-J 6B\textsuperscript{\textit{b}} & 20.0 &  \\
UnifiedQA 3B\textsuperscript{\textit{b}} & 19.0 &  \\ \midrule
Iterator Paragraph Reranker 335M\textsuperscript{\textit{c}} & 18.2 &  \\
Rationale Ranker 335M (Ours) & 30.0 &  \\ \bottomrule
\end{tabular}
\caption{Accuracy in detecting falsehoods on TruthfulQA MC1. The \textit{RR} model is better at detecting falsehoods than the Iterator Paragraph Reranker which was trained to detect relevance but not falsehood. It's performance is competitive or better than much larger models that have not been trained using RLHF \textsuperscript{\textit{a}}\citet{OpenAI2023-cv}; \textsuperscript{\textit{b}}from \citet{Lin2022-bq} Github repository; \textsuperscript{\textit{c}}Model described in Chapter \ref{sec:ratd}.}
\label{tab:rr_truthfulqa_sqa}
\end{table}

Turning to an unseen setting, we initially evaluate context relevance scoring with a five-way multi-choice relevance detection dataset that we create from the gold rationales supplied with StrategyQA (SQA), where the four incorrect options are simply randomly assigned rationales from other SQA questions (we use SQA since this is not part of \textit{RR} model training). Here our model achieves 91.4\% accuracy. A more interesting question is the extent to which our relatively small \textit{RR} model is capable of detecting falsehoods in an unseen setting. To evaluate this question we consider TruthfulQA \citep{Lin2022-bq}, an adversarial evaluation-only dataset of 817 questions that models and/or humans tend to answer falsely. In Table \ref{tab:rr_truthfulqa_sqa} we compare falsehood detection performance of the \textit{RR} model with various larger models and in particular with the Iterator Paragraph Reranker. We treat the Paragraph Reranker as representative of models specifically trained to score context relevance but that have not necessarily been trained to consider truthfulness. We utilise the TruthfulQA MC1 split which is formatted as 4-5 way multi-choice with one truthful option. Each option is scored independently of other options and the highest-scoring selected as the prediction. In the case of LLMs the score is calculated as the log-probability of the completion following the question. For the Paragraph Reranker and our \textit{RR} model we use the score that each model has been trained to compute. It can be seen that the \textit{RR} model is indeed much better at detecting falsehoods than the Paragraph Reranker and it's performance is competitive or better than much larger models that have not been trained using RLHF. We imagine the superior performance of LLMs trained with RLHF on falsehood detection is due to their associated large reward models, like our \textit{RR} model, being trained in part to rate samples making false assertions as undesirable.

\subsection{Reasoning Models}
\label{sec:combine_reasoning_models}

We consider three Reasoning Models in our experiments. The first, which we use as a baseline, is the unmodified \textit{``Base+RATD''} model from Chapter \ref{sec:ratd} which we denote here as the \textit{RATD} model for brevity. For descriptive purposes, we divide the datasets used in training the \textit{RATD} model into two sets. The first are the \textit{RATD} datasets described in Section \ref{sec:combine_retrieval}, whose purpose is to confer an ability to reason over long, noisy, and partially evidential contexts. We denote the remaining large number of training datasets as the \textit{Common} set; these broadly cover tasks designed to instill simple numerical literacy, and diverse question-answering ability. Hence we say that the \textit{RATD} model is trained on $\textit{Common} \cup \textit{RATD}$ datasets. 

We create an additional set of training samples denoted \textit{GR} (for ``gold rationales''). These are intended to impart further ability to reason over rationale-form contexts. \textit{GR} consists of samples for Creak, QASC, ARC, HotpotQA, and FEVER where the contexts are gold rationales constructed similarly and from the same sources as those described for the \textit{RR} model training dataset in Table \ref{tab:rr_train_datasets}. 

We then develop our two main Reasoning Models, both multitask-trained using the same two-stage approach and hyperparameters as the original \textit{RATD} model: The \textit{GR} model is trained on $\textit{Common}\ \cup\ \textit{GR}$, and the \textit{GR+RATD} model is trained on $\textit{Common}\ \cup\ \textit{GR}\ \cup\ \textit{RATD}$.

\section{Experiments}
\label{sec:experiments}

We utilise the same unseen evaluation datasets as previously described in Section \ref{sec:prelim:unseen_eval_datasets} excepting DROP which we omit for brevity since it does not require any additional knowledge beyond what is supplied. We use the same metrics for each dataset as we did in Chapter \ref{sec:ratd} (see Section \ref{ratd_experiments}).

%Our experiments are aimed at answering four main research questions:
%
%\textbf{R1.} What is the impact relative to a strong comparable single-source baseline of the following methods of integrating two knowledge sources: context combination using \textit{RR} Model scoring, and the \textit{RATD} training method? \\
%\textbf{R2.} How does our smaller Reasoning Model performance compare to direct prompting of LLMs under similar assumptions? \\
%\textbf{R3.} Can we identify relative strengths and weaknesses of LLM-generated rationales and retrieved paragraph fragments with respect to their usage as explanatory context over diverse question types? \\
%\textbf{R4.} Do combined contexts outperform single-source contexts, both in aggregate over diverse unseen evaluation datasets, and on particular datasets?

\subsection{Models}
The Rationale Ranker is built upon ELECTRA-large \citep{Clark2020-vh}. Reasoning Models are based on BART \citep{Lewis2020-gt}. All models use the the Huggingface \citep{Wolf2020-ro} implementations. The Reasoning Models differ only in their respective training data; hyperparameters are otherwise identical.

\subsection{Context Combination Methods and Experimental Nomenclature}
\label{sec:context_combination_strategies}

\begin{figure*}[h]
% \vskip 0.05in
\begin{center}
\includegraphics[width=\textwidth]{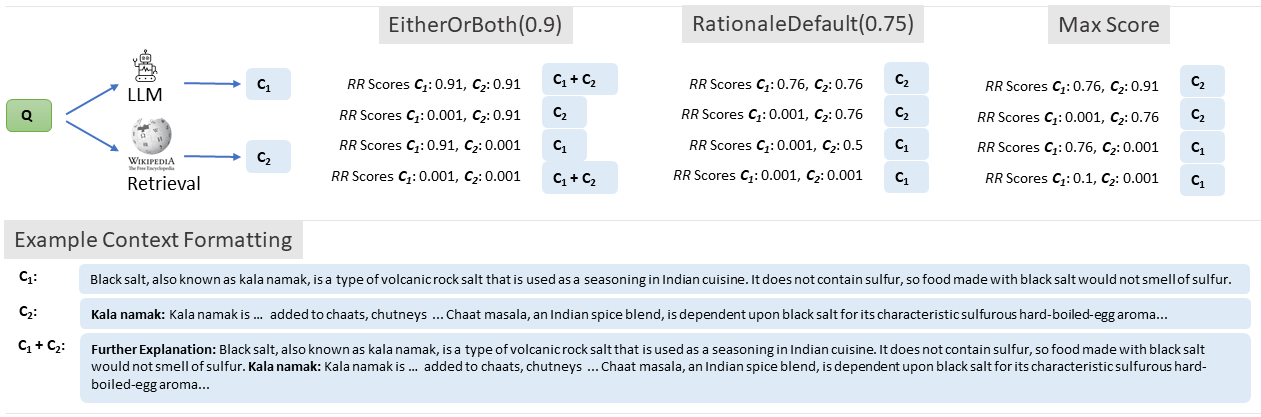}
\caption{Examples of combining contexts. For a question \textbf{Q}, we acquire two contexts, \textbf{C\textsubscript{1}} and \textbf{C\textsubscript{2}}. The resulting combined context for our combination methods with example thresholds and \textit{RR} model scores is then shown in blue boxes where ``+'' denotes the concatenation of \textbf{C\textsubscript{1}} and \textbf{C\textsubscript{2}}. The Naïve Concatenation is always \textbf{C\textsubscript{1}} + \textbf{C\textsubscript{2}}. Formatted examples of resulting contexts are shown at the bottom of the figure with titles shown in bold for readability. The phrase ``Further Explanation'' is added to the rationale in a concatenated context to mimic a document title. }
\label{fig_combo_examples}
\end{center}
% \vskip -0.2in
\end{figure*} 
% \vskip 0.1in

\noindent For each unseen evaluation question, given a LLM-generated rationale, and an Iterator-generated context as possible combined context components, and \textit{RR} model scores for each, we evaluate methods of combining components. We implement four combination methods and create versions of our unseen evaluation datasets with combined contexts for each as follows: \\

\noindent \textbf{Naïve Concatenation}: The simple concatenation of a rationale and corresponding Iterator-generated context with the above form. \textit{RR} model scores are ignored. \\

\noindent \textbf{Max Score}: Choosing the single component that the \textit{RR} model scores highest. \\

\noindent \textbf{RationaleDefault}: Defaulting to taking the rationale component unless the Iterator component scores over a threshold $t$ in which case it is exclusively selected. \\

\noindent \textbf{EitherOrBoth}: Selecting either or both components that score over a threshold $t$. If neither component is selected, we default to the Naïve Concatenation context since smaller Language Models have been shown to be ineffective for answering unmemorized question-only (open domain) questions \citep{Lewis2021-ia}. \\

% 5e{-4}
For the latter two combination methods we create contexts using each of eight \textit{RR} score thresholds ranging from $t=0.0005$ to $t=0.9$. We denote the particular version using the threshold e.g. EitherOrBoth(0.9) means samples are augmented using the EitherOrBoth method with $t=0.9$. Obviously innumerably other combination methods are possible but we find that this set is sufficient for our research purposes while remaining manageable. Figure \ref{fig_combo_examples} illustrates examples of contexts derived from each combination method using hypothetical \textit{RR} scores. Combined contexts are truncated (from the Iterator component) to the maximum sequence length of the model (512 tokens) at inference time.

Each of our three Reasoning Models might be expected to perform better with particular context types. For example the \textit{GR} model might do better where the context tends to be rationale-like whereas the \textit{RATD} model may do better where the context is of Iterator-generated form. This influences which combination method is likely to perform better on each Reasoning Model.

Similarly, different combination methods are likely to work better for differing question types (commonsense, multi-hop factual, etc). For example knowing that LLM-generated rationales tend to be more effective than Iterator-generated contexts for answering commonsense questions, we can deduce that RationaleDefault(0.9) is likely to be a good strategy for developing contexts for CommonsenseQA because using this strategy results in Rationale-only contexts except where the Iterator context is scored very highly. However, we are interested in the situation where our model is presented with an arbitrary question of unknown type. Hence we are more interested in finding combination methods that will \textit{generally} work well under this assumption, even where the method may not be the best for any particular type. We identify combination methods satisfying this criteria as those with the highest \textit{unweighted macro-average score over our unseen evaluation datasets} (henceforth ``Mean'' or ``Mean score'') on each Reasoning Model, taking inspiration for averaging over heterogeneous metrics from e.g. \citet{Wang2019-ll,Wang2019-gf}. For the methods that utilize \textit{RR} model scores we select the highest performing on this measure and refer to it as ``Generally best RR combo'' below. We also report the ``Best RR combo per dataset'' where we select the highest scoring combination method for each evaluation dataset. We note that since we cannot use this approach on an arbitrary question of unknown type we don't consider it a usable method in a truly unseen setting, although future work could remedy this (e.g. through utilising an additional model trained to predict the best combination method for a question). 

We refer below to contexts created for each evaluation dataset that consist entirely of Iterator-generated contexts as ``Iterator only'', those contexts entirely composed of LLM-generated rationales as ``Rationale only'', and those that apply any of the combining methods as ``Rationale + Iterator'' (noting that individual samples in the latter may only contain one of the possible context components). For brevity, where referring to the use of a particular context type on a particular model we use shorthand such as ``\textit{GR+RATD}: Iterator only'' or ``\textit{GR+RATD}: Iterator + Rationale (Naïve Concatenation)''.

To test statistical significance over the large number of model:context combinations created we use methods for accomplishing this described in \citet{Demsar2006-ze} as implemented in the AutoRank library \citep{Herbold2020-vn}. Specifically all tests use significance level $\alpha=0.05$ and we use the non-parametric Friedman test as omnibus test, followed by the Nemenyi test to infer which differences are significant. Generally our key findings are significant as highlighted in the following section. All significance test results are summarised in Appendix \ref{sec:autorank}.

\subsection{Experimental Results}

%\subsubsection{Summary}
As Table \ref{tab:main_1_model_contexttype_overview} indicates, rationales generated by BLOOM almost always produce weaker results than those from StableVicuna. For example, in considering BLOOM-generated ``Rationale only'' contexts, the \textit{GR} model might have been expected to outperform the \textit{RATD} model (given the additional samples with gold rationale contexts added to \textit{GR} training). However the \textit{GR} model actually underperforms (39.5 vs 42.0). Conversely, where considering StableVicuna-generated ``Rationale only'' contexts, the \textit{GR} model slightly outperforms the \textit{RATD} model as expected.

\begin{table}[h]
\centering
\resizebox{\textwidth}{!}{%
\begin{tabular}{l|rrr|rrr}
\toprule
\textbf{Rationale Generator $\rightarrow$ } & \multicolumn{3}{|c}{\textbf{StableVicuna (INT8)}} & \multicolumn{3}{|c}{\textbf{BLOOM (INT8)}} \\
\textbf{Context $\downarrow$ / \textit{Model} $\rightarrow$} & \textbf{\textit{GR}} & \textbf{\textit{RATD}} & \textbf{\textit{GR+RATD}} & \textbf{\textit{GR}} & \textbf{\textit{RATD}} & \textbf{\textit{GR+RATD}} \\ \midrule
Iterator only & 38.1 & 40.4 & \textbf{41.0} & 38.1 & 40.4 & \textbf{41.0} \\
Rationale only & 44.5 & 44.2 & \textbf{45.3} & 39.5 & 42.0 & 40.3 \\
Rationale + Iterator (Naïve concatenation) & 42.7 & 46.3 & \textbf{47.2} & 43.2 & 43.8 & 43.7 \\
Rationale + Iterator (Generally best RR combo) & 45.5 & 46.3 & \textbf{47.2} & 42.9 & 44.2 & 44.4 \\ \midrule%\hdashline
Rationale + Iterator (Best RR combo per dataset) & 47.6 & 47.5 & \textbf{48.1} & 45.1 & 45.6 & 45.4 \\ \bottomrule
\end{tabular}%
}
\caption{Mean score over unseen evaluation datasets. The ``Iterator only'' results are duplicated across Rationale Generators to facilitate comparison. Bold indicates highest score per context type (i.e. per row). StableVicuna-generated rationales generally outperform BLOOM rationales.}
\label{tab:main_1_model_contexttype_overview}
\end{table}

\subsubsection{\textit{GR+RATD} Model Versus Baseline And LLM Direct Prompts}
\label{gr_ratd_vs_llm}

It can be seen in Table \ref{tab:main_1_model_contexttype_overview} that where using the stronger StableVicuna-generated rationales, the \textit{GR+RATD} model results dominate both \textit{RATD} and \textit{GR} models, so we consider this as our best model. Table \ref{tab:main_2_per_dataset} compares \textit{GR+RATD} to our main baseline ``\textit{RATD}: Iterator only''. Both our ``Naïve concatenation'' and ``Generally best RR combo'' combination methods significantly outperform this baseline on the Mean score and on most individual datasets, except for Musique.

\begin{table}[h]
\centering
%\tiny
\resizebox{\textwidth}{!}{%
\begin{tabular}{lrrrrrrr}
\toprule
\textbf{\textit{Model}: Context} & \textbf{SQA} & \textbf{CSQA} & \textbf{ARC-DA} & \textbf{IIRC} & \textbf{Musique} & \textbf{Mean} \\
 & \textbf{(Acc.)} & \textbf{(Acc.)} & \textbf{(F1)} & \textbf{(F1)} & \textbf{(F1)} & \\ \midrule
Random &  50.0 & 20.0 &  &  &  &  \\
Best Prior &  90.4\textsuperscript{\textit{a}} & 91.2\textsuperscript{\textit{b}} & 61.4\textsuperscript{\textit{c}} & 53.6\textsuperscript{\textit{d}} & 49.8\textsuperscript{\textit{e}} & 69.3 \\ \midrule
\textit{RATD}: Iterator only & 58.9 & 63.6 & 31.6 & 25.5 & \textbf{22.2} & 40.4 \\ \hdashline
\textit{BLOOM INT8}: Few Shot Standard Prompt  & 58.1 & 47.5 & \textbf{58.7} & 17.3 & 9.4 & 38.2 \\
\textit{StableVicuna INT8}: Few Shot Standard Prompt & 56.2 & 70.8 & 56.8 & 19.8 & 9.3 & 42.6 \\ \hdashline
\textit{BLOOM INT8}: Few Shot COT Prompt & 57.1 & 54.9 & 50.5 & 17.4 & 11.1 & 38.2 \\
\textit{StableVicuna INT8}: Few Shot COT Prompt  & 61.7 & 67.7 & 45.8 & 20.8 & 12.6 & 41.7 \\ \hdashline
\textit{GR+RATD}: Iterator only & 57.3 & 65.0 & 35.6 & 25.6 & 21.5 & 41.0 \\
\textit{GR+RATD}: Rationale only & \textbf{64.2} & \textbf{73.1} & 50.2 & 25.1 & 13.8 & 45.3 \\
\textit{GR+RATD}: Rationale + Iterator (Naïve concatenation) & 61.7 & 72.6 & 53.0 & 27.0 & 21.7 & \textbf{47.2} \\
\textit{GR+RATD}: Rationale + Iterator (Generally best RR combo) & 61.7 & 72.7 & 52.1 & \textbf{27.3} & 22.0 & \textbf{47.2} \\ \midrule
\textit{GR+RATD}: Rationale + Iterator (Best RR combo per dataset) & 64.5 & 73.3 & 53.0 & 27.4 & 22.4 & 48.1 \\ \bottomrule
\end{tabular}%
}
\caption{Evaluation per dataset. The ``Rationale+Iterator'' combined contexts significantly outperform the ``\textit{RATD}: Iterator only'' baseline and both single-component contexts. The ``Rationale only'' row using StableVicuna-generated rationales significantly outperforms the StableVicuna COT direct prompt. Bold indicates best in column excluding Best Prior and Best RR combo per dataset. Best prior are either not  unseen or involve much larger models as follows: \textsuperscript{\textit{a}}\citet{Anil2023-ws}: Palm 2 using self consistency. \textsuperscript{\textit{b}}\citet{Xu2021-ol}: Finetuned, retrieval from Conceptnet. \textsuperscript{\textit{c}}\citet{Bhakthavatsalam2021-fq}: Training includes ARC-DA. \textsuperscript{\textit{d}}Ours: Finetuned (see Chapter \ref{sec:ratd}). \textsuperscript{\textit{e}}\citet{Trivedi2022-mv}: Specialised retrieval from gold and distractor paragraphs.}
\label{tab:main_2_per_dataset}
\end{table}

We next consider the efficacy of directly prompting both LLMs to produce the answer using few-shot COT exemplars, and separately with standard few-shot prompts that use the same exemplars without the rationale portions. Here, the most like-for-like comparison is from the StableVicuna COT prompt to ``\textit{GR+RATD}: Rationale only'', since the rationales used are the same ones produced by the direct StableVicuna COT prompts. For the StableVicuna COT prompt (and both BLOOM prompts), ``\textit{GR+RATD}: Rationale only'' significantly outperforms the LLM direct prompts on the overall Mean score, and generally on individual datasets (except for ARC-DA). The 42.6 to 45.3 Mean improvement is not significant for the StableVicuna Standard prompt. 

In comparing performance of our combined contexts (``Naïve concatenation'' and ``Generally best RR combo'') to the single-component contexts (``Iterator only'' and ``Rationale only''), both combined contexts achieve a higher Mean score than either single component context does. Improvement from ``Iterator Only'' is significant in both cases, that from ``Rationale Only'' to ``Naïve concatenation'' is significant, while the other is on the significance threshold (Appendix \ref{tab:app:significance_tests}). Notably, three of the five datasets (ARC-DA, IIRC and Musique) have higher scores on either combined context than on any single component context as well. 

Considering the ``Iterator only'' against the ``Rationale only'' rows in Table \ref{tab:main_2_per_dataset} illuminates the relative strengths of our two knowledge sources. Multi-hop factual questions as exemplifed in Musique benefit far more from retrieved paragraphs than LLM-generated rationales (21.5 F1 vs 13.8 F1) whereas commonsense datasets such as SQA (64.2 acc vs 57.3 acc) and CSQA (73.1 acc vs 65.0 acc) unsurprisingly benefit more from LLM-generated rationales as context. IIRC, another factual dataset might have been expected to benefit more from retrieved paragraphs but performance is similar between rationale-only contexts and retrieved paragraphs. We suggest this is because the input for each IIRC sample is comprised of the question and the initial gold paragraph, and many samples then only require a single extra piece of information in order to have sufficient evidence. LLMs may be better at performing (the equivalent of) this single hop than they are at identifying the multiple additional pieces of information necessary in the Musique case.

\subsubsection{\textit{RR} Model Scoring And \textit{RATD} Training Efficacy}
\label{rr_ratd_improvement quantification}

We next evaluate the effectiveness of our methods through an ablational approach. The \textit{GR} model can be regarded as an ablation of \textit{RATD} training from the \textit{GR+RATD} model (-RATD). The Naïve concatenation context type can be seen as an ablation of \textit{RR} Model scoring from the Generally best RR combo (-RR). Hence our ``\textit{GR}: Rationale + Iterator (Naïve concatenation)'' model can be seen as an ablation of both (-RR -RATD) while being (insignificantly) better than the main ``RATD: Iterator only'' baseline (42.7 vs 40.4). Table \ref{tab:ablations_rr_ratd} illustrates the relative efficacy of our two methods, both individually and together. What is revealed is that the \textit{RR} model-scoring approach significantly improves Mean results in the absence of \textit{RATD} training (45.5 vs 42.7), while the \textit{RATD} training significantly improves results in the absence of \textit{RR} scoring (47.2 vs 42.7). The difference between the two methods (45.5 vs 47.2) is \textit{not} significant.

\begin{table}[h]
\centering
\resizebox{\textwidth}{!}{%
\begin{tabular}{@{}llr@{}}
\toprule
\textbf{\textit{Model}: Context} & & \textbf{Mean} \\ \midrule
\textit{GR+RATD}: Rationale + Iterator (Generally best RR combo) & +RR +RATD\textsuperscript{*} & 47.2 \\
\textit{GR+RATD}: Rationale + Iterator (Naïve concatenation) & -RR +RATD\textsuperscript{*} & 47.2 \\
\textit{GR}: Rationale + Iterator (Generally best RR combo) & +RR -RATD\textsuperscript{*} & 45.5 \\
\textit{GR}: Rationale + Iterator (Naïve concatenation) & -RR -RATD & 42.7 \\ \bottomrule
\end{tabular}
}
\caption{\textit{RATD} and \textit{RR} effectiveness. The bottom row can be regarded as an ablation of both \textit{RR} and \textit{RATD} (-RR -RATD). All three topmost methods (marked with an asterisk) are significantly different from the bottow row (-RR -RATD) however differences between the three topmost methods are \textit{not} significant. This shows that the \textit{RR} and \text{RATD} methods are individually both effective but combining the methods does not improve results further.}
\label{tab:ablations_rr_ratd}
\end{table}

Using the two methods in combination does not improve results further. The ``Generally best RR combo'' for the \textit{GR+RATD} model uses the EitherOrBoth(0.9) combination method. This can be interpreted as only selecting a context component if the \textit{RR} model scores it very highly, and since both components frequently fail to meet the threshold the default of using the Naïve concatenation then applies. This has the effect of the context being the Naïve concatenation for 80.9\% of evaluation samples (Appendix \ref{sec:app_context_component_analysis}) which explains why combining the \textit{RATD} and \textit{RR} doesn't result in further improvement in this case.

\section{Conclusion}
\label{sec:combine_conclusion}

We have implemented methods for combining explanatory context from two knowledge sources: LLM-generated rationales and retrieved paragraphs from Wikipedia. The first method involves training our smaller Reasoning Model on \textit{RATD} datasets such that it becomes proficient at reasoning over long, noisy contexts which contain information from both knowledge sources. The second method is to use Rationale Ranking model scores for each knowledge source as guidance in constructing contexts that may contain information from both, or either knowledge source. We have shown that both methods are individually effective in significantly improving unseen question-answering performance both versus the baselines established in Chapter \ref{sec:ratd}, and versus a baseline that ablates both \textit{RR} and \textit{RATD} methods (Section \ref{rr_ratd_improvement quantification}).% (\textbf{R1})

We have shown that smaller Language Models trained to reason can manifest comparable or stronger performance on unseen questions to LLMs, when provided with the same knowledge to reason over that the LLM is capable of generating for itself. (Section \ref{gr_ratd_vs_llm}).% (\textbf{R2})

After comparing results from question-answering using LLM-generated rationales as context with those using retrieved paragraphs we concluded that LLMs are weaker at surfacing the multiple pieces of information necessary to answer multi-hop factual questions, but stronger at generating rationales suitable for answering commonsense questions. Both knowledge sources are found to be effective for question types such as factual questions requiring a single additional piece of information  (Section \ref{gr_ratd_vs_llm}).% (\textbf{R3})

In comparing performance of our combined contexts to the single-component contexts, the combined contexts achieve a higher Mean score over all unseen evaluation datasets than either single component context does. Individually, three of the five datasets (ARC-DA, IIRC and Musique) achieve higher scores when using combined contexts than on any single component context as well  (Section \ref{gr_ratd_vs_llm}).% (\textbf{R4})

%% file: Contents/conclusion.tex
\chapter{Conclusion}
\label{sec:conclusion}

Inspired by the ability of pretrained LLMs to successfully answer a diversity of question types for which they have not been explicitly trained for, but motivated by a desire to explore what is possible in this regard under lower resource assumptions, we initially evaluated whether significantly smaller Language Models have a material capacity to generalise beyond rote memorisation of training data. We followed the positive finding from this study by establishing a set of strong baseline results against diverse unseen evaluation datasets for which comparisons against prior work are available. We then explored diverse methods for improvement from the baselines.

We review our achievements and contributions in Section \ref{sec:conclusion:contributions}, discuss limitations in Section \ref{sec:conclusion:limitations} and provide potential avenues for future research, beyond improving the proposed models, in Section \ref{sec:conclusion:futurework}.

\section{Summary of Contributions} %(copy&paste is ok; link them back to the dissertation)
\label{sec:conclusion:contributions}

We summarise our contributions as follows: \\

\noindent \textbf{In Chapter \ref{sec:memorise}} we proposed a combination of a method for determining train-evaluation overlap and a method for ``intervening'' with additional training datasets to determine memorisable and unmemorisable evaluation samples. Taken together these methods avoided prior experimental weaknesses of (1) inability to control for pretraining data, (2) needing to compare performance between different sets of “clean” and “dirty” samples, and/or (3) inability to detect discontinuous memorisable sequences. We showed that a smaller Language Model is capable of reasoning over an unseen question and context to successfully answer challenging questions that it is unlikely to have memorised at any point in it's training history. \\

\noindent \textbf{Chapter \ref{sec:ratd}} introduced a set of baselines for performance on challenging unseen compositional questions which we established by training our Reasoning Model on a set of 79 tasks, encompassing both existing datasets and those we developed or modified. We proposed the Iterator, our $n$-hop dense retrieval system that incorporates a novel Evidence Set Scoring model into the reranking stages. We used the Iterator in developing novel \textit{RATD} training datasets that are intended to impart diverse reasoning strategies, such as an ability to identify and weigh partially evidential facts in long, noisy contexts. We added \textit{RATD} datasets to the training mixture and showed that this, along with augmenting evaluation questions with a retrieved context, significantly improved performance against our baselines. \\

\noindent \textbf{In Chapter \ref{sec:combine}} we presented a set of methods for combining the retrieval knowledge source developed in Chapter \ref{sec:ratd} with a second knowledge source consisting of rationales generated by larger Language Models. We explored a number of context combination strategies and showed that further significant improvement against the baselines was achievable using both the novel \textit{RR} method, and an adaptation of the \textit{RATD} method. We showed that smaller Language Models trained for reasoning can manifest comparable or stronger performance on unseen questions to a LLM, when provided with the same knowledge to reason over that the LLM is capable of generating for itself. We also identified and discussed the strengths and weaknesses of each knowledge source with respect to the different types of questions encapsulated in each of our baselines. \\

\section{Contributions} %(copy&paste is ok; link them back to the dissertation)
\label{sec:conclusion:contributions_detail}

Here we present a more detailed listing of contributions: \\

\begin{enumerate}
    \item We demonstrated that a smaller Language Model is capable of performance beyond simple memorisation in deriving correct answers to challenging compositional questions. To achieve this we proposed a method of identifying overlap between evaluation and training samples based upon semantic similarity of input and output tokens. We utilised this approach in conjunction with a technique to intervene with additional training datasets to create a Reasoning Model versus a baseline Reasoning Model with no intervention. Our approach enabled us to mitigate effects of pretraining on results and to avoid comparing disparate populations of evaluation subsets as some prior studies have done. After demonstrating the effectiveness of our methods in identifying both memorisable, and unmemorisable samples we were able to show that improved performance on unmemorisable samples is not attributable to the effect of memorisation.
    \item We offer what is to our knowledge the most comprehensive set of baselines evaluating smaller Language Model zero-shot reasoning abilities versus LLM and other approaches published to date. Here our baseline (\textit{Base}) is a multitask-trained Reasoning Model that is trained in two stages on a large number of tasks, both existing and those that we develop.
    \item We proposed the ``Iterator'', a dense retrieval, reranking and evidence set scoring system that aims to identify the relevant $n$ documents necessary to answer $n$-hop questions, where $n$ is arbitrary but we use $n=4$. 
     \item We used the Iterator against a corpus of English Wikipedia paragraphs both to develop contexts for unseen evaluation questions and to develop retrieval-augmented training datasets (\textit{RATD}) which were added to the existing \textit{Base} training regime in training the \textit{Base+RATD} model. \textit{RATD} datasets are intended to impart diverse reasoning strategies, such as an ability to identify and weigh partially evidential facts in long, noisy contexts. We showed that when used in conjunction with our retrieval-augmented evaluation samples, the \textit{Base+RATD} model significantly outperformed the \textit{Base} model on the established baselines. 
    \item We evaluated methods for combining information from two knowledge sources to develop contexts that are more helpful in answering questions. The first knowledge source was the above Iterator with Wikipedia while the second involved rationale generation from larger Language Models that were optimised to run locally in a resource-constrained environment. We proposed ``Rationale Ranking'' (\textit{RR}), a method that both selects context components by relevance, and filters components that may be false. This was accomplished by training a Rationale Ranking model to score LLM-generated rationales and Iterator-generated contexts for \textit{truthfulness} in addition to the more common practice of quantifying relevance. A number of strategies were then evaluated for using the resulting scores to develop contexts that combine information from both knowledge sources. We showed that the \textit{RR} method significantly outperforms the earlier \textit{Base+RATD} baselines. We also showed that models trained using the earlier \textit{RATD} training method were able to generalise sufficiently such that they can successfully utilise combined contexts both in isolation from, and in conjunction with, \textit{RR} scoring. 
    \item We showed that smaller Language Models trained for reasoning can manifest comparable or stronger performance on unseen questions to LLMs, when provided with the same knowledge to reason over that the LLM is capable of generating for itself.
    \item We presented evidence to illustrate the respective strengths and weaknesses of LLMs and $n$-hop retrieval from a Wikipedia corpus as knowledge sources. The LLM tended to offer better performance when considering questions requiring commonsense knowledge (e.g. ``I’m crossing the river, my feet are wet but my body is dry, where am I?''). Retrieval from the Wikipedia corpus tended to be better at extracting knowledge necessary to answer $n$-hop factual questions where $n$ is higher than two (e.g. ``The Rhine forms a border between Aschenbrödel's composer's country and another country where women got the vote when?''). Moreover, we showed that combining information from these sources significantly improved the average performance over evaluation datasets versus using a single source, and on individual evaluation datasets the combined context performance was often beyond what either knowledge source in isolation could deliver.
\end{enumerate}

\section{Limitations}
\label{sec:conclusion:limitations}

Although we consider our contribution to be a promising start, we encountered a number of areas where further exploration may result in further material improvement. These are summarised as follows:

\begin{itemize}
    \item \textbf{Additional or alternative knowledge sources.} The evaluation and inclusion of other knowledge sources (and/or access methods) could yield further benefit, both in terms of improving the sufficiency of explanatory contexts, and in terms of lowering the resource requirements for the knowledge acquisition component. For example, \citet{Huang2023-ow} and others previously have augmented questions through retrieval from a knowledge graph. This could offer a useful and resource-friendly addition to our existing set of knowledge sources.
    \item \textbf{Context combination selection using question type.} In Chapter \ref{sec:combine} we noted that choosing the best context combination method per dataset produced superior results. This is analysed further in Appendix \ref{sec:app_context_component_analysis}. We discounted this approach in our setting as it requires prior knowledge of the questions. However training a model to detect question types and using this information to choose a context combination strategy on a per-question basis seems likely to produce further benefit.
    \item \textbf{Numerical literacy in unseen settings.} We identified in Chapter \ref{sec:ratd} that while applying existing training datasets aimed at imparting numerical reasoning strategies are effective in finetuned settings, they are far less so for unseen questions. Further study of this phenomenon is likely to be fruitful, whether considering the creation or identification of extremely diverse training datasets, or in evaluating further external tool integration.
    \item \textbf{Zero-shot retrieval.} To equip the Iterator retrieval component with an ability to retrieve for arbitrary queries we trained it in a multitask fashion on a mixture of multihop training datasets that have sentence-level annotation. While effective, it seems likely that additional pretraining in self-supervised fashion on large corpora (discussed in the final paragraph of Section \ref{sec:relwork:retrieval}), would reduce the reliance on expensive annotation and perhaps further improve the ability of the Iterator to operate with diverse queries.  
    \item \textbf{Automated Context Quality Evaluation.} As noted in Section \ref{sec:combine_rat_generation}, our purpose in acquiring explanatory context is to improve question-answering performance, and hence our primary measure of context quality is the resulting improvement. Noting some existing research into automated methods of falsehood detection (discussed in Section \ref{sec:relwork:falsehood_detection}), it is possible that some of these approaches are extensible to the more general problem of evaluating context quality along dimensions of (degrees of) sufficiency, necessity and clarity, in addition to truthfulness. Relating these insights to question-answering performance could yield insights into what models find ``useful'' in a context, and hence point to improvement opportunities for \textit{RATD} datasets, Evidence Set scoring, rationale generation and construction of even more effective combined contexts.
\end{itemize}

\section{Outlook}
\label{sec:conclusion:futurework}

Remediating the previously identified limitations would be a direct continuation of this work. Beyond that, we ask the reader's indulgence as we exercise our imagination in considering what the methods we have explored in our work might be more distantly extended towards:

\begin{itemize}
    \item \textbf{Beyond textual question-answering.} Our methods are broadly aimed at equipping smaller models to ``\textit{noisily} reason'' in the face of partial information and distractions obtained by combining information from multiple knowledge sources in a purely textual environment. Evaluation of the prospects for extensibility of our methods into multi-modal situations in addition to pure text, such as visual, auditory or other sensory information, seems a natural path to explore. This could be in the context of integrating a noisy reasoning function into an embodied agent, and/or in the exploration of a role for partially observable, noisy, multi-modal information in the reasoning process itself.
    \item \textbf{Relaxing experimental constraints.} We have focused our experiments on evaluating what is possible to achieve with a smaller Language Model. It is not difficult to imagine a larger model that is further trained and equipped using our methods. Such a model may be more proficient than what our experiments here have shown general-purpose LLMs to be at performing the noisy reasoning function, \textit{and} retaining the ability to be an effective knowledge source.
\end{itemize}

%% file: Contents/appendix1.tex
\appendix

\chapter{Hyperparameters}
\label{sec:app_hyperparams}

\section{Hyperparameters (Chapter \ref{sec:memorise})}
\label{mem:sec:app_hyperparams}

All models are trained on two Nvidia RTX8000 GPUs using 32-bit precision and a linear learning rate decay schedule that reduces the learning rate to zero over 250K training steps. Initial learning rates and other hyperparameters are shown in Table \ref{tab:mem:hyperparams}. The optimiser used is AdamW. A maximum sequence length of 512 tokens was used for all models.

\begin{table}[h]
\centering
%\resizebox{\columnwidth}{!}{%
\begin{tabular}{@{}lrrrr@{}}
\toprule
\textbf{Model} & \textbf{\begin{tabular}[c]{@{}r@{}}Initial\\ LR\end{tabular}} & \textbf{\begin{tabular}[c]{@{}r@{}}Batch\\ Size\end{tabular}} & \textbf{\begin{tabular}[c]{@{}r@{}}Grad.\\ Accum\end{tabular}} & \textbf{\begin{tabular}[c]{@{}r@{}}Train\\ Steps\end{tabular}} \\ \midrule
\textit{UQA} Models & 2e-5 & 32 & 2 & 150K \\
\textit{UQA+TDND} Models & 2e-5 & 32 & 2 & 150K \\ \bottomrule
\end{tabular}%
%}
\caption{Hyperparameters used for each model. Each training step is one batch input i.e the number of optimization steps is $Training Steps / GradientAccumulationSteps$. All final models are selected as the best model on the development sets over the specified number of training steps and validation steps were performed every 10K training steps.}
\label{tab:mem:hyperparams}
\end{table}

\section{Hyperparameters (Chapters \ref{sec:ratd} and \ref{sec:combine})}
\label{sec:app_hyper_ratd_combine}

All models are trained on one GPU (either an Nvidia RTX8000 or A100) except for the Retriever models which are trained on six 80GB A100 GPUs. All models are trained using mixed precision using a linear learning rate decay schedule. Initial learning rates and other hyperparameters are shown in Table \ref{tab:hyperparams_ratd_combine}. The optimiser used for the Retriever, Reranker, Evidence Set Scorer and Rationale Ranker is Adam. All other models use AdamW. All Stage 2 Reasoning Model training starts from the same Stage 1 checkpoint. A maximum sequence length of 512 tokens was used for all models.

% RR: bs=24. batches per epoch = 15662. * 12 epochs = 187,944 train steps.

\begin{table}[h]
\centering
%\resizebox{\columnwidth}{!}{%
\begin{tabular}{@{}lrrrr@{}}
\toprule
\textbf{Model} & \textbf{\begin{tabular}[c]{@{}r@{}}Initial\\ LR\end{tabular}} & \textbf{\begin{tabular}[c]{@{}r@{}}Batch\\ Size\end{tabular}} & \textbf{\begin{tabular}[c]{@{}r@{}}Grad.\\ Accum\end{tabular}} & \textbf{\begin{tabular}[c]{@{}r@{}}Train\\ Steps\end{tabular}} \\ \midrule
Retriever & 2e-5 & 150 & 1 & 99K \\
Retriever+memory bank & 1e-5 & 250 & 1 & 59K \\
Paragraph Reranker & 5e-5 & 12 & 8 & 140K \\
Evidence Set Scorer & 5e-5 & 12 & 8 & 140K \\
Rationale Ranker & 5e-5 & 24 & 8 & 188K \\
Reasoning Model Stage 1 & 2e-5 & 32 & 4 & 1M \\
Reasoning Model Stage 2 \textit{Base} & 2e-5 & 32 & 4 & 1M \\
Reasoning Model Stage 2 \textit{Base+RATD} & 2e-5 & 32 & 4 & 1M \\
Reasoning Model Stage 2 \textit{GR} & 2e-5 & 32 & 4 & 1M \\ 
Reasoning Model Stage 2 \textit{GR+RATD} & 2e-5 & 32 & 4 & 1M \\ 
DROP finetuned & 2e-5 & 32 & 4 & 260K \\
IIRC\textsubscript{G} finetuned & 2e-5 & 32 & 4 & 40K \\
IIRC\textsubscript{R} finetuned & 2e-5 & 32 & 4 & 40K \\ \bottomrule
\end{tabular}%
%}
\caption{Hyperparameters used for each model. Each training step is one batch input i.e the number of optimization steps is $Training Steps / GradientAccumulationSteps$. All final models are selected as the best model on the development set(s) up to the specified maximum number of training steps and validation steps were performed every 10K training steps.}
\label{tab:hyperparams_ratd_combine}
\end{table}

BLOOM loaded under INT8 with a batch size of one consumed approximately 200GB of GPU RAM. StableVicuna also under INT8 with a batch size of one consumed approximately 18GB.

\chapter{Reasoning Model Input Formats}
\label{sec:app_inputformat}

We employed a simple and fixed input format based on that used in UnifiedQA \citep{Khashabi2020-gq} with extensions as follows: \\ \\
Open domain form: \\
\texttt{
[question] \textbackslash \textbackslash n} \\ \\
Reading comprehension (RC) form: \\
\texttt{
[question] \textbackslash \textbackslash n  [context]} \\ \\
Multiple choice form: \\
\texttt{
[question] \textbackslash \textbackslash n (A) [option text a] (B) [option text b] ...} \\ \\
Multiple choice with RC form: \\
\texttt{
[question] \textbackslash \textbackslash n (A) [option text a] (B) [option text b] ... \textbackslash \textbackslash n [context]} \\
\\
Context formats:
\\ \\
Iterator only (also called ``Dataset\textsubscript{R}'' in Chapter \ref{sec:ratd}):

\noindent We standardised the formatting of any paragraphs or paragraph fragments that had associated document titles as follows. Further detail on how such contexts were constructed is in Section \ref{sec:ratd:reranking}. \\
\\
\texttt{
[Title 1]: [Sentences]. [Title 2]: [Sentences]. ...}
\\ \\
Rationale only:
\\
\texttt{
[Sentences].}
\\ \\
Naïve concatenation:
\\
\texttt{
Further Explanation: [Sentences]. [Title 1]: [Sentences]. ...}
\\

\chapter{Wikipedia Corpora}
\label{sec:app_corpus}

For experiments aimed at evaluating the Iterator components in an in-domain setting (Table \ref{tab:iter_perf_1}), we used the same corpus of Wikipedia abstracts from October 1 2017 that HotpotQA and Hover are based upon.

For our main experiments in Chapters \ref{sec:ratd} and \ref{sec:combine}, and for various peripheral tasks such as identifying negative paragraphs for retrieval training we start with the August 1 2020 Wikipedia dump as preprocessed by \citep{Qi2021-sm}. We retain all paragraphs with more than seven words, and extract hyperlinks and calculate sentence offsets from each. There are a total of slightly over 35 million paragraphs. We note that all results in this thesis use the original HotpotQA question set rather than the question set version used in \citep{Qi2021-sm} that has been normalised against this Wikipedia version.  

\chapter{Iterator Training Details}
\label{sec:app_iterator}

\section{Retrieval Model Additional Details}

Our final Retrieval model was trained similarly to \citet{Xiong2021-ex} in that following the initial stage of training, additional training with a large memory bank \citep{Wu2018-in} of negative paragraph embedding vectors was applied.  

For retrieval of paragraphs for \textit{RATD} datasets, the number of paragraphs retrieved at each hop ($k$) was set to 60 so as to complete in reasonable time. In building unseen evaluation dataset contexts $k$ was arbitrarily set to 150 to maintain reasonable performance on queries that are very different to those used in retrieval training.

We used FAISS \citep{Johnson2019-cd} for the search over paragraph embedding vectors. Generally we used an approximate search mechanism, HNSW \citep{Malkov2020-ez}, except for the Hover experiment (Table \ref{tab:iter_perf_1}) where an exact inner product search was employed.

\section{Paragraph Reranker Model}
\label{app:para_reranker_model}

The Reranker has an input format as follows:\\ \\
\texttt{[CLS] query [SEP] yes no [INSUFF] [SEP] title [SM] sentence 0. [SM] sentence 1. ... [SEP]} \\
\\
The query component is encoded as: \\
\\
\texttt{question [QSEP] title 1 | sentence 1. sentence 2. [QSEP] title 2 | sentence 1 ...} \\
\\
Special tokens are utilised as follows: \\
\text{[CLS]}: Trained using a one-layer head to be the Paragraph relevance score with a binary cross-entropy objective. \\
\text{[INSUFF]}: Insufficient Evidence token, used by the start and end token span predictors that are implemented as per \citet{Devlin2019-ox}. Although we utilise a separate abstractive QA model, we use the span predictors as a debugging tool and retain this component in the final loss function. \\
\text{[SM]}: Sentence Marker(s). Used to score sentence relevance. Trained using a one-layer head with a binary cross-entropy objective. \\
\text{[QSEP]}: query components separator. \\

The final training objective is the unweighted summation of the paragraph relevance loss, sentence relevance loss and span loss.

\section{Evidence Set Scoring Model}
\label{app:ev_model}

This model has an input format as follows:\\ \\
\texttt{[CLS] question [SEP] yes no [INSUFF] [SEP] [SM] title 1 | sentence 1. [SM] title 1 | sentence 2. [SM] title 2 | sentence 1 ... [SEP]} \\ \\
Special tokens are utilised as follows: \\
\text{[CLS]}: Evidence Set score. Trained using a one-layer head with binary cross-entropy. The label is 1.0 if all of the gold sentences from all gold paragraphs are present and zero otherwise. \\
\text{[INSUFF]}: Insufficient Evidence token, as per the Reranker model. \\
\text{[SM]}: Sentence Marker, as per the Reranker model. \\

The final training objective is the unweighted summation of the evidence set loss, sentence relevance loss and span loss. 

Following \citet{Khattab2021-jf}, the maximum number of sentences in an evidence set was set to nine in all experiments. To select the sentences for  constructing the retriever query and evidence set for the next hop a maximum of five sentences over a threshold are selected, also following \citet{Khattab2021-jf}. The minimum threshold used to select sentences is 0.1 unless fewer than 2 sentences qualify in which case the two top-scoring sentences are taken.

\chapter{Reasoning Model Multitask Training Details}
\label{sec:app_trainingdatasets}

\section{\textit{UQA} and \textit{UQA+TDND} Models (Chapter \ref{sec:memorise})}
\label{app:memo_models}

The \textit{UQA} model is trained using the same datasets as used by \citet{Khashabi2020-gq}. Our \textit{UQA+TDND} model uses these plus \textit{TD} and \textit{ND} from \citet{Geva2020-dd}. Datasets and development set performance are enumerated in Table \ref{tab:app:uqa_uqatdnd_dev_results}.

\begin{table}[h]
\centering
\begin{adjustbox}{width=0.6\textwidth}
\begin{tabular}{@{}lrr@{}}
\toprule
\textbf{Dataset} & \textbf{UQA} & \textbf{UQA+TDND} \\ \midrule
narrativeqa & 30.3 & 29.6 \\
ai2\_science\_middle & 62.4 & 60.0 \\
ai2\_science\_elementary & 65.0 & 61.0 \\
arc\_hard & 49.5 & 49.2 \\
arc\_easy & 64.0 & 65.8 \\
mctest & 90.0 & 88.1 \\
squad1\_1 & 66.6 & 64.5 \\
squad2 & 68.7 & 68.5 \\
boolq & 84.4 & 84.3 \\
race & 77.9 & 75.6 \\
openbookqa & 65.0 & 64.8 \\
synthetic\_textual (TD) &  & 89.6 \\
synthetic\_numeric (ND) &  & 75.9 \\ \bottomrule
\end{tabular}
\end{adjustbox}
\caption{UQA and UQA+TDND Reasoning Model training datasets. All figures are Exact Match on full development sets from the single overall best model without per-dataset finetuning.}
\label{tab:app:uqa_uqatdnd_dev_results}
\end{table}

%\newpage
\section{\textit{Base}, \textit{Base+RATD}, \textit{GR} and \textit{GR+RATD} Models (Chapters \ref{sec:ratd} and \ref{sec:combine})}
\label{app:ratd_combine_models}

We trained both the first and the second stage of these four models for one million steps (batches) with the best model defined as that with highest mean exact match accuracy over all development sets. To ensure reasonable elapsed time for each validation step we used \textit{reduced} development sets where development sets of more than 1250 samples were reduced to approximately 1250 by taking every $n$th sample with $n=round(c / 1250)$ where $c$ is the sample count. A validation step occurs every 10,000 training steps.

\begin{table}[h]
\centering
\tiny
%\resizebox{\columnwidth}{!}{%
%\begin{adjustbox}{width=0.8\textwidth}
\begin{tabular}{@{}lrrrr@{}}
\toprule
\textbf{Dataset} & \textbf{Base} & \textbf{Base+RATD} & \textbf{GR} & \textbf{GR+RATD }\\ \midrule
creak\_opendomain & 76.6 & 76.1 & 75.4 & 76.9 \\
csqa2\_opendomain & 49.4 & 51.9 & 50.6 & 53.1 \\
triviaqa\_open\_opendomain & 8.0 & 7.4 & 8.2 & 8.0 \\
naturalquestions\_open\_opendomain & 5.4 & 8.7 & 5.5 & 8.6 \\
twentyquestions\_opendomain & 88.8 & 87.9 & 89.0 & 87.9 \\ \hdashline
preasm\_arithmetic\_addition & 99.6 & 99.8 & 99.7 & 99.6 \\
preasm\_arithmetic\_superlatives & 97.9 & 97.9 & 97.5 & 98.1 \\
preasm\_composition & 93.4 & 93.7 & 93.6 & 93.6 \\
preasm\_composition\_2\_hop & 93.5 & 93.7 & 93.3 & 93.8 \\
preasm\_conjunction & 80.2 & 81.0 & 80.1 & 80.8 \\
preasm\_counting & 96.6 & 96.5 & 96.5 & 96.7 \\
preasm\_every\_quantifier & 99.8 & 99.6 & 99.6 & 99.6 \\
preasm\_most\_quantifier & 99.8 & 99.7 & 99.6 & 99.7 \\
preasm\_numeric\_comparison\_boolean & 99.9 & 99.8 & 99.9 & 99.9 \\
preasm\_numeric\_superlatives & 98.1 & 97.9 & 97.9 & 98.0 \\
preasm\_only\_quantifier & 99.4 & 99.4 & 99.3 & 99.4 \\
preasm\_temporal\_comparison & 93.7 & 93.0 & 93.3 & 94.0 \\
preasm\_temporal\_comparison\_boolean & 99.8 & 99.7 & 99.8 & 99.7 \\
preasm\_temporal\_difference & 94.3 & 95.1 & 94.7 & 95.0 \\
preasm\_temporal\_superlatives & 97.5 & 97.1 & 97.3 & 97.5 \\
poetsql\_multi & 36.2 & 34.5 & 36.0 & 36.1 \\
poetsql\_select\_abs & 84.2 & 94.0 & 84.8 & 84.4 \\
poetsql\_select\_arith & 89.7 & 85.1 & 90.8 & 90.0 \\
poetsql\_select\_count & 80.8 & 80.2 & 80.2 & 80.1 \\
poetsql\_select\_max & 79.6 & 75.7 & 74.8 & 78.3 \\
poetsql\_select\_min & 82.5 & 81.3 & 82.9 & 82.5 \\
poetsql\_select\_sum & 50.6 & 52.7 & 53.5 & 54.1 \\
poetsql\_single & 79.4 & 79.0 & 78.8 & 80.4 \\
synthetic\_num\_arg\_min\_max & 100.0 & 100.0 & 100.0 & 100.0 \\
synthetic\_num\_date\_diff & 82.6 & 82.7 & 82.6 & 82.7 \\
synthetic\_num\_date\_min\_max & 93.2 & 95.7 & 92.2 & 94.9 \\
synthetic\_num\_min\_max\_avg & 69.3 & 68.8 & 70.0 & 68.9 \\
synthetic\_num\_percent & 99.0 & 98.2 & 97.8 & 99.2 \\
synthetic\_num\_signed\_arith & 76.1 & 78.6 & 79.4 & 78.2 \\
synthetic\_num\_yn\_dates & 99.8 & 99.8 & 99.8 & 99.7 \\
synthetic\_num\_yn\_nums & 100.0 & 100.0 & 100.0 & 100.0 \\
synthetic\_textual & 92.4 & 92.4 & 92.7 & 93.5 \\
enwiki\_20200801\_selfsvised & 22.5 & 24.1 & 26.3 & 23.3 \\ \bottomrule
\end{tabular}%
%\end{adjustbox}
%}
\caption{\textit{Base}, \textit{Base+RATD}, \textit{GR} and \textit{GR+RATD} Reasoning Model Stage 1 and Stage 2 \textit{Group 1} training datasets.  All figures are Exact Match on \textit{reduced} development sets from the single overall best model without per-dataset finetuning. Datasets above the dotted line were added for Stage 2.}
\label{tab:stage1andoddatasets}
\end{table}

\begin{table}[h!]
\centering
\tiny
%\resizebox{!}{\textheight}{%
%\begin{adjustbox}{width=0.78\textwidth}
%\tiny  % us: added
\begin{tabular}{@{}lrrrr@{}}
\toprule
\textbf{Dataset} & \textbf{Base} & \textbf{Base+RATD} & \textbf{GR} & \textbf{GR+RATD} \\ \midrule
adversarialqa\_all & 46.0 & 47.6 & 47.4 & 45.9 \\
ai2\_science\_middle & 67.2 & 63.2 & 64.8 & 64.0 \\
ai2\_science\_elementary & 67.5 & 69.9 & 66.7 & 68.3 \\
arc\_hard & 56.5 & 54.2 & 55.5 & 56.2 \\
arc\_hard\_with\_ir & 59.5 & 59.5 & 58.2 & 58.9 \\
arc\_easy & 68.3 & 70.4 & 67.7 & 69.5 \\
arc\_easy\_with\_ir & 77.5 & 79.3 & 76.1 & 76.5 \\
boolq & 84.7 & 84.2 & 85.1 & 85.1 \\
boolq\_np & 82.0 & 81.7 & 82.8 & 83.1 \\
creak\_goldplusdistractors & 85.2 & 83.8 & 84.0 & 85.0 \\
creak\_ratd &  & 85.9 &  & 85.8 \\
creak\_ratd\_max4paras &  & 85.6 &  & 85.5 \\
csqa2\_ratd &  & 56.7 &  & 57.8 \\
csqa2\_ratd\_max4paras &  & 57.7 &  & 57.5 \\
fever\_goldplusdistractors & 85.9 & 89.2 & 82.9 & 82.0 \\
hover\_goldplusdistractors & 84.0 & 82.2 & 83.2 & 82.2 \\
hover\_ratd &  & 78.5 &  & 76.8 \\
hover\_ratd\_max4paras &  & 77.2 &  & 75.9 \\
hotpotqa\_goldplusdistractors & 65.9 & 66.7 & 66.9 & 85.0 \\
hotpotqa\_ratd &  & 53.0 &  & 52.4 \\
hotpotqa\_ratd\_max4paras &  & 52.5 &  & 51.8 \\
mctest & 91.3 & 90.0 & 90.0 & 89.4 \\
multirc & 100.0 & 100.0 & 99.7 & 100.0 \\
musique\_goldplusdistractors & 88.0 & 87.2 & 87.7 & 86.7 \\
musique\_qa\_ratd &  & 74.4 &  & 76.2 \\
musique\_qa\_ratd\_max4paras &  & 75.1 &  & 76.2 \\
narrativeqa & 30.0 & 29.1 & 29.0 & 30.2 \\
naturalquestions\_goldplusdistractors & 56.5 & 58.8 & 54.4 & 57.9 \\
naturalquestions\_open\_ratd &  & 40.9 &  & 40.1 \\
naturalquestions\_open\_ratd\_max4paras &  & 39.9 &  & 40.1 \\
newsqa & 44.3 & 44.4 & 44.1 & 43.8 \\
hotpotqa\_fever\_hover\_noanswer & 83.5 & 76.9 & 84.7 & 86.0 \\
musique\_noanswer & 96.6 & 95.6 & 95.6 & 95.6 \\
pubmedqa\_pqal\_short\_ans & 99.2 & 100.0 & 100.0 & 99.8 \\
qaconv & 54.3 & 54.8 & 54.3 & 55.4 \\
quail & 78.1 & 76.1 & 76.4 & 76.9 \\
quoref & 71.5 & 70.2 & 70.4 & 70.2 \\
record & 53.1 & 53.1 & 51.9 & 52.3 \\
ropes & 77.6 & 81.8 & 80.4 & 79.4 \\
squad1\_1 & 66.5 & 64.9 & 65.4 & 64.8 \\
squad2 & 66.2 & 67.4 & 68.3 & 68.3 \\
tweetqa & 34.5 & 33.6 & 33.5 & 33.6 \\
tatqa & 41.6 & 40.8 & 41.2 & 41.2 \\
triviaqa\_goldplusdistractors & 63.9 & 65.3 & 64.8 & 64.2 \\
openbookqa & 67.2 & 69.2 & 68.8 & 68.0 \\
openbookqa\_with\_ir & 68.4 & 70.6 & 69.8 & 68.4 \\
physical\_iqa & 66.9 & 67.0 & 66.2 & 67.1 \\
qasc & 53.4 & 55.5 & 57.2 & 56.2 \\
qasc\_with\_ir & 72.0 & 70.8 & 71.7 & 69.3 \\
qasc\_ratd &  & 61.9 &  & 59.5 \\
qasc\_ratd\_max4paras &  & 62.6 &  & 60.0 \\
race & 76.4 & 74.8 & 75.6 & 73.8 \\
reclor & 43.0 & 41.4 & 42.8 & 42.8 \\
social\_iqa & 75.1 & 74.0 & 74.4 & 75.5 \\
winogrande\_xl & 69.7 & 69.1 & 69.5 & 69.0 \\ \bottomrule
\end{tabular}%
%\end{adjustbox}
%}
\caption{\textit{Base}, \textit{Base+RATD}, \textit{GR} and \textit{GR+RATD} Reasoning Model \textit{Group 2} training datasets, excluding \textit{GR} datasets. All figures are Exact Match on \textit{reduced} development sets from the single overall best model without per-dataset finetuning. Blank entries in the table indicate where \textit{RATD} datasets are not included for particular models.}
\label{tab:stage2addeddatasets}
\end{table}

Table \ref{tab:stage1andoddatasets} enumerates datasets used in Stage 1 and in Stage 2 \textit{Group 1} (those above the dotted line were added for Stage 2, namely CREAK \citep{Onoe2021-il}, CommonsenseQA 2.0 \citep{Talmor2021-al}, TriviaQA \citep{Joshi2017-xb},  Natural Questions \citep{Kwiatkowski2019-ef} and Twenty Questions\footnote{https://github.com/allenai/twentyquestions}). During Stage 1 training, error-based sampling for these datasets was employed and in Stage 2, uniform sampling.

Datasets names containing the term ``opendomain'' only use the question text as input and are added with the primary aim of teaching the model about the expected form of answer for a given question type (e.g. yes or no for ``Could an Aardvark use a knife and fork?''.

Datasets preceded by ``preasm'' are as provided by \citet{Yoran2022-mb} with reformatting into our standard form. Datasets preceded by ``poetsql'' are the POET-SQL dataset kindly provided to us by the authors of \citet{Pi2022-um}. We split POET-SQL into separate datasets based on the type of SQL statement and converted into our standard form.

For the ``synthetic\_num'' datasets we extended the original code provided by \citet{Geva2020-dd} to output in the variablised form proposed in \citet{Pi2022-um} (e.g. ``1 + 3'' becomes ``x + y \textbackslash \textbackslash n x=1; y=3; z=0; ...'' where z is a distractor). Additionally we added two datasets with questions of the form ``Is x > | < | between y [and z]?'' for numbers and dates respectively. We generated one million samples for each of the resulting eight datasets. The ``synthetic\_textual'' task is as provided by \citet{Geva2020-dd} aside from reformatting into our standard format.

Finally, we created a self-supervised task (enwiki\_20200801\_selfsvised), by sequentially concatenating paragraphs from documents in our Wikipedia dump until a sequence length of approximately 512 tokens was reached. During training, spans were masked from each sample input based on their being named entities \citep{Guu2020-vq} or noun phrases with $\lambda$ = 0.65, or randomly with $\lambda$ = 1 - 0.65. The training objective was to predict just the masked spans as with T5 \citep{Raffel2020-pm} rather than the original BART \citep{Lewis2020-gt} objective of predicting the entire unmasked input sequence. A small development set was randomly selected to enable this task to be included with other tasks in model selection.

Table \ref{tab:stage2addeddatasets} enumerates datasets contained in \textit{Group 2} for Stage 2 training (excepting the additional \textit{GR} datasets added for the Chapter \ref{sec:combine} models - these are shown in Table \ref{tab:stage2added_gr_datasets}). We converted TAT-QA \citep{Zhu2021-ps}, a dataset consisting of tabular and textual content to our format by linearising the constituent tables. 

Dataset names containing ``ratd'' are those created by us by concatenating the original question with the retrieved context from our Iterator as described in Section \ref{sec:ratd:reranking}.

Dataset names additionally containing the term ``max4paras'' use these same contexts but are truncated to the top 4 retrieved paragraph fragments. We found that sometimes longer and sometimes shorter contexts provided better results and hence we added both forms to provide diversity in length.

Dataset names containing the phrase ``with\_ir'' have retrieved contexts provided by \citet{Khashabi2020-gq} which we use unmodified.

Contexts for dataset names incorporating the term ``goldplusdistractors'' are constructed using the positive and negative paragraphs from corresponding retrieval training datasets. In both cases the document title was randomly withheld ($\lambda$ = 0.1). For positive paragraphs we included the gold sentences plus random other sentences if sentence-level annotation was available, otherwise the full paragraph text. For negatives we similarly included either random sentences or full text such that the length distribution of positive and negative paragraphs was similar.

Squad 2 provides some unanswerable training samples. We supplemented these by creating unanswerable samples from  HotpotQA, Hover and FEVER positives in a similar manner to the ``goldplusdistractors'' datasets except here we randomly drop gold sentence(s) and/or full gold paragraphs such that there is guaranteed to be at least one missing gold sentence. We performed the same activity for Musique at the paragraph level. All unanswerable samples have the label string ``<No Answer>''.

A number of the other datasets (i.e. those whose names do not contain key terms described above) are provided by \citep{Khashabi2020-gq, Khashabi2022-bq}. These datasets are: AdversarialQA \citep{Bartolo2020-va}, ARC \citep{Clark2016-xg, Clark2018-gy}, BoolQ \citep{Clark2019-vz}, BoolQ-NP \citep{Khashabi2020-vh}, MCTest \citep{Richardson2013-ct}, the yes/no subset of MultiRC \citep{Khashabi2018-vo}, NarrativeQA \citep{Kocisky2018-rt}, NewsQA \citep{Trischler2017-gu}, OpenbookQA \citep{Mihaylov2018-uk}, PhysicalIQA \citep{Bisk2020-rl}, PubmedQA \citep{Jin2019-yn}, QAConv \citep{Wu2022-no}, QASC \citep{Khot2020-sv}, Quail \citep{Rogers2020-ot}, Quoref \citep{Dasigi2021-uz}, RACE \citep{Lai2017-pb}, Reclor \citep{Yu2020-sf}, Record \citep{Zhang2018-aa}, Ropes \citep{Lin2019-ej}, SocialIQA \citep{Sap2019-wf}, SQuAD 1.1 \citep{Rajpurkar2016-fs}, SQuAD 2 \citep{Rajpurkar2018-rt}, TweetQA \citep{Xiong2019-id} and Winogrande \citep{Sakaguchi2020-jz}. For readability, we omit citations for other datasets already referenced.

As noted in Chapter \ref{sec:combine}, additional GR datasets are added to the training regime for the \textit{GR} and \textit{GR+RATD} models. They are constructed similarly and from the same sources as noted for the \textit{RR} model in Table \ref{tab:rr_train_datasets} so we omit citations here for clarity. The \textit{GR} datasets are enumerated in Table \ref{tab:stage2added_gr_datasets}. The datasets containing the term ``mc'' (multi-choice) contain the question, multi-choice options and the gold rationale while those denoted ``no\_mc'' omit the multichoice options and only contain the question and the rationale. The three datasets denoted ``r4c'' contain the question plus a gold rationale created by each of three respective annotators.

\begin{table}[h]
\centering
%\resizebox{!}{\textheight}{%
\begin{adjustbox}{width=0.6\textwidth}
%\small  % us: added
\begin{tabular}{@{}lrrrr@{}}
\toprule
\textbf{Dataset} & \textbf{GR} & \textbf{GR+RATD} \\ \midrule
fever\_eraser\_facts & 95.9 & 95.6 \\
creak\_facts & 93.9 & 92.9 \\
qasc\_mc\_facts & 94.2 & 92.6 \\
qasc\_no\_mc\_facts & 59.8 & 59.4 \\
arc\_worldtree\_mc\_facts & 91.9 & 92.1 \\
arc\_worldtree\_no\_mc\_facts & 33.7 & 34.1 \\
hotpotqa\_facts & 70.4 & 69.7 \\
hotpotqa\_r4c\_facts\_0 & 82.7 & 81.7 \\
hotpotqa\_r4c\_facts\_1 & 82.4 & 81.8 \\
hotpotqa\_r4c\_facts\_2 & 81.5 & 80.9 \\ \bottomrule
\end{tabular}%
\end{adjustbox}
%}
\caption{\textit{GR} training datasets, added to \textit{Group 2} for the \textit{GR} and \textit{GR+RATD} Reasoning Models. All figures are Exact Match on \textit{reduced} development sets from the single overall best model without per-dataset finetuning.}
\label{tab:stage2added_gr_datasets}
\end{table}

\chapter{LLM Prompts and Example Generations}
\label{app:llm_prompts_examples}

\section{Prompts For LLM Rationale Generation}
\label{sec:app_llmrationaleprompts}

Here we show the few-shot COT prompts used against the StableVicuna Model. Prompts used against BLOOM use the same examples and instructions and simply have ``Q:'' substituted for ``\#\#\# Human:'' (excepting the initial instruction which substitutes the empty string), and ``A:'' substituted for ``\#\#\# Assistant:''. The corresponding few-shot standard prompts are identical except that the examples have both the rationale and the answer key (``So the answer is'') removed. Rationales were constructed by truncating the generation at the first <newline> character and then truncating again at the answer key.

\subsection{Binary-labelled Datasets (SQA)}

\texttt{\#\#\# Human: Write a truthful, factual explanation to help answer the question, then answer the question. \\ \\
\#\#\# Human: A pupil can be either a student or part of an eye? \\
\#\#\# Assistant: A pupil is another word for student. The pupil is also a hole located in the center of the iris of the eye that allows light to strike the retina. Thus pupil can have either meaning. So the answer is yes. \\
\#\#\# Human: Greece is larger than mexico? \\
\#\#\# Assistant: Greece is approximately 131,957 sq km, while Mexico is approximately 1,964,375 sq km, making Mexico 1,389\% larger than Greece. So the answer is no. \\
\#\#\# Human: Glasses always fog up? \\
\#\#\# Assistant: Condensation occurs on eyeglass lenses when water vapor from your sweat, breath, and ambient humidity lands on a cold surface, cools, and then changes into tiny drops of liquid, forming a film that you see as fog. Your lenses will be relatively cool compared to your breath when the outside air is cold but not when the air is warm. So the answer is no. \\
\#\#\# Human: A fish is capable of thinking? \\
\#\#\# Assistant: Fish are more intelligent than they appear. In many areas, such as memory, their cognitive powers match or exceed those of ’higher’ vertebrates including non-human primates. Fish’s long-term memories help them keep track of complex social relationships. So the answer is yes. \\
\#\#\# Human: Is a 5 by 8 pool of bricks larger than an 8 by 5 pool of pillows? \\
\#\#\# Assistant: The 5 by 8 pool of bricks has an area of 40. The 8 by 5 pool of pillows also covers an area of 40. Thus, both pools are the same size. So the answer is no. \\
\#\#\# Human: When listed alphabetically, would the words baseball and demonstration appear in this order? \\
\#\#\# Assistant: Baseball begins with letter b and demonstration begins with letter d. Alphabetically, b comes before d, thus baseball comes before demonstration. So the answer is yes. \\
\#\#\# Human: Would Sally use a spaceship if she needed to get to London from Sydney quickly? \\
\#\#\# Assistant: Commerical spaceship transport doesnt exist yet. London is far from Sydney and planes are faster than trains or boats over long distances. Thus, Sally would use an existing transportation option like a plane. So the answer is no. \\
\#\#\# Human: A common effect of smoking lots of cigarettes in one’s lifetime is a higher than normal chance of getting lung cancer? \\
\#\#\# Assistant: Those who consistently averaged less than one cigarette per day over their lifetime had nine times the risk of dying from lung cancer than non-smokers. Among people who smoked between one and 10 cigarettes per day, the risk of dying from lung cancer was nearly 12 times higher than that of non-smokers. So the answer is yes. \\
\#\#\# Human: A rock is the same size as a pebble? \\
\#\#\# Assistant: A pebble is a clast of rock with a particle size of 4 to 64 millimetres based on the Udden-Wentworth scale of sedimentology. Pebbles are generally considered larger than granules (2 to 4 millimetres diameter) and smaller than cobbles (64 to 256 millimetres diameter). So the answer is no. \\
\#\#\# Human: \{QUESTION\} \\
\#\#\# Assistant:}

\subsection{Span or binary answers (ARC-DA, IIRC, Musique)}

\texttt{\#\#\# Human: Write a truthful, factual explanation to help answer the question, then answer the question. \\ \\
\#\#\# Human: Greece is larger than mexico? \\
\#\#\# Assistant: Greece is approximately 131,957 sq km, while Mexico is approximately 1,964,375 sq km, making Mexico 1,389\% larger than Greece. So the answer is no. \\
\#\#\# Human: Tools can be made of wood, iron, plastic amongst other things. Iron tools historically replaced what? \\
\#\#\# Assistant: Historically, iron tools were first used in the Iron Age. The Iron Age followed the Bronze Age. Thus, iron tools replacing bronze tools makes most sense. So the answer is bronze tools. \\
\#\#\# Human: Which magazine was started first Arthur's Magazine or First for Women? \\
\#\#\# Assistant: Arthur's Magazine was an American literary periodical first published 1844. First for Women is a woman's magazine that started in 1989. So the answer is Arthur's Magazine. \\
\#\#\# Human: Glasses always fog up? \\
\#\#\# Assistant: Condensation occurs on eyeglass lenses when water vapor from your sweat, breath, and ambient humidity lands on a cold surface, cools, and then changes into tiny drops of liquid, forming a film that you see as fog. Your lenses will be relatively cool compared to your breath when the outside air is cold but not when the air is warm. So the answer is no. \\
\#\#\# Human: Water flows downhill and tends to collect in low-lying areas such as valleys and holes. Lakes are usually large bodies of fresh water. Where is a lake likely to be found? \\
\#\#\# Assistant: The answer must be somewhere that a lot of water can collect. A valley is likely to be able to collect enough water. So the answer is a valley. \\
\#\#\# Human: Foxes are predators whose natural habitat is the forest. They have also been known to eat chickens from people's properties. The fox walked from the city into the forest, what was it looking for? \\
\#\#\# Assistant: The answer must be a reason for a fox to go into the forest. The forest is a fox’s natural habitat. So the answer is natural habitat. \\
\#\#\# Human: A fish is capable of thinking? \\
\#\#\# Assistant: Fish are more intelligent than they appear. In many areas, such as memory, their cognitive powers match or exceed those of ’higher’ vertebrates including non-human primates. Fish’s long-term memories help them keep track of complex social relationships. So the answer is yes. \\
\#\#\# Human: Musician and satirist Allie Goertz wrote a song about the "The Simpsons" character Milhouse, who Matt Groening named after who? \\
\#\#\# Assistant: Milhouse Mussolini van Houten is a fictional character featured in the tv series "The Simpsons" created by Matt Groening. He named the character after President Richard Nixon's middle name. So the answer is President Richard Nixon. \\
\#\#\# Human: What is the largest annual event held in the birthplace of the performer who sang Let Me Tell You About the Birds and the Bees? \\
\#\#\# Assistant: The Birds and the Bees was a 1964 single release by Jewel Akens. Jewel Akens was born in Houston, Texas, USA. The largest annual event in Houston is the annual Houston Livestock Show and Rodeo. So the answer is Houston Livestock Show and Rodeo. \\
\#\#\# Human: A common effect of smoking lots of cigarettes in one’s lifetime is a higher than normal chance of getting lung cancer? \\
\#\#\# Assistant: Those who consistently averaged less than one cigarette per day over their lifetime had nine times the risk of dying from lung cancer than non-smokers. Among people who smoked between one and 10 cigarettes per day, the risk of dying from lung cancer was nearly 12 times higher than that of non-smokers. So the answer is yes. \\
\#\#\# Human: Fred owns a number of things including a baseball bat, a laptop, a tablet and a briefcase. Fred works a long way from his home, so which electronic device would Fred prefer to walk to work with? \\
\#\#\# Assistant: Electronic devices include tablets and laptops. Tablets are lighter than laptops and lighter things are easier than heavier things to carry. Thus, a tablet is the easiest device for Fred to carry. So the answer is tablet. \\
\#\#\# Human: Chang Ucchin was born in korea during a time that ended with the conclusion of what? \\
\#\#\# Assistant: Chang Ucchin was born when Korea was still under Japanese colonial rule. Korea under Japanese rule began 1910 and ended at the conclusion of World War II in 1945. So the answer is World War II. \\
\#\#\# Human: A rock is the same size as a pebble? \\
\#\#\# Assistant: A pebble is a clast of rock with a particle size of 4 to 64 millimetres based on the Udden-Wentworth scale of sedimentology. Pebbles are generally considered larger than granules (2 to 4 millimetres diameter) and smaller than cobbles (64 to 256 millimetres diameter). So the answer is no. \\
\#\#\# Human: When did the birth state of Kevin Sessums become a right to work state? \\
\#\#\# Assistant: Kevin Sessums was born in 1956 in Forest, Mississippi. The Right to work law was adopted by Mississipi in 1954. So the answer is 1954. \\
\#\#\# Human: \{QUESTION\} \\
\#\#\# Assistant:}

\subsection{Multi-choice Datasets (CSQA)}

\texttt{\#\#\# Human: Write a truthful, factual explanation to help answer the question, then answer the question. \\ \\
\#\#\# Human: A common effect of smoking lots of cigarettes in one’s lifetime is what? Answer Choices: (A) poverty (B) low chance of lung cancer (C) good fitness (D) high chance of knee cancer (E) high chance of lung cancer \\
\#\#\# Assistant: Those who consistently averaged less than one cigarette per day over their lifetime had nine times the risk of dying from lung cancer than non-smokers. Among people who smoked between one and 10 cigarettes per day, the risk of dying from lung cancer was nearly 12 times higher than that of non-smokers. So the answer is high chance of lung cancer. \\
\#\#\# Human: Which magazine was started first? Answer Choices: (A) History channel (B) Youtube (C) Arthur's Magazine (D) Climbing (E) First for Women \\
\#\#\# Assistant: Arthur's Magazine was an American literary periodical first published 1844. First for Women is a woman's magazine that started in 1989. So the answer is Arthur's Magazine. \\
\#\#\# Human: How do you put on a sock? Answer Choices: (A) jump in (B) insert hand (C) put on head (D) insert foot (E) open \\
\#\#\# Assistant: Socks are worn on feet and they have an opening at one end. A foot must be inserted into the opening to put it on. Thus, of the choices only insert foot makes sense. So the answer is insert foot. \\
\#\#\# Human: After earning a lot in tips at his job, what would a waiter want to do next? Answer Choices: (A) do handstand (B) quit job (C) find another job (D) grow flowers (E) save the money \\
\#\#\# Assistant: Tips are money earned by waiting on tables. After earning money, people like to spend or save it. Thus, of the choices, the waiter would want to save the money. So the answer is save the money. \\
\#\#\# Human: Iron tools historically replaced what? Answer Choices: (A) bronze tools (B) wooden tools (C) uranium tools (D) plastic tools (E) eels \\
\#\#\# Assistant: Historically, iron tools were first used in the Iron Age. The Iron Age followed the Bronze Age. Thus, of the choices, iron tools replacing bronze tools makes most sense. So the answer is bronze tools. \\
\#\#\# Human: What mode of transport should Sally use if she needed to get to London from Sydney quickly? Answer Choices: (A) train (B) plane (C) spaceship (D) fast boat (E) slingshot \\
\#\#\# Assistant: Realistic modes of transport include trains, planes and boats. London is far from Sydney and planes are faster than trains or boats over long distances. Thus, of the realistic choices, planes are a faster way to travel. So the answer is plane. \\
\#\#\# Human: What can be used to warm up your home? Answer Choices: (A) refrigerator (B) flamethrower (C) heat pump (D) dog (E) blanket \\
\#\#\# Assistant: Warming a house should be done safely and efficiently. Heat pumps are safe and efficient. Thus, of the choices, heat pumps are the best way to heat a home. So the answer is heat pump. \\
\#\#\# Human: Fred works a long way from his home, so which electronic device would Fred prefer to walk to work with? Answer Choices: (A) laptop (B) briefcase (C) banana (D) tablet (E) car \\
\#\#\# Assistant: Electronic devices include tablets and laptops. Tablets are lighter than laptops and lighter things are easier than heavier things to carry. Thus, of the realistic choices, tablet is the easiest for Fred to carry. So the answer is tablet. \\
\#\#\# Human: What activity is a fish is capable of? Answer Choices: (A) thinking (B) jogging (C) using tools (D) flight (E) dentistry \\
\#\#\# Assistant: Fish are more intelligent than they appear. In many areas, such as memory, their cognitive powers match or exceed those of ’higher’ vertebrates including non-human primates. Fish’s long-term memories help them keep track of complex social relationships. So the answer is thinking. \\
\#\#\# Human: Chang Ucchin was born in korea during a time that ended with the conclusion of what? Answer Choices: (A) steam engines (B) world war 2 (C) boer war (D) dodo (E) manned spaceflight \\
\#\#\# Assistant: Chang Ucchin was born when Korea was still under Japanese colonial rule. Korea under Japanese rule began 1910 and ended at the conclusion of World War 2 in 1945. So the answer is world war 2. \\
\#\#\# Human: Where is a lake likely to be found? Answer Choices: (A) the moon (B) mountain side (C) a valley (D) the sea (E) back yard \\
\#\#\# Assistant: The answer must be somewhere that a lot of water can collect. Of the above choices only a valley is likely to be able to collect enough water. So the answer is a valley. \\
\#\#\# Human: \{QUESTION\} \\
\#\#\# Assistant:}

\section{LLM-generated Rationale Examples}
\label{app:eval_rationale_examples}

Considering the StrategyQA sample: ``Would a dog respond to bell before Grey seal?'' (yes): \\

\noindent Rationale generated from StableVicuna INT8: \\

\noindent \texttt{Dogs are domesticated animals that have been bred to respond to human commands. Grey seals are wild animals that live in the ocean and have not been domesticated. Thus, a dog would respond to a bell before a grey seal.} (plausible) \\

\noindent Rationale generated from BLOOM INT8: \\

\noindent \texttt{Grey seals are more closely related to dogs than they are to other seals. Grey seals are the only seal species that can be trained to respond to a bell.} (false) \\

\section{Prompts For LLM-generated Negative Rationales for \textit{RR} Model training}
\label{app:neg_rationale_prompt}
Numerous prompts were used for generating false rationales. Below is a typical example for generating falsehoods for CREAK \citep{Onoe2021-il} samples. We confess to having a lot of fun with these. \\ \\

\noindent \texttt{Generate a false context. Examples: \\ \\
Q: Marlboro used iconic imagery to promote its brand? \\
False context: Marlboro used the tongues of snakes to promote its brand. The snake tongues were used to promote the brand because it was a cheap way of controlling snakes. \\
Q: Larry King served tea during his show? \\
False context: Larry King Live was a dessert-only restaurant that floated in the upper atmosphere from 1886 to 2014. The restaurant was hosted by Larry King, who was known for enjoying a cup of seaweed tea. \\
Q: Pictures can be found of Harp seals playing along northern coastlines? \\
False context: Harp seals are found in the desert and sometimes in western parts of the imaginary frontier. Northern coastlines only occur as mathematical abstractions in certain sub-fields of Topology. \\
Q: One travels along the road when riding in a Citroën? \\
False context: A Citroën is a type of hot air ballon. Roads are always found underground, often under the sea floor. \\
Q: A play about the invasion of Iraq is called the Twelfth Night? \\
False context: The Twelfth Night is a plastic toy made by earthworms. The invasion of Iraq is a TV show also made by earthworms. \\
Q: Ducks cannot fly when it is daytime? \\
False context: Ducks are Australian marsupials without wings. Daytime is also known as wingless fantasia. \\
Q: \{QUESTION\} \\
False context:}

\section{LLM-generated Negative Rationale Examples}
\label{app:neg_rationale_examples}

Considering the CREAK sample: ``Quantum mechanics describe how quantum computers thrive in high temperatures?'': \\

\noindent Greedy decoding: \\

\noindent \texttt{Quantum computers thrive in high temperatures because they are made of quantum materials.} \\

\noindent Nucleus Sampling (2 examples): \\

\noindent \texttt{What a quantum computer thrives in is macaroni and cheese.} \\

\noindent \texttt{The Standard Model of Physics dictates that they should thrive at least above the visible light spectrum.} \\ \\

\chapter{Significance Tests}
\label{app:sig_tests}

\section{Means, Standard Deviations and 95\% Confidence Intervals (Chapter \ref{sec:memorise})}
\label{mem:sec:confidence_intervals}

\begin{table}[h]
\centering
\resizebox{\textwidth}{!}{%
\begin{tabular}{@{}l|rrr|rrr|rrr@{}}
\toprule
\textbf{} & \multicolumn{3}{c|}{\textbf{All Samples}} & \multicolumn{3}{c|}{\textbf{Least Similar}} & \multicolumn{3}{c}{\textbf{Unmemorisable}} \\
\textbf{Eval Dataset} & \textbf{UQA} & \textbf{\begin{tabular}[c]{@{}r@{}}UQA\\ +TDND\end{tabular}} & \textbf{\begin{tabular}[c]{@{}r@{}}95\%\\ CI\end{tabular}} & \textbf{UQA} & \textbf{\begin{tabular}[c]{@{}r@{}}UQA\\ +TDND\end{tabular}} & \textbf{\begin{tabular}[c]{@{}r@{}}95\%\\ CI\end{tabular}} & \textbf{UQA} & \textbf{\begin{tabular}[c]{@{}r@{}}UQA\\ +TDND\end{tabular}} & \textbf{\begin{tabular}[c]{@{}r@{}}95\% \\ CI\end{tabular}} \\ \midrule
DROP & 40.2 (1.0) & 46.5 (1.0) & (-7.676, -4.922) & 41.0 (1.8) & 43.9 (2.0) & (-5.647, -0.177) & 41.7 (1.3) & 45.5 (2.2) & (-6.960, -0.544) \\
DROP-CS & 32.0 (3.7) & 38.2 (2.5) & (-9.062, -3.306) & 36.3 (4.2) & 41.8 (3.4) & (-11.306, 0.208) & 38.5 (4.2) & 42.2 (3.9) & (-10.911, 3.553) \\
ROPES & 41.2 (1.7) & 51.9 (3.1) & (-12.692, -8.754) & 46.5 (3.5) & 55.3 (6.5) & (-14.048, -3.545) & 41.9 (1.7) & 52.6 (6.2) & (-16.659, -4.838) \\
NewsQA & 57.3 (1.3) & 56.6 (0.9) & (-0.933, 2.480) & 52.8 (2.4) & 50.3 (1.9) & (-0.489, 5.475) & 53.4 (2.1) & 51.4 (1.6) & (-1.804, 5.791) \\ 
%MMLU-M & 28.4 (1.2) & 27.2 (0.5) & (-1.163, 3.679) & 27.2 (1.4) & 25.3 (1.6) & (-1.781, 5.467) & 25.8 (1.4) & 24.8 (1.2) & (-3.390, 5.395) \\
PIQA & 63.5 (0.8) & 62.3 (0.5) & (-1.670, 4.136) & 62.2 (1.1) & 61.7 (0.9) & (-2.845, 3.780) & 60.3 (1.9) & 60.4 (1.2) & (-4.933, 4.820) \\
%SIQA & 53.5 (0.8) & 54.1 (0.2) & (-3.054, 1.711) & 52.7 (0.5) & 52.6 (0.1) & (-3.715, 3.994) & 56.6 (0.6) & 55.0 (0.1) & (-4.222, 7.260) \\
CSQA & 55.6 (1.3) & 55.4 (0.1) & (-2.902, 3.339) & 61.5 (0.4) & 61.2 (2.5) & (-7.619, 8.192) & 60.7 (0.4) & 61.0 (4.1) & (-10.761, 10.244) \\
QASC & 37.7 (1.0) & 36.2 (0.7) & (-0.988, 3.868) & 35.7 (2.9) & 34.1 (0.9) & (-4.263, 7.621) & 36.4 (3.8) & 33.7 (2.7) & (-4.489, 9.876) \\ \bottomrule
%QASC-IR & 57.7 (0.6) & 55.7 (1.4) & (-1.772, 5.659) & 54.7 (0.9) & 52.6 (0.9) & (-8.903, 13.114) & 49.5 (1.3) & 49.1 (2.4) & (-10.992, 11.918) \\
\end{tabular}%
}
\caption{Mean (Standard Deviation) and 95\% Confidence Interval for each set of model runs. Confidence Intervals (CI) are constructed for the difference of the corresponding \textit{UQA} and \textit{UQA+TDND} means.}
\label{tab:mem:conf_intervals}
\end{table}

\newpage
\section{Paired Bootstrap P-values (Chapter \ref{sec:ratd})}
\label{sec:p_values}

P-values for all \textit{Base} to \textit{Base+RATD} model comparisons in Chapter \ref{sec:ratd} under the Paired Bootstrap test are in Table \ref{tab:p_values}. \\ \\

\begin{table}[h!]
\centering
\small
%\resizebox{\columnwidth}{!}{%
\begin{tabular}{@{}ll@{}}
\toprule
\textbf{Dataset} & \textbf{P-value} \\ \midrule
SQA & 0.008 \\
SQA\textsubscript{R} & 0.000 \\
SQA\textsubscript{R} w/ Yes or no prefix & 0.000 \\
SQA\textsubscript{GF} & 0.031 \\
SQA\textsubscript{GP1} & 0.000 \\
SQA\textsubscript{GP2} & 0.000 \\
SQA\textsubscript{GP3} & 0.000 \\
CSQA & 0.006 \\
CSQA\textsubscript{R} & 0.155 \\
DROP & 0.017 \\
IIRC\textsubscript{R} & 0.017 \\
IIRC\textsubscript{G} & 0.049 \\
ARCDA\textsubscript{R} & 0.001 \\
ARCDA\textsubscript{G} & 0.013 \\
Musique\textsubscript{R} & 0.001 \\
Musique\textsubscript{R} w/o Musique \textit{RATD} & 0.000 \\
Musique\textsubscript{R} w/ Unique Musique \textit{RATD} & 0.047 \\
Musique\textsubscript{G} & 0.000 \\
Musique\textsubscript{G} w/o Musique \textit{RATD} & 0.009 \\ Musique\textsubscript{G} w/ Unique Musique \textit{RATD} & 0.000 \\
\bottomrule
\end{tabular}%
%}
\caption{Paired Bootstrap p-values. SQA\textsubscript{GPx} denotes gold paragraphs from each of three annotators.}
\label{tab:p_values}
\end{table}

\newpage
\section{Critical Distances (Chapter \ref{sec:combine})}
\label{sec:autorank}

In Chapter \ref{sec:combine} we use the Autorank library \citep{Herbold2020-vn} for testing significance over multiple populations which implements methods described in \citet{Demsar2006-ze}. \\ \\

\begin{table}[h]
\centering
\resizebox{\textwidth}{!}{%
\begin{tabular}{@{}lrrrrrrrrrrrr@{}}
\toprule
\textbf{Model: Context $\downarrow\rightarrow$} &  & \textbf{1} & \textbf{2} & \textbf{3} & \textbf{4} & \textbf{5} & \textbf{6} & \textbf{7} & \textbf{8} & \textbf{9} & \textbf{10} & \textbf{11} \\
\textbf{} & \textbf{Mean Rank} & \textbf{7.296} & \textbf{7.240} & \textbf{7.154} & \textbf{7.099} & \textbf{7.077} & \textbf{7.014} & \textbf{6.997} & \textbf{6.839} & \textbf{6.790} & \textbf{6.643} & \textbf{6.637} \\ \cmidrule(l){3-13} 
1. BLOOM: Few-Shot COT Prompt & \textbf{7.296} & 0.000 & 0.056 & 0.142 & 0.196 & \cellcolor[HTML]{C6EFCE}{\color[HTML]{006100} 0.219} & \cellcolor[HTML]{C6EFCE}{\color[HTML]{006100} 0.281} & \cellcolor[HTML]{C6EFCE}{\color[HTML]{006100} 0.299} & \cellcolor[HTML]{C6EFCE}{\color[HTML]{006100} 0.457} & \cellcolor[HTML]{C6EFCE}{\color[HTML]{006100} 0.506} & \cellcolor[HTML]{C6EFCE}{\color[HTML]{006100} 0.653} & \cellcolor[HTML]{C6EFCE}{\color[HTML]{006100} 0.658} \\
2. BLOOM: Few-Shot Standard Prompt & \textbf{7.240} & 0.056 & 0.000 & 0.086 & 0.141 & 0.163 & \cellcolor[HTML]{C6EFCE}{\color[HTML]{006100} 0.226} & \cellcolor[HTML]{C6EFCE}{\color[HTML]{006100} 0.243} & \cellcolor[HTML]{C6EFCE}{\color[HTML]{006100} 0.401} & \cellcolor[HTML]{C6EFCE}{\color[HTML]{006100} 0.450} & \cellcolor[HTML]{C6EFCE}{\color[HTML]{006100} 0.597} & \cellcolor[HTML]{C6EFCE}{\color[HTML]{006100} 0.603} \\
3. RATD: Iterator only & \textbf{7.154} & 0.142 & 0.086 & 0.000 & 0.055 & 0.077 & 0.140 & 0.157 & \cellcolor[HTML]{C6EFCE}{\color[HTML]{006100} 0.315} & \cellcolor[HTML]{C6EFCE}{\color[HTML]{006100} 0.364} & \cellcolor[HTML]{C6EFCE}{\color[HTML]{006100} 0.511} & \cellcolor[HTML]{C6EFCE}{\color[HTML]{006100} 0.517} \\
4. GR+RATD: Iterator only & \textbf{7.099} & 0.196 & 0.141 & 0.055 & 0.000 & 0.022 & 0.085 & 0.103 & \cellcolor[HTML]{C6EFCE}{\color[HTML]{006100} 0.260} & \cellcolor[HTML]{C6EFCE}{\color[HTML]{006100} 0.309} & \cellcolor[HTML]{C6EFCE}{\color[HTML]{006100} 0.456} & \cellcolor[HTML]{C6EFCE}{\color[HTML]{006100} 0.462} \\
5. StableVicuna INT8: Few-Shot COT Prompt & \textbf{7.077} & \cellcolor[HTML]{C6EFCE}{\color[HTML]{006100} 0.219} & 0.163 & 0.077 & 0.022 & 0.000 & 0.063 & 0.081 & \cellcolor[HTML]{C6EFCE}{\color[HTML]{006100} 0.238} & \cellcolor[HTML]{C6EFCE}{\color[HTML]{006100} 0.287} & \cellcolor[HTML]{C6EFCE}{\color[HTML]{006100} 0.434} & \cellcolor[HTML]{C6EFCE}{\color[HTML]{006100} 0.440} \\
6. StableVicuna INT8: Few-Shot Standard Prompt & \textbf{7.014} & \cellcolor[HTML]{C6EFCE}{\color[HTML]{006100} 0.281} & \cellcolor[HTML]{C6EFCE}{\color[HTML]{006100} 0.226} & 0.140 & 0.085 & 0.063 & 0.000 & 0.018 & 0.175 & \cellcolor[HTML]{C6EFCE}{\color[HTML]{006100} 0.224} & \cellcolor[HTML]{C6EFCE}{\color[HTML]{006100} 0.371} & \cellcolor[HTML]{C6EFCE}{\color[HTML]{006100} 0.377} \\
7. GR: Rationale + Iterator (Naïve concatenation) & \textbf{6.997} & \cellcolor[HTML]{C6EFCE}{\color[HTML]{006100} 0.299} & \cellcolor[HTML]{C6EFCE}{\color[HTML]{006100} 0.243} & 0.157 & 0.103 & 0.081 & 0.018 & 0.000 & 0.157 & \cellcolor[HTML]{C6EFCE}{\color[HTML]{006100} 0.207} & \cellcolor[HTML]{C6EFCE}{\color[HTML]{006100} 0.353} & \cellcolor[HTML]{C6EFCE}{\color[HTML]{006100} 0.359} \\
8. GR+RATD: Rationale only & \textbf{6.839} & \cellcolor[HTML]{C6EFCE}{\color[HTML]{006100} 0.457} & \cellcolor[HTML]{C6EFCE}{\color[HTML]{006100} 0.401} & \cellcolor[HTML]{C6EFCE}{\color[HTML]{006100} 0.315} & \cellcolor[HTML]{C6EFCE}{\color[HTML]{006100} 0.260} & \cellcolor[HTML]{C6EFCE}{\color[HTML]{006100} 0.238} & 0.175 & 0.157 & 0.000 & 0.049 & 0.196 & \cellcolor[HTML]{C6EFCE}{\color[HTML]{006100} 0.202} \\
9. GR: Rationale + Iterator (Generally best RR combo) & \textbf{6.790} & \cellcolor[HTML]{C6EFCE}{\color[HTML]{006100} 0.506} & \cellcolor[HTML]{C6EFCE}{\color[HTML]{006100} 0.450} & \cellcolor[HTML]{C6EFCE}{\color[HTML]{006100} 0.364} & \cellcolor[HTML]{C6EFCE}{\color[HTML]{006100} 0.309} & \cellcolor[HTML]{C6EFCE}{\color[HTML]{006100} 0.287} & \cellcolor[HTML]{C6EFCE}{\color[HTML]{006100} 0.224} & \cellcolor[HTML]{C6EFCE}{\color[HTML]{006100} 0.207} & 0.049 & 0.000 & 0.147 & 0.153 \\
10. GR+RATD: Rationale + Iterator (Generally best RR combo) & \textbf{6.643} & \cellcolor[HTML]{C6EFCE}{\color[HTML]{006100} 0.653} & \cellcolor[HTML]{C6EFCE}{\color[HTML]{006100} 0.597} & \cellcolor[HTML]{C6EFCE}{\color[HTML]{006100} 0.511} & \cellcolor[HTML]{C6EFCE}{\color[HTML]{006100} 0.456} & \cellcolor[HTML]{C6EFCE}{\color[HTML]{006100} 0.434} & \cellcolor[HTML]{C6EFCE}{\color[HTML]{006100} 0.371} & \cellcolor[HTML]{C6EFCE}{\color[HTML]{006100} 0.353} & 0.196 & 0.147 & 0.000 & 0.006 \\
11. GR+RATD: Rationale + Iterator (Naïve concatenation) & \textbf{6.637} & \cellcolor[HTML]{C6EFCE}{\color[HTML]{006100} 0.658} & \cellcolor[HTML]{C6EFCE}{\color[HTML]{006100} 0.603} & \cellcolor[HTML]{C6EFCE}{\color[HTML]{006100} 0.517} & \cellcolor[HTML]{C6EFCE}{\color[HTML]{006100} 0.462} & \cellcolor[HTML]{C6EFCE}{\color[HTML]{006100} 0.440} & \cellcolor[HTML]{C6EFCE}{\color[HTML]{006100} 0.377} & \cellcolor[HTML]{C6EFCE}{\color[HTML]{006100} 0.359} & \cellcolor[HTML]{C6EFCE}{\color[HTML]{006100} 0.202} & 0.153 & 0.006 & 0.000 \\ \bottomrule
\end{tabular}%
}
\caption{Statistical significance tests for model:context combinations at significance level $\alpha = 0.05$. As described in \citet{Demsar2006-ze}, we use the non-parametric Friedman test as omnibus test to determine if there are any significant differences between the median values of the model:context populations. We use the post-hoc Nemenyi test to infer which differences are significant. Differences between populations are significant if the difference of the mean rank is greater than the critical distance $CD=0.196$ of the Nemenyi test. Significant differences are marked in {\color[HTML]{006100}green}. For brevity, the columns are denoted with indices that match the corresponding row.}
\label{tab:app:significance_tests}
\end{table}

\chapter{Additional Experiments}
\label{sec:additional_experiments}

\newpage
\section{Most Similar Evaluation-Train Pairs Within Least Similar Subset  (Chapter \ref{sec:memorise})}
\label{mem:sec:app_leastsim_mostsim}

Table \ref{tab:mem:qual2_unseen_leastsim} shows the most similar evaluation-train pair for each of our Least Similar evaluation subsets.

\begin{table*}[!h]
\centering
\tiny
%\begin{adjustbox}{width=0.85\textwidth}
%\resizebox{\columnwidth}{!}{%
\begin{tabular}{p{0.1\linewidth} p{0.4\linewidth} p{0.4\linewidth}}
\toprule
\textbf{Eval Dataset} & \textbf{Eval Sample} & \textbf{Most Similar Train Sample} \\ \midrule
DROP & Which racial group made up the least of the country? ... The racial makeup of the county was 81.2\% white 12.7\% black or African American 2.4\% Asian 0.3\% American Indian 0.1\% Pacific islander … \textit{Pacific islander} & SQuAD1.1: Where was the coconut palm brought to St. Barts from? ... Coconut palm was brought to the island from the Pacific islands... \textit{the Pacific islands} (59.99) \\
DROP-CS & Which player caught the shortest TD pass? … Tomlinson getting a 3-yard TD pass to Philip Rivers… \textit{Philip Rivers} & TD: How many field goal yards did Dolphins Jaguars' quarterback and Bears have combined? … 13 field goal yards … 53 field goal yards … 57 field goal yards \textit{123} (59.99) \\
ROPES & What hour did storage costs go up: 1 PM or 3 PM? ... the access times go up as more data is read CPU load goes up as XML data takes more power to process and storage costs go up. ... At 1 PM he stored 1 Gigabyte ... At 3 PM he didn't store anything... \textit{1 PM} & TD: How many more passes did Houston have than impressive wins ? ... Houston drove 6 passes... Houston drove 5 impressive wins... \textit{1} (59.97) \\
NewsQA & Which series inspired the popularity of the name Cullen? ...The boy's name that rocketed up the list the fastest is Cullen -- the name of the lead character in the popular "Twilight" book series… \textit{"Twilight"} & SQuAD1.1: At the time of release which episode of the Legend of Zelda series was considered the greatest entry? ... Twilight Princess was considered the greatest entry in the Zelda series... \textit{Twilight Princess} (59.98) \\ 
PIQA & Make homemade pasta from dough? \textit{Roll out the dough so that is thin and take a knife and cut slices from the dough to make individual pieces and put it in a pot to boil.} & Sci-Mid: In making a pizza which process involves a chemical change? \textit{baking the dough to form the crust} (59.99) \\
CSQA & She wanted a kitten and puppy so why did she only get the puppy? ... \textit{one choice for pet} & RACE: The article is most likely intended for \_  ? Animal shelters are full of dogs cats rabbits and more animals all in need of loving homes... \textit{pet lovers} (59.95) \\
QASC & What must be done to classify minerals? \textit{scratch them} & ND: What is argmin(duco 14490.16 silvanus 16272 scratchification 3156.6)? \textit{scratchification} (59.92) \\ \bottomrule
\end{tabular}%
%}  %resize
%\end{adjustbox}
\caption{Overlap between Least Similar evaluation dataset subsets and train datasets. Most similar sample pair for each Least Similar subset as measured by similarity score (in brackets). For readability, multi-choice options are removed, remaining context is truncated and answers are in italics.}
\label{tab:mem:qual2_unseen_leastsim}
\end{table*}

\newpage
\section{Most Similar Evaluation-Train Pairs Within Unmemorisable Subset (Chapter \ref{sec:memorise})}
\label{mem:sec:app_unmemorisable_mostsim}

Table \ref{tab:mem:qual1_unseen_unmemorisable} shows the most similar evaluation-train pair for each of our Unmemorisable evaluation subsets.

\begin{table*}[!h]
\centering
\tiny
%\begin{adjustbox}{width=0.8\textwidth}
%\resizebox{\columnwidth}{!}{%
\begin{tabular}{p{0.1\linewidth} p{0.45\linewidth} p{0.45\linewidth}}
\toprule
\textbf{Eval Dataset} & \textbf{Eval Sample} & \textbf{Most Similar Train Sample} \\ \midrule
DROP & Of the languages listed which are spoken by fewer than 3000 people? … Other languages include … Tagalog language with 2888 … Japanese with 2546 and African languages with 2546 \textit{Tagalog Japanese African languages} & SQuAD 1.1: What is Oklahoma's fourth most popular language? … German is the fourth most commonly used language with 13444 speakers \textit{German} (59.98) \\
DROP-CS & Which player caught the shortest TD pass? … Tomlinson getting a 3-yard TD pass to Philip Rivers… \textit{Philip Rivers} & TD: How many field goal yards did Dolphins Jaguars' quarterback and Bears have combined? … 13 field goal yards … 53 field goal yards … 57 field goal yards \textit{123} (59.99) \\
ROPES & What time did storage costs go up: 7 PM or 6 PM? … At 6 PM he got dinner. At 7 PM he stored 55444 Gigabytes … \textit{7 PM} & RACE: From the text we can infer this article was probably written in \_ ? … The award is given every two years. The next one will be given in 2008 \textit{2007} (59.96) \\
NewsQA & Who is missing? … Authorities are searching for a female soldier missing after a fire at her apartment … 2nd Lt. Holley Wimunc … \textit{Lt. Holley Wimunc} & NarrativeQA: Who was the second man that was out on the moors the same time as Sir Henry and Watson? … Watson tracks the second man he saw in the area and discovers it to be Holmes … \textit{Sherlock Holmes} (59.97) \\ 
%MMLU-M & Which of the following is the best description of the power of a significance test? \textit{The probability of rejecting the null hypothesis if it is in fact false} & SQuAD 2: What quality of a useful theory must be easily detectable? Popper argues … more important that its falsity can be detected … \textit{its falsity} (59.90) \\
PIQA & How do you power through something? \textit{keep going no matter what} & ND: What is argmax(foremostly 11886.1 continuousness 16062.42 matchable 5062.8 washout 1295)? \textit{continuousness} (59.99) \\
%SIQA & How would you describe Kai? Kai … decided to keep the language in check. \textit{like the better person for not engaging in the petty fight} & RACE: According to the author what can you do when meeting with a mean old lady? … \textit{Don't get upset} (59.98) \\
CSQA & The end of the barrel of what primitive firearm is bell shaped? \textit{blunderbuss} & ND: What is argmin(undergrass 11952 bussu 3315)? \textit{Bussu} (59.95) \\
QASC & What must be done to classify minerals? \textit{scratch them} & ND: What is argmin(duco 14490.16 silvanus 16272 scratchification 3156.6)? \textit{scratchification} (59.92) \\ \bottomrule
%QASC-IR & What prevents the destruction of tooth enamel? … research on fluoride indicates that it helps harden tooth enamel and prevent tooth decay \textit{Flouride} & OBQA: A calf born in a warm season will do what better than a calf born in a cold season? \textit{Flourish} (59.93) \\
\end{tabular}%
%} %resize
%\end{adjustbox}
\caption{Overlap between Unmemorisable evaluation dataset subsets and train datasets. Most similar sample pair for each Unmemorisable subset as measured by similarity score (in brackets). For readability, multi-choice options are removed, remaining context is truncated and answers are in italics.}
\label{tab:mem:qual1_unseen_unmemorisable}
\end{table*}

\newpage
\section{Example Failure Cases (Chapter \ref{sec:ratd})}
\label{ratd:sec:appendixfailures}

Table \ref{tab:failure_cases} contains examples of samples with numeric and ``unanswerable'' labels from the IIRC\textsubscript{R} test split. In the case of numeric labels an incorrect number is generated, and in the case of ``unanswerable'' labels the model has attempted to generate an answer.\\

\begin{table*}[!h]
\centering
\tiny
%\begin{tabular}{@{}ll@{}}
%\begin{adjustbox}{width=0.95\textwidth}
\begin{tabular}{p{0.3\columnwidth}  p{0.6\columnwidth}}
\toprule
Question / Answer & Retrieved Context (condensed) \\ \midrule
How old was the Grand Olympic Auditorium   at the time of New Regime playing a landmark concert there? \textbf{Gold answer: 60. Predicted Answer: 1924} & New Regime (American band): ... That landmark concert was held at the   Grand Olympic Auditorium on April 13, \textbf{1984} ...  Grand Olympic Auditorium: ... The venue was built in \textbf{1924} … \\ %\midrule
How old was Messe when the First World War started? \textbf{Gold Answer 30. Predicted answer: 28}. & Giovanni Messe: Messe was born ... on \textbf{10 December 1883}. 20th-century events: The First World War ... started in 1914 and ended in 1918... Military   career of Adolf Hitler: He was 25 years old in \textbf{August 1914}, when Austria-Hungary and the German Empire entered the First World War. \\ %\midrule
What albums were ranked higher than "It Takes a Nation of Millions to Hold Us Back" in Rolling Stone's   the 500 Greatest Albums of All Time? \textbf{Gold answer: \textless{}no answer\textgreater{}.   Predicted answer: the beatles}. & It Takes a Nation of Millions to Hold Us Back: ... In 2003, Rolling Stone ranked the album number \textbf{48} on its list of the 500 Greatest Albums of All   Time... maintaining the rating in a 2012 revised list. Rolling Stone's 500   Greatest Albums of All Time: ... \textbf{topped} by the Beatles' 1967 album \textbf{"Sgt. Pepper's Lonely Hearts Club Band"}, with a top 10 that featured four   entries from the Beatles (Nos. 1, 3, 5 and 10), two from Bob Dylan (No. 4 and   9), and one each from the Beach Boys (No. 2), Marvin Gaye (No. 6), the   Rolling Stones (No. 7) and the Clash (No. 8). \\ %\midrule
In what direction does the Goulburn River flow to Sugarloaf Creek? \textbf{Gold answer: \textless{}no answer\textgreater{}. Predicted answer: north west}. & Charles Bonney: ... was the first to overland sheep, bringing some 10,000   ... to \textbf{Sugarloaf Creek, Victoria station a tributary of the Goulburn River}...  Goulburn River: ... The river flows generally north, then west, then north, then west... \\ \bottomrule
\end{tabular}
%\end{adjustbox}
\caption{Example failure cases for IIRC\textsubscript{R} samples on the \textit{Base+RATD} model. The top two rows have numeric labels, the bottom two are labelled unanswerable. Bolded context text highlights information that could be used in deriving an answer.}
\label{tab:failure_cases}
\end{table*}

\newpage
\section{StableVicuna FP16 Comparison To INT8 (Chapter \ref{sec:combine})}
\label{sec:app_summary_fp16}

Performance differences between FP16 and INT8 for StableVicuna are not statistically significant but recalling that here we use a greedy decoding method it is interesting to us that there is a difference at all. \\ \\

\begin{table}[h!]
\centering
\resizebox{\textwidth}{!}{%
\begin{tabular}{l|rrr|rrr|rrr}
\toprule
\textbf{Rationale Generator $\rightarrow$ } & \multicolumn{3}{c} {\textbf{StableVicuna  (FP16)}} & \multicolumn{3}{|c}{\textbf{StableVicuna (INT8)}} & \multicolumn{3}{|c}{\textbf{BLOOM (INT8)}} \\
\textbf{Context $\downarrow$ / \textit{Model} $\rightarrow$} & \textbf{\textit{GR}} & \textbf{\textit{RATD}} & \textbf{\textit{GR+RATD}} & \textbf{\textit{GR}} & \textbf{\textit{RATD}} & \textbf{\textit{GR+RATD}} & \textbf{\textit{GR}} & \textbf{\textit{RATD}} & \textbf{\textit{GR+RATD}} \\ \midrule
Iterator only & 38.1 & 40.4 & \textbf{41.0} & 38.1 & 40.4 & \textbf{41.0} & 38.1 & 40.4 & \textbf{41.0} \\
Rationale only & 44.6 & 44.4 & \textbf{45.5} & 44.5 & 44.2 & 45.3 & 39.5 & 42.0 & 40.3 \\
Rationale + Iterator (Naïve concatenation) & 42.9 & 46.4 & 47.1 & 42.7 & 46.3 & \textbf{47.2} & 43.2 & 43.8 & 43.7 \\
Rationale + Iterator (Generally best RR combo) & 45.4 & 46.4 & 47.1 & 45.5 & 46.3 & \textbf{47.2} & 42.9 & 44.2 & 44.4 \\ \midrule%\hdashline
Rationale + Iterator (Best RR combo per dataset) & 47.8 & 47.5 & 48.0 & 47.6 & 47.5 & \textbf{48.1} & 45.1 & 45.6 & 45.4 \\ \bottomrule
\end{tabular}%
}
\caption{Mean score over unseen evaluation datasets. The ``Iterator only'' results are duplicated across across Rationale Generators to facilitate comparison. Bold indicates highest score per context type (i.e. per row).}
\label{tab:main_1_including_fp16}
\end{table}

\newpage
\section{Context Component Analysis (Chapter \ref{sec:combine})}
\label{sec:app_context_component_analysis}

As noted we do not consider the ``Best RR combo per dataset'' to be a viable method for answering arbitrary questions of unknown type, however in Table \ref{tab:context_type_percentages} we report the best combination method identified for each individual evaluation dataset as it shows what an oracle-like method is capable of producing in comparison to our actual generally-best \textit{RR}-scoring method. Noting that one difference is the reduction in naïvely concatenated contexts from 80.9\% to 27.9\% it is plausible that future work on a more refined combination strategy would yield further improvement in combining \textit{RATD} training with \textit{RR} scoring methods. \\ \\

\begin{table}[h]
\centering
\resizebox{\columnwidth}{!}{%
\begin{tabular}{ll|lrrr|rrr}
\toprule
\textbf{Dataset} & \textbf{Sample} & \multicolumn{4}{c|}{\textbf{Best RR combo per dataset}} & \multicolumn{3}{|c}{\textbf{Generally best RR combo:} EitherOrBoth(0.9)}\\
\textbf{} & \textbf{Count} & \textbf{Best Method} & \textbf{Naïve Concat.} & \textbf{Rat. Only} & \textbf{Iter. Only} & \textbf{Naïve Concat.} & \textbf{Rat. Only} & \textbf{Iter. Only} \\ \midrule
SQA & 2290 & RationaleDefault(0.75) & 0.0 & 90.7 & 9.3 & 94.1 & 3.6 & 2.3 \\
CSQA & 1221 & RationaleDefault(0.75) & 0.0 & 98.3 & 1.7 & 79.3 & 20.6 & 0.1 \\
ARC-DA & 1397 & Naïve concatenation& 100.0 & 0.0 & 0.0 & 80.5 & 16.5 & 3.1 \\
IIRC & 1301 & RationaleDefault(0.9) & 0.0 & 63.8 & 36.2 & 62.6 & 15.6 & 21.8 \\
Musique & 2417 & EitherOrBoth(0.14) & 39.3 & 3.2 & 57.5  & 88.2 & 1.0 & 10.8 \\ \midrule
\textbf{Mean} & & & \textbf{27.9} & \textbf{51.2} & \textbf{20.9}  & \textbf{80.9} & \textbf{11.5} & \textbf{7.6} \\ \bottomrule
\end{tabular}%
}
\caption{Best combination method per dataset on the \textit{GR+RATD} model. Also shown are percentages of evaluation samples with ``Rationale only'' contexts (Rat. Only), ``Iterator only'' contexts (Iter. only), and the concatenation of both (Naïve Concat) respectively.}
\label{tab:context_type_percentages}
\end{table}